%% file: main.tex
\RequirePackage{fix-cm}
\documentclass{article}
\usepackage[OT1]{fontenc}
\usepackage{iclr2027_conference,times}

\usepackage{amsmath,amsfonts}
\usepackage{xcolor}
\usepackage{booktabs}
\usepackage{graphicx}

\definecolor{mycitecolor}{RGB}{65,123,110}
\usepackage[colorlinks=true,
            citecolor=mycitecolor,
            linkcolor=black,
            urlcolor=black]{hyperref}
\usepackage{pifont}

\usepackage{colortbl}
\definecolor{ppgrowblue}{RGB}{232,243,255}

\usepackage{url}
\usepackage{wrapfig}
\usepackage{multirow}

\title{\textbf{PPG-LM}: A Photoplethysmography--Language Model with Multi-Level Clinical Alignment}

\author{%
\makebox[\textwidth][l]{\hspace*{3pt}\textbf{Xiaoda Wang\textsuperscript{1,2}, Minxiao Wang\textsuperscript{1}, Maxwell A Xu\textsuperscript{3}, Patrick Langer\textsuperscript{4}, Kaiqiao Han\textsuperscript{2}, Defu Cao\textsuperscript{5}}}\\
\makebox[\textwidth][l]{\hspace*{3pt}\textbf{Xiao Luo\textsuperscript{6}, Yuzhe Yang\textsuperscript{2,3}, Yan Liu\textsuperscript{5}, Xiao Hu\textsuperscript{1}, Yizhou Sun\textsuperscript{2}, Wei Wang\textsuperscript{2}, Carl Yang\textsuperscript{1}}}\\
\makebox[\textwidth][l]{\hspace*{3pt}\textsuperscript{1}Emory University\quad
\textsuperscript{2}University of California, Los Angeles\quad
\textsuperscript{3}Google\quad
\textsuperscript{4}Stanford University}\\
\makebox[\textwidth][l]{\hspace*{3pt}\textsuperscript{5}University of Southern California\quad
\textsuperscript{6}University of Wisconsin--Madison}
}

\newcommand{\ppglm}{\textsc{PPG-LM}}

\iclrfinalcopy

\begin{document}

\begingroup
\setlength{\tabcolsep}{0pt}
\maketitle
\endgroup
\fancyhead{}
\lhead{Preprint}

\begin{abstract}

Photoplethysmography (PPG) is widely recorded by clinical monitors and consumer wearables, providing a scalable source of continuous physiological information. These recordings offer an opportunity for physiological assessment at scale, but realizing this potential requires models to learn from both signal-derived physiological supervision and broader clinical context captured in electronic health records (EHRs). This involves aligning information spanning local observations, care events, and entire visits with PPG representations at corresponding temporal scales. However, existing PPG foundation models primarily rely on task-specific prediction heads, while the medical knowledge of large language models does not necessarily translate into waveform understanding.
To bridge this gap, we introduce \textbf{\ppglm{}}, the first PPG--language model family to learn physiological representations from both signal-derived supervision and broader clinical context captured in EHRs.
To construct clinically grounded captions, we develop an automatic captioning pipeline that generates \mbox{segment-,} \mbox{event-,} and visit-level descriptions from signal measurements and structured EHR records.
We then learn from these pairs through a two-stage framework that first establishes segment--language correspondence through contrastive learning and waveform-conditioned captioning, then extends alignment to events and visits through time-aware aggregation and temporal statement matching.
Pretrained on approximately 73k hours of PPG, \ppglm{} supports language-based recognition, cross-modal retrieval, and segment captioning.
Experiments on MC-MED, MIMIC-III, and VitalDB show improved retrieval and caption factuality over language-model baselines and gains over PPG and time-series foundation models on multiple clinical prediction tasks.

\end{abstract}

\input{Sections/1_intro}

\input{Sections/2_related_work}

\input{Sections/3_method}

\input{Sections/4_experiment}
\input{Sections/5_conclusion}

\subsection*{AI use statement}

During manuscript preparation, we used generative AI tools solely for language polishing, including grammar correction, sentence refinement, and improvements to clarity and readability. The authors reviewed all suggested edits to ensure that the revised text preserved the intended technical meaning and accurately reflected the methods and results. The authors retained final editorial control and take full responsibility for the manuscript's content, claims, and conclusions.

\subsection*{Ethics statement}

This study involves secondary analysis of de-identified data from MC-MED, MIMIC-III, and VitalDB \citep{mcmed,mimic3,vitaldb}, in accordance with their respective access requirements and data-use agreements, with no new patient recruitment.
\ppglm{} is intended for research and has not been validated for clinical decision-making.
Medication-event predictions reflect associations and do not establish causal treatment effects.
Prospective validation and assessment of subgroup performance and generalizability are needed before clinical deployment, as discussed in Appendix~\ref{app:limitations}.

\subsection*{Reproducibility statement}

Sections~\ref{sec:captions}--\ref{sec:experiments} describe caption construction, the model framework, and the experiments.
Appendices~\ref{app:captions} and~\ref{app:prompts} provide captioning rules, language templates, and evaluation prompts, while Appendix~\ref{app:training-details} details the architecture, training objectives, hyperparameters, and computational setup.
Dataset preprocessing, patient-disjoint splits, baseline configurations, task definitions, and evaluation protocols are documented in Appendix~\ref{app:experiment-details}.

\bibliography{reference}
\bibliographystyle{iclr2027_conference}

\appendix
\input{Sections/6_appendix}

\end{document}

%% file: Sections/1_intro.tex
\section{Introduction}

Photoplethysmography (PPG) is a low-cost optical measure of peripheral blood-volume changes, widely recorded by clinical monitors and wearables \citep{allen2007ppg,charlton2022wearable}.
These recordings provide cardiac and vascular information at scale, motivating foundation models that learn generalizable waveform representations \citep{papagei,pulseppg,anyppg}.
However, existing PPG foundation models primarily support task-specific prediction, with limited support for recognizing, retrieving, or describing physiological observations through clinical language.
Recent efforts to adapt large language models (LLMs) to PPG question answering \citep{pulselm} address a complementary gap: textual medical knowledge alone does not ensure an understanding of pulse morphology and timing \citep{wang2026position}.
PPG and Language alignment offers a shared interface for these capabilities, with demonstrated benefits in medical imaging \citep{gloria}, wearable sensing \citep{sensorlm}, and sleep analysis \citep{sleeplm}.
For PPG, electronic health records (EHRs) provide clinical context, including patient presentation, medical history, and care events, that complements signal-derived physiological measurements \citep{cap,mcmed}.
Expressing these complementary sources as language supervision offers an opportunity to learn both waveform characteristics and broader clinical associations.

Training a PPG--language model to learn both fine-grained waveform characteristics and broader clinical information poses two coupled challenges:
(i) \textit{Clinically grounded caption generation.}
Clinical records contain heterogeneous measurements, histories, and care events but do not directly provide descriptions paired with PPG observations.
Constructing useful supervision requires selecting and verbalizing information from both signal measurements and EHRs, distinguishing measured waveform properties from documented clinical context, and organizing the resulting descriptions at segment, event, and visit levels.
(ii) \textit{Multi-level temporal alignment.}
These descriptions correspond to individual windows, observations around care events, and variable-length visits.
Event- and visit-level alignment therefore requires aggregating multiple windows, but pooling without temporal information can obscure physiological changes.
A model must capture these temporal relationships while maintaining segment--language correspondence as broader clinical supervision is introduced.

\begin{figure}[t]
    \centering
    \includegraphics[width=0.99\linewidth]{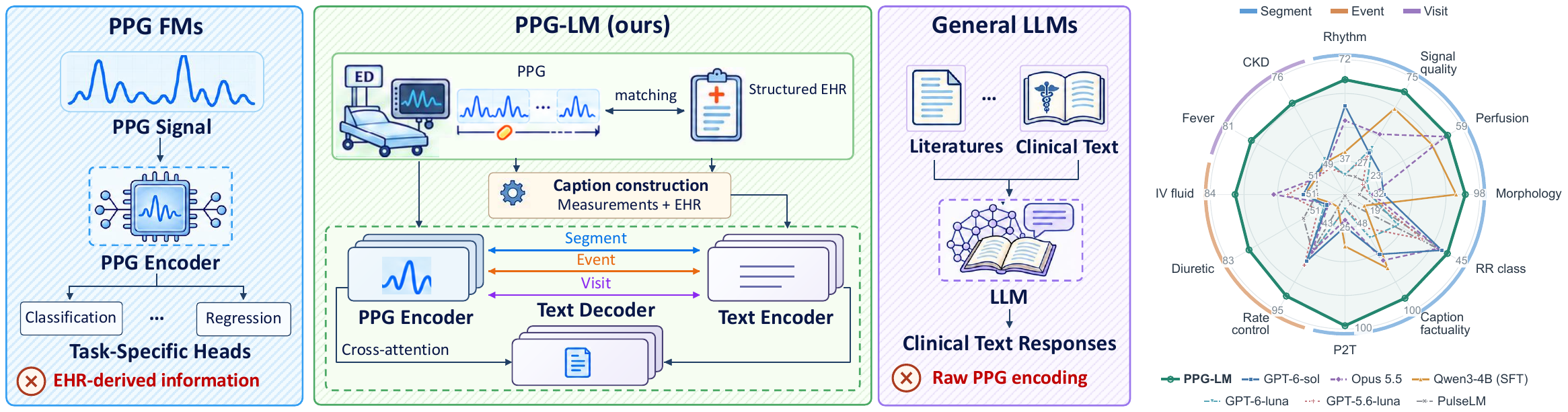}
    \caption{
        \textbf{Motivation and overview of PPG-LM.}
    \textcolor[HTML]{2878C8}{\textbf{PPG FMs}}
    learn waveform representations with limited integration of
    language and EHR context.
    \textcolor[HTML]{8854C8}{\textbf{General LLMs}}
    capture medical knowledge from text but require adaptation
    to encode raw PPG.
    \textcolor[HTML]{2E8B57}{\textbf{PPG-LM}}
    bridges this gap by pairing PPG with captions derived from
    physiological measurements and structured EHRs.
    The radar plot highlights PPG-LM's superior performance
    over baselines on selected segment-, event-,
    and visit-level tasks.
    } 
    \label{fig:ppglm_intro}
\end{figure}

\begin{wraptable}{r}{0.5\linewidth}
  \vspace{-\intextsep}
  \setlength{\abovecaptionskip}{0pt}
  \caption{Comparison of PPG--language studies.}
  \vspace{0.2em}
  \label{tab:ppg_text_comparison}
  \centering
  \scriptsize
  \setlength{\parskip}{0pt}
  \setlength{\tabcolsep}{0.6pt}
  \renewcommand{\arraystretch}{1.08}

  \newcommand{\ppgcheck}{\textcolor{green!45!black}{\ding{51}}}
  \newcommand{\ppgcross}{\textcolor{red!80!black}{\ding{55}}}
  \newcommand{\modelcite}[2]{%
    #1\,{\fontsize{6}{7}\selectfont\citep{#2}}}

  \resizebox{\linewidth}{!}{%
  \begin{tabular}{@{}lcccccc@{}}
    \toprule
    \multirow{2}{*}[-2pt]{\textbf{Study}}
      & \multicolumn{2}{c}{\textbf{Signal}}
      & \multirow{2}{*}[-2pt]{\textbf{Text Source}}
      & \multicolumn{3}{c}{\textbf{Text scope}} \\
    \cmidrule(lr){2-3}\cmidrule(lr){5-7}
    & Raw & Context & & Segment & Event & Visit \\
    \midrule

    \modelcite{SensorLM}{sensorlm}
      & \ppgcross & 1 day & Sensor
      & \ppgcheck & \ppgcross & \ppgcross \\
    \modelcite{NormWear}{normwear}
      & \ppgcheck & 6 s & Labels
      & \ppgcheck & \ppgcross & \ppgcross \\
    \modelcite{CSFM}{csfm}
      & \ppgcheck & 10 s & Labels
      & \ppgcheck & \ppgcross & \ppgcross \\
    \modelcite{PulseLM}{pulselm}
      & \ppgcheck & 10 s & Labels
      & \ppgcheck & \ppgcross & \ppgcross \\
    \modelcite{CAP}{cap}
      & \ppgcheck & 5 min & EHR
      & \ppgcheck & \ppgcross & \ppgcross \\

    \midrule
    \rowcolor{ppgrowblue}
    \textbf{PPG-LM}
      & \ppgcheck
      & \textbf{Variable}\hspace{2pt}
      & \textbf{Signal+EHR}
      & \ppgcheck & \ppgcheck & \ppgcheck \\
    \bottomrule
  \end{tabular}%
  }
\vspace{-\intextsep}
\end{wraptable}

To address these challenges, we introduce \textbf{\ppglm{}}, a family of PPG--language models trained with clinically grounded, multi-level supervision.
To construct clinically grounded captions, we develop an automatic captioning pipeline combines signal measurements and structured EHR records to generate segment-, event-, and visit-level descriptions, each paired with PPG observations over the corresponding temporal scope.
We then learn these correspondences through a two-stage alignment framework with waveform and text encoders shared across levels.
Stage~1 learns segment--language correspondence from captions derived from signal measurements and time-matched vital signs, combining contrastive learning and waveform-conditioned captioning with auxiliary clinical-attribute supervision.
Stage~2 extends alignment to medication events and variable-length visits: signed time offsets guide event aggregation, while global and time-anchored visit latents support clinical-caption alignment and temporal statement matching.
Our main contributions are as follows:

\ding{182} \textbf{\ppglm{}, a family of PPG--language models.}
    We introduce the first PPG--language model family to learn physiological representations from multi-level clinical language supervision, training on approximately 73,700 hours of PPG to support recognition, retrieval, and captioning.

\ding{183} \textbf{Clinically grounded multi-level caption generation.}
    We develop an automatic pipeline that combines signal measurements and structured EHR records to generate segment-, event-, and visit-level captions, preserving the evidence source and temporal scope of each description.

\ding{184} \textbf{Two-stage multi-level alignment.}
    We propose a two-stage framework that first learns segment--language correspondence through contrastive learning and captioning, then extends alignment to event and visit levels through time-aware aggregation and temporal statement matching.

\ding{185} \textbf{Comprehensive evaluation.}
    Experiments on MC-MED, MIMIC-III, and VitalDB assess recognition, retrieval, captioning, temporal understanding, and clinical prediction against PPG FMs, and LLMs, demonstrating language capabilities and transferable representations across care settings.

\vspace{2em}

%% file: Sections/2_related_work.tex
\section{Related Work}
\label{sec:related}

\textbf{Foundation models for PPG and time series.}
PPG foundation models learn transferable representations through contrastive and non-contrastive learning with participant and temporal cues \citep{wearablefm,siamquality}, morphology-aware objectives, and uncurated wearable data \citep{papagei,pulseppg}.
Generative approaches use autoregressive prediction and masked reconstruction \citep{gptppg,sigmappg,ppgpt}, while cross-signal methods use synchronized ECG and respiratory supervision \citep{anyppg,robustppg}.
However, their signal-centered objectives leave waveform--language correspondence unmodeled, and clinical applications typically require task-specific predictors.

\textbf{LLMs for PPG modeling.}
LLM-based methods use numerical prompting and parameter-efficient adaptation for physiological estimation \citep{llmbp,peftvitals}.
PulseLM~\citep{pulselm} pairs pretrained PPG encoders with instruction-tuned LLMs for question answering, while arrhythmia systems combine signal--text alignment with retrieval-augmented report generation \citep{ppgreport}.
Video-based remote PPG methods use text prototypes and physiological prompts for hemodynamic modeling \citep{physllm}.
These methods center on prescribed estimates, answers, or reports, leaving joint waveform--language learning across clinical temporal levels unaddressed.

\textbf{Time series and language alignment.}
Time series--language models pair sensor signals with activity, trend, behavioral, and affective descriptions \citep{imu2clip,sensorllm,sensorlm,eevr,chen2026learning}, and clinical signals with text across ECG, EEG, sleep, and multivariate recordings \citep{melp,neurolm,sleeplm,opentslm}.
PPG--EHR alignment uses medical histories as patient-level anchors \citep{cap}.
Multi-level objectives capture local and global semantics in medical imaging and video \citep{gloria,mgca,hiervl,hecvl,biovilt}, and finer physiological structure in ECG and sleep \citep{melp,sleeplm}.
However, these approaches do not directly support PPG--language alignment across multi-level clinical timescales: none jointly grounds local physiological descriptions, care-event context, and visit-level clinical information.

%% file: Sections/3_method.tex

\section{Clinically Grounded Caption Generation}
\label{sec:captions}

\begin{figure}[h]
    \centering
    \includegraphics[width=0.98\linewidth]{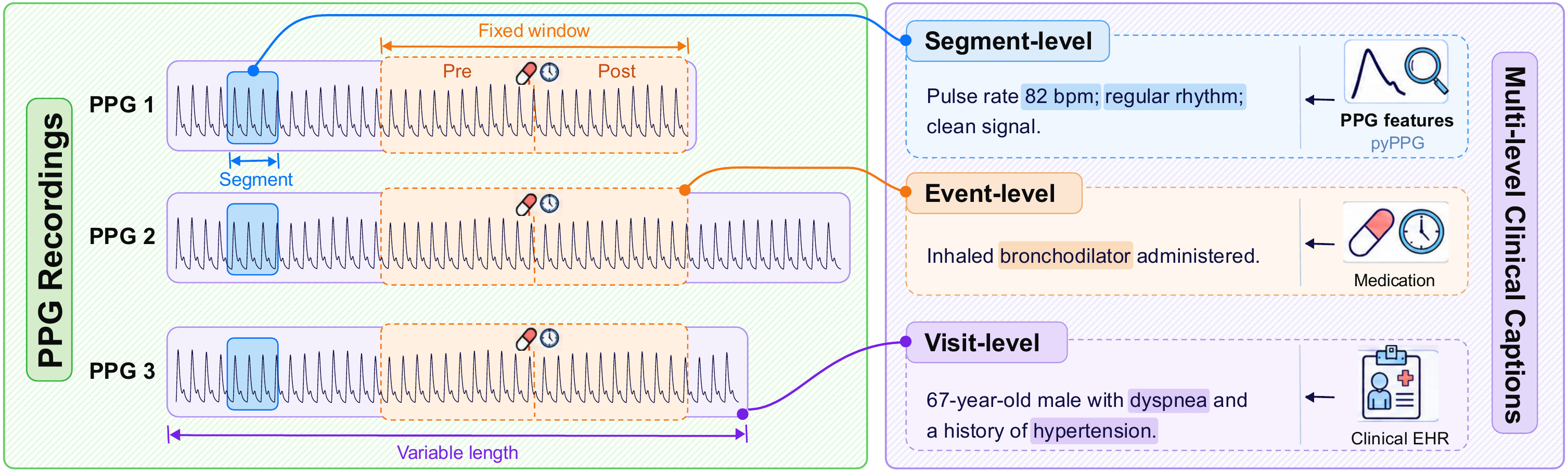}
    \vspace{-0.8em}
    \caption{\textbf{Clinically Grounded Caption Generation:}
    (1) \textit{\textcolor[HTML]{1677FF}{Segment-level}} captions describe physiological features and signal quality within 30-second PPG windows.
    (2) \textit{\textcolor[HTML]{E87516}{Event-level}} captions provide medication context for fixed windows before and after administration.
    (3) \textit{\textcolor[HTML]{7C2AE8}{Visit-level}} captions associate variable-length recordings with demographics, and medical history from clinical EHRs.}
    \vspace{-0.8em}
    \label{fig:caption-generation}
\end{figure}

To train \ppglm{}, we construct a corpus from MC-MED \citep{mcmed}, which pairs PPG recordings with structured electronic health records (EHRs) from emergency-department visits.
We generate segment-, event-, and visit-level captions from signal measurements and EHRs, pairing them with PPG observations at the corresponding temporal scales (Figure~\ref{fig:caption-generation}).
Captions use fixed language templates with reproducible template selection and sentence shuffling.
Corpus details, captioning rules, and template examples are provided in Appendices~\ref{app:experiment-setup}, \ref{app:captions}, and~\ref{app:prompts}, respectively.

\textbf{\textit{\textcolor[HTML]{1677FF}{Segment-level}} captions.}
\label{sec:captions-seg}
Using pyPPG \citep{pyppg}, we describe up to eight attributes per PPG window:
pulse rate, rate trend, rhythm regularity, pulsatile amplitude, signal quality,
augmentation-related morphology, a vascular-response index, and beat-to-beat
amplitude variability. Available fields yield numerical or qualitative descriptions.

\textbf{\textit{\textcolor[HTML]{E87516}{Event-level}} captions.}
\label{sec:captions-evt}
We extract administration timestamps, drug names, and routes from EHR medication
records. Route filtering and drug-name rules map eligible records
to ten classes; near-duplicates are removed within each visit and class.
For an administration at time $a$, we select up to three latest windows in
$[a-30,a)$ and up to five evenly spaced windows in $[a,a+60]$ (times in minutes).
The caption names the administered class and is paired with these windows.
Controls follow the same sampling rule within the visit, excluding anchors near
recorded administrations.

\textbf{\textit{\textcolor[HTML]{7C2AE8}{Visit-level}} captions.}
\label{sec:captions-vis}
We combine encounter records and demographics to summarize age, sex, arrival
mode, triage acuity, normalized chief complaints, and first recorded vital signs.
Prior-diagnosis codes are mapped to clinical descriptions, deduplicated, and
ranked by relevance and recency; active home medications are summarized by class.
These fields form one caption paired with observations across the encounter.
Ordered segment rates and rhythm labels additionally yield temporal statements
of state \emph{occurrence}, \emph{proportion}, contiguous \emph{duration}, and
rate \emph{ordering}.

\section{Method}
\label{sec:method}


\begin{figure}[t]
    \centering
    \includegraphics[width=0.98\linewidth]{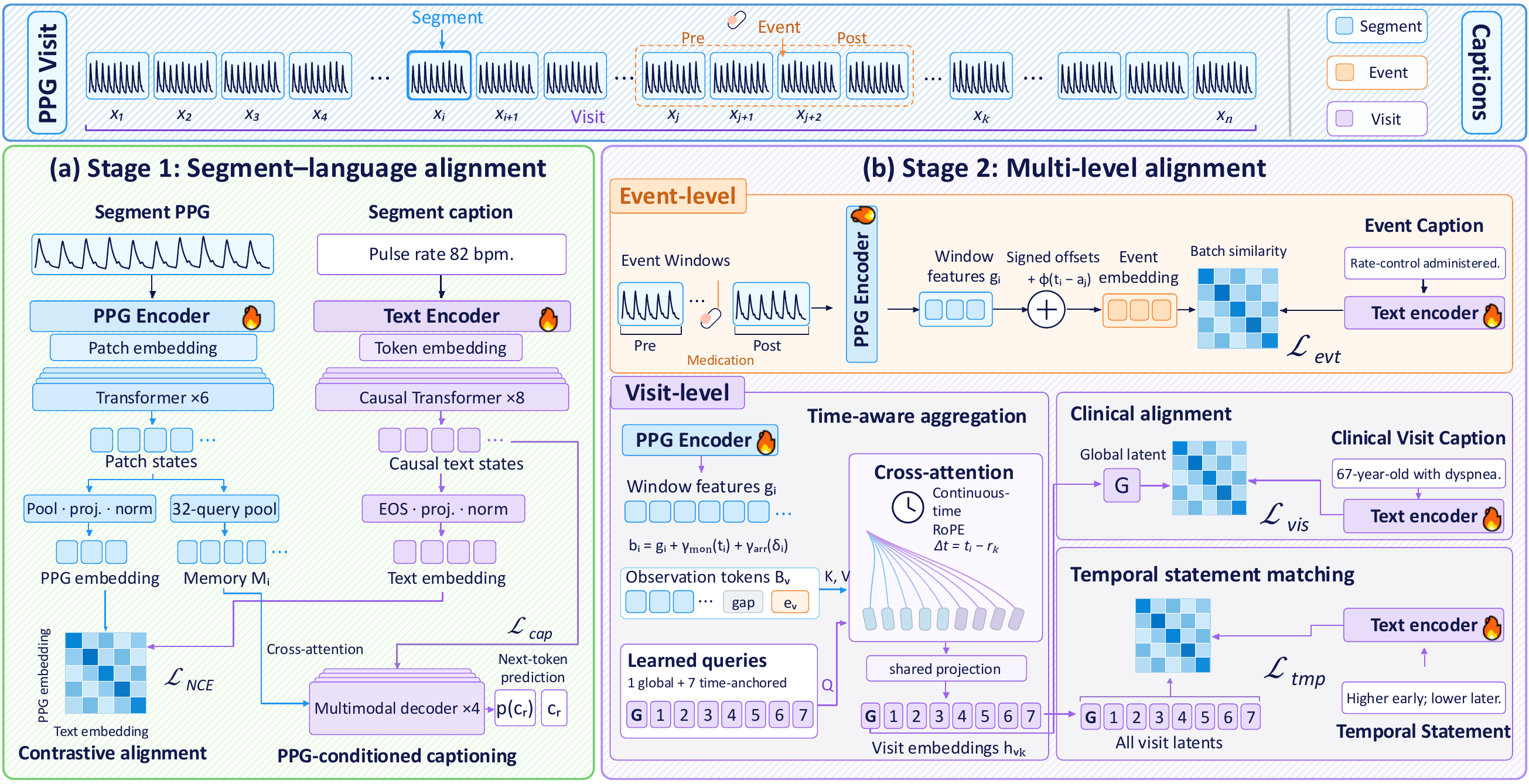}
    \vspace{-0.8em}
    \caption{\textbf{Two-stage multilevel alignment.}
    \textbf{(a)} Stage~1 learns PPG and text encoders through contrastive segment--language alignment and PPG-conditioned caption generation.
    \textbf{(b)} Stage~2 extends the shared encoders to event- and visit-level alignment.}
    \vspace{-1em}
    \label{fig:method-overview}
\end{figure}

The captions in Section~\ref{sec:captions} pair local physiological measurements
and broader clinical context with PPG observations at different temporal scopes.
Learning from these pairs requires both a window-level language interface and
representations that combine observations across time. \ppglm{} therefore uses
shared waveform and text encoders with scope-specific read-outs, trained in two
stages (Figure~\ref{fig:method-overview}). Stage~1 establishes segment--language
correspondence through contrastive learning and captioning. Stage~2 learns to
aggregate window features for event and visit descriptions, adding temporal
statement supervision while retaining the segment objective.

\textbf{Problem definition.}
A visit is a time-ordered sequence $V=((x_i,t_i))_{i=1}^{n_V}$, where
$x_i\in\mathbb{R}^{3750}$ is a standardized 30-second PPG window and $t_i$
is its start time in minutes since the first window. A caption $c^{\ell}$,
$\ell\in\{\mathrm{seg},\mathrm{evt},\mathrm{vis}\}$, is paired with a
single window, windows around a medication or control anchor, or observations
distributed across a visit, respectively. In addition to its clinical caption,
a visit may have temporal statements $T_V=\{d_{Vr}\}_{r=1}^{m_V}$ describing
state occurrence, proportion, contiguous duration, and rate ordering.

\textbf{PPG and text encoders.}
We share both encoders across levels to reuse waveform and language features,
while adapting their read-outs to each description's scope.
The PPG encoder $f_\theta$ divides a window into non-overlapping patches,
adds learned positional embeddings, and applies Transformer blocks
\citep{transformer}. Attention pooling summarizes the resulting patch states
$H_i$ into a window feature $g_i$. The text encoder $\psi_\phi$ is a causal
Transformer using CLIP's byte-pair encoding (BPE) vocabulary \citep{clip}.
A learned prefix $e_\ell$ identifies the description level, and a
level-specific projection $P_\ell$ maps the end-of-sequence (EOS) state
to a normalized text embedding:
\(
g_i=f_\theta(x_i)
\),
\(
u^\ell(c)=\operatorname{norm}\!\left(P_\ell\,
\psi_\phi([e_\ell;c])_{\mathrm{EOS}}\right),
\label{eq:representations}
\)
where $\operatorname{norm}(v)=v/\|v\|_2$. Waveform read-outs operate on
$g_i$ for segment alignment and on collections of window features for
event and visit alignment. Encoder details are given in
Appendix~\ref{app:training-architecture}.

\subsection{Stage 1: Segment--Language Pretraining}
\label{sec:method-stage1}

Following CoCa \citep{coca}, Stage~1 combines contrastive alignment of
windows and descriptions with token-level caption supervision.
For a batch of $B$ window--caption pairs $(x_i,c_i)$ from
Section~\ref{sec:captions}, we form normalized waveform embeddings
$z_i^{\mathrm{seg}}=\operatorname{norm}(W_{\mathrm{seg}}g_i)$ and logits
$Q_{ij}=s_{\mathrm{seg}}(z_i^{\mathrm{seg}})^\top u^{\mathrm{seg}}(c_j)$,
with learned scale $s_{\mathrm{seg}}>0$. To avoid treating captions with
identical binned measurements as negatives, we assign content keys $\kappa_i$
and use $\mathcal C_i^{\mathrm{seg}}=\{i\}\cup
\{j:\kappa_j\neq\kappa_i\}$ within the batch. For candidate sets
$\mathcal C$, InfoNCE \citep{cpc} in CLIP's symmetric form \citep{clip} is
\begin{equation}
\mathcal{L}_{\mathrm{NCE}}(Q;\mathcal{C})
=-\frac{1}{2B}\sum_{i=1}^{B}\left[
\log\frac{\exp(Q_{ii})}{\sum_{j\in\mathcal{C}_i}\exp(Q_{ij})}
+\log\frac{\exp(Q_{ii})}{\sum_{j\in\mathcal{C}_i}\exp(Q_{ji})}\right].
\label{eq:contrastive}
\end{equation}
The two terms align signals and text in both directions; later levels reuse
this loss with their respective logits and candidate sets.

For complementary token-level supervision, a separate attention pooler forms
waveform memory $M_i$ from the shared patch states $H_i$. A multimodal
decoder receives causal token states from the shared text encoder and
cross-attends to $M_i$ to predict caption tokens. With teacher forcing, we
minimize $\mathcal L_{\mathrm{cap}}=-N_{\mathrm{tok}}^{-1}\sum_{i,r}
\log p_{\phi,\eta}(c_{i,r}\mid c_{i,<r},M_i)$ over the
$N_{\mathrm{tok}}$ non-padding targets, including EOS; $\eta$ denotes
the additional captioning parameters. The combined objective
$\mathcal L_{\mathrm{seg}}=\mathcal L_{\mathrm{NCE}}(Q;\mathcal C^{\mathrm{seg}})
+\mathcal L_{\mathrm{cap}}$ trains both shared encoders from scratch and
is retained in Stage~2.

\subsection{Stage 2: Multi-Level Alignment}
\label{sec:method-stage2}

Event and visit descriptions concern multiple observations. Stage~2 therefore
adds temporal aggregation over the shared window features and aligns the
resulting representations with descriptions at the corresponding scope.

\textbf{Event-level alignment.}
Event captions pair a medication class or no-administration control with
nearby PPG windows (Section~\ref{sec:captions}). For view $j$, we preserve
timing relative to its anchor $a_j$ by adding signed offset embeddings
$\varphi(t_i-a_j)$ and pooling features over selected window indices $E_j$
with a learned query:
\begin{equation}
z_j^{\mathrm{evt}}=\operatorname{norm}\!\left(
W_{\mathrm{evt}}\operatorname{Attn}
\left(q_{\mathrm{evt}};\{g_i+\varphi(t_i-a_j)\}_{i\in E_j}\right)\right).
\label{eq:event}
\end{equation}
Offsets are measured in minutes on the visit clock, distinguishing
observations before and after the anchor and their temporal distance.
Using Equation~\ref{eq:contrastive}, we minimize
$\mathcal L_{\mathrm{evt}}=\mathcal L_{\mathrm{NCE}}(Q^{\mathrm{evt}};\mathcal C^{\mathrm{evt}})$,
with logits $Q_{jk}^{\mathrm{evt}}=s_{\mathrm{evt}}(z_j^{\mathrm{evt}})^\top
u^{\mathrm{evt}}(c_k^{\mathrm{evt}})$ and learned scale $s_{\mathrm{evt}}>0$.
Candidate sets retain the matched pair and exclude other views of the same
class to avoid contrasting identical treatment categories; controls form
one class. Attention details appear in Appendix~\ref{app:training-aggregator}.

\textbf{Visit-level alignment.}
A clinical visit caption summarizes the encounter, while temporal statements
may require evidence from different parts of its trajectory. To bound
computation, we select at most 256 windows indexed by $\mathcal I_V$ across
the monitoring span. We encode monitoring time $t_i$ and time since ED arrival $\delta_i$
as $b_i=g_i+\gamma_{\mathrm{mon}}(t_i)+\gamma_{\mathrm{arr}}(\delta_i)$.
The input $B_V=[(b_i)_{i\in\mathcal I_V};G_V;e_V]$ also includes gap tokens
$G_V$ for monitoring interruptions and an evidence token $e_V$ for coverage
and quality.

Two attention blocks aggregate $B_V$ through learned queries \citep{perceiver}.
We fix their number at eight: one global query and seven with reference
times $r_k$ relative to the first window \citep{mtan}.
We apply RoFormer's rotary position embeddings (RoPE) \citep{roformer} to
actual time offsets in minutes. For an anchored query $q_k$ and a window key $k_i$,
\begin{equation}
\tilde q_k^\top\tilde k_i
=(q_k^{\mathrm{rot}})^\top R_\Theta(t_i-r_k)k_i^{\mathrm{rot}}
 +(q_k^{\mathrm{free}})^\top k_i^{\mathrm{free}},
\qquad k=1,\ldots,7.
\label{eq:time-rope}
\end{equation}
Here $R_\Theta$ rotates the $\mathrm{rot}$ channels and $\mathrm{free}$
channels remain unchanged. All queries attend to every valid input token;
anchors do not restrict attention to fixed intervals. The output latents $\ell_{Vk}$
share a projection $h_{Vk}=\operatorname{norm}(W_{\mathrm{lat}}\ell_{Vk})$,
$k=0,\ldots,7$. We align the global embedding $z_V^{\mathrm{vis}}=h_{V0}$
with $u^{\mathrm{vis}}(c_V^{\mathrm{vis}})$ through in-batch InfoNCE,
$\mathcal L_{\mathrm{vis}}$, and use all eight $h_{Vk}$ for temporal statement
matching. Encoding and anchor details appear in Appendix~\ref{app:training-aggregator}.

\textbf{Temporal statement matching.}
We adapt ColBERT's late interaction \citep{colbert} to match each statement
to its most compatible visit latent. For trajectory $i$ and a visit
$j$ with $m_j>0$ available statements, let $v_{jr}=u^{\mathrm{vis}}(d_{jr})$:
\(
R_{ij}=\frac{1}{m_j}\sum_{r=1}^{m_j}\max_{k=0,\dots,7}
h_{ik}^{\top}v_{jr}.
\label{eq:temporal-score}
\)
The mean normalizes for statement count, and each statement can select a
different latent without a prescribed statement-to-anchor assignment.
Alongside in-batch alignment, we rank the true set above a version $i^-$
with one incompatible statement:
\begin{equation}
\mathcal{L}_{\mathrm{tmp}}
=\mathcal{L}_{\mathrm{NCE}}(s_{\mathrm{vis}}R;\mathcal{C}^{\mathrm{full}})
 +\tfrac12\,\mathbb{E}_i\!\left[
 -\log\sigma\!\left(s_{\mathrm{vis}}(R_{ii}-R_{i,i^-})\right)\right],
\label{eq:temporal-objective}
\end{equation}
where $\sigma$ is the logistic function and $\mathcal{C}^{\mathrm{full}}$
admits all in-batch pairs. Changing one statement preserves most of the
description while making its temporal content the discriminating target.
Missing statements and unavailable corruptions are masked.

Stage~2 initializes from Stage~1 and alternates segment, visit, and event
batches, retaining segment supervision to preserve local alignment
\citep{hiervl}. For batch type
$b\in\{\mathrm{seg},\mathrm{evt},\mathrm{vis}\}$ and windows $\mathcal Q$
re-encoded with gradients, the alignment objective is
\begin{equation}
\mathcal L^{(b)}
=\mathcal L_{\mathrm{seg}}(\mathcal Q)
+\lambda_{\mathrm{dst}}\mathcal L_{\mathrm{dst}}(\mathcal Q)
+\mathbb I_{b=\mathrm{evt}}\mathcal L_{\mathrm{evt}}
+\mathbb I_{b=\mathrm{vis}}(\mathcal L_{\mathrm{vis}}+\mathcal L_{\mathrm{tmp}}).
\label{eq:joint}
\end{equation}
Here $\mathbb I$ is an indicator, and $\mathcal L_{\mathrm{dst}}$ is the
mean cosine distance to frozen Stage-1 window embeddings.
Training schedules, hyperparameter settings, and inference procedures are
given in Appendix~\ref{app:training-schedule}.

%% file: Sections/4_experiment.tex
\section{Experiments}
\label{sec:experiments}

\textbf{Datasets.}
We use MC-MED \citep{mcmed} for pretraining and in-domain evaluation, with patient-disjoint training, validation, and test splits. MC-MED pairs PPG recordings from Stanford's adult emergency department (ED) with structured EHRs. For external evaluation, we use MIMIC-III \citep{mimic3}, which links intensive-care bedside waveforms from Beth Israel Deaconess Medical Center to clinical records, and VitalDB \citep{vitaldb}, which pairs intraoperative biosignals from Seoul National University Hospital with perioperative clinical information. 

\textbf{Baselines.}
For language-based tasks, we compare with four general-purpose LLMs---GPT-5.6-luna \citep{gpt56}, GPT-6-luna \citep{gpt6luna}, GPT-6-sol \citep{gpt6sol}, and Claude Opus~5.5 \citep{claudeopus55}. We also include the Qwen2.5-7B-Instruct variant of PulseLM \citep{pulselm}, trained with a PaPaGei PPG encoder \citep{papagei}. Separately, we train a Qwen3-4B (SFT) baseline \citep{qwen3} using a frozen PaPaGei-S encoder and supervised fine-tuning on PPG signals paired with captions generated by our captioning pipeline. For clinical prediction, we include PPG foundation models PaPaGei-S \citep{papagei}, Pulse-PPG \citep{pulseppg}, AnyPPG \citep{anyppg}, and SIGMA-PPG \citep{sigmappg}, together with general time-series foundation models MOMENT-L \citep{moment} and Chronos-2 \citep{chronos2}.

\subsection{Zero-Shot Physiological Understanding}
\label{sec:exp-language}

\begin{table}[!htbp]
\vspace{-\intextsep}
\centering\small
\vspace{-0.3em}
\caption{\textbf{Zero-shot physiological recognition.} Macro-F1 ($\uparrow$) for rhythm,
quality (Qual.), perfusion (Perf.), morphology (Morph.), and respiratory-rate
class (RR). Bold marks the best in each column.}
\label{tab:recognition}
\scriptsize
\setlength{\tabcolsep}{2pt}
\begin{tabular*}{\linewidth}{@{\extracolsep{\fill}}l*{11}{c}@{}}
\toprule
\multicolumn{1}{c}{\multirow{2}{*}[-0.5ex]{\textbf{Model}}} & \multicolumn{5}{>{\columncolor[HTML]{EAF2E6}}c}{\textbf{MC-MED}} & \multicolumn{3}{>{\columncolor[HTML]{FDF1E6}}c}{\textbf{MIMIC-III}} & \multicolumn{3}{>{\columncolor[HTML]{F2ECFC}}c}{\textbf{VitalDB}}\\
\cmidrule(lr){2-6}\cmidrule(lr){7-9}\cmidrule(lr){10-12}
 & Rhythm$\uparrow$ & Qual.$\uparrow$ & Perf.$\uparrow$ & Morph.$\uparrow$ & RR$\uparrow$ & Rhythm$\uparrow$ & Qual.$\uparrow$ & RR$\uparrow$ & Rhythm$\uparrow$ & Qual.$\uparrow$ & RR$\uparrow$\\
\midrule
GPT-5.6-luna & 0.322 & 0.314 & 0.174 & 0.259 & 0.332 & 0.376 & 0.291 & 0.266 & 0.317 & 0.162 & 0.313\\
GPT-6-luna & 0.319 & 0.361 & 0.193 & 0.263 & 0.259 & 0.357 & 0.315 & 0.253 & 0.326 & 0.175 & 0.307\\
GPT-6-sol & 0.560 & 0.332 & 0.250 & 0.354 & 0.390 & 0.572 & 0.295 & 0.380 & 0.375 & 0.128 & 0.326\\
Opus~5.5 & 0.508 & 0.436 & 0.521 & 0.284 & 0.380 & 0.598 & 0.343 & 0.397 & 0.330 & 0.228 & \textbf{0.330}\\
\midrule
PulseLM & 0.319 & 0.199 & 0.188 & 0.195 & 0.405 & 0.242 & 0.297 & 0.354 & 0.181 & 0.326 & 0.165\\
Qwen3-4B (SFT) &  0.399 & 0.578 & 0.464 & 0.826 & 0.158 & 0.272 & 0.564 & 0.171 & 0.231 & 0.515 & 0.015\\
\midrule
\noalign{\begingroup\color{ppgrowblue}\hrule height\dimexpr\ht\strutbox+\dp\strutbox\relax\endgroup\vskip-\dimexpr\ht\strutbox+\dp\strutbox\relax}
\textbf{PPG-LM} & \textbf{0.651} & \textbf{0.673} & \textbf{0.531} & \textbf{0.886} & \textbf{0.407} & \textbf{0.670} & \textbf{0.786} & \textbf{0.428} & \textbf{0.389} & \textbf{0.583} & 0.329\\
\bottomrule
\end{tabular*}
\end{table}
\vspace{-0.7em}

\textbf{Zero-shot physiological recognition.}
We test recognition of physiological categories from PPG in Table~\ref{tab:recognition}. Macro-F1 is reported for rhythm regularity, signal quality (Qual.), waveform-derived perfusion strength (Perf.) and augmentation-related morphology (Morph.), and respiratory-rate class (RR). \ppglm{} achieves the highest scores on 10 of 11 dataset--task pairs, including rhythm and quality on all three datasets and both perfusion and morphology on MC-MED. The exception is VitalDB RR, where it closely matches Opus~5.5 (0.329 vs.\ 0.330). These results support recognition of multiple waveform attributes through a shared language interface, with transfer across emergency, intensive-care, and surgical settings.

\textbf{Zero-shot physiological estimation.}
We assess whether generated descriptions preserve physiological facts and
numerical measurements in Table~\ref{tab:caption}. Fact measures the accuracy
of stated facts across six shared attributes; HR, RR, SBP, and MAP report
MAE for heart rate, respiratory rate, systolic blood pressure, and mean
arterial pressure, respectively. \ppglm{} achieves the highest fact accuracy
on MC-MED, MIMIC-III, and VitalDB (0.923, 0.920, and 0.820) and the lowest
reported errors for all displayed measurements. HR MAE is 3.90--4.27\,bpm against an independent
ECG reference, supporting physiological agreement beyond the PPG-derived
training captions. RR errors are computed on valid
outputs, so differences in output coverage limit their direct comparison.

\begin{table}[!htbp]
\vspace{-\intextsep}
\centering\small
\vspace{-0.5em}
\caption{\textbf{Zero-shot physiological estimation.}
Fact is the accuracy of stated facts across six shared attributes ($\uparrow$). Numerical columns
report MAE ($\downarrow$): HR in bpm, RR in breaths/min, and SBP/MAP in mmHg.
MAE uses valid numerical outputs; bold marks the best reported value in each column.}
\label{tab:caption}
\footnotesize
\setlength{\tabcolsep}{1.4pt}
\begin{tabular*}{\linewidth}{@{\extracolsep{\fill}}lccccccccccc@{}}
\toprule
\multicolumn{1}{c}{\multirow{2}{*}[-0.5ex]{\textbf{Model}}} & \multicolumn{5}{>{\columncolor[HTML]{EAF2E6}}c}{\textbf{MC-MED}} & \multicolumn{3}{>{\columncolor[HTML]{FDF1E6}}c}{\textbf{MIMIC-III}} & \multicolumn{3}{>{\columncolor[HTML]{F2ECFC}}c}{\textbf{VitalDB}}\\
\cmidrule(lr){2-6}\cmidrule(lr){7-9}\cmidrule(lr){10-12}
 & HR$\downarrow$ & RR$\downarrow$ & SBP$\downarrow$ & MAP$\downarrow$ & Fact$\uparrow$ & HR$\downarrow$ & RR$\downarrow$ & Fact$\uparrow$ & HR$\downarrow$ & RR$\downarrow$ & Fact$\uparrow$\\
\midrule
GPT-5.6-luna & 10.06 & 3.95 & 21.73 & 15.12 & 0.499 & 9.42 & 5.15 & 0.456 & 8.49 & 2.99 & 0.486\\
GPT-6-luna & 11.24 & 3.93 & 22.24 & 14.51 & 0.559 & 11.52 & 4.71 & 0.533 & 8.86 & 2.34 & 0.579\\
GPT-6-sol & 4.58 & 3.85 & 21.56 & 14.48 & 0.659 & 5.21 & 5.10 & 0.660 & 5.51 & 2.26 & 0.619\\
Opus~5.5 & 4.67 & 3.32 & 19.80 & 14.26 & 0.696 & 4.37 & 3.99 & 0.647 & 4.88 & 2.44 & 0.642\\
\midrule
PulseLM & 29.20 & 4.21 & 20.44 & 15.47 & 0.356 & 29.44 & 4.59 & 0.440 & 37.12 & 7.03 & 0.481\\
Qwen3-4B (SFT) & 7.36 & 5.10 & 20.14 & 13.93 & 0.743 & 8.54 & 5.20 & 0.665 & 8.98 & 7.80 & 0.674\\
\midrule
\noalign{\begingroup\color{ppgrowblue}\hrule height\dimexpr\ht\strutbox+\dp\strutbox\relax\endgroup\vskip-\dimexpr\ht\strutbox+\dp\strutbox\relax}
\textbf{PPG-LM} & \textbf{3.90} & \textbf{3.10} & \textbf{12.50} & \textbf{6.17} & \textbf{0.923} & \textbf{4.27} & \textbf{3.83} & \textbf{0.920} & \textbf{3.98} & \textbf{2.23} & \textbf{0.820}\\
\bottomrule
\end{tabular*}
\end{table}
\vspace{-0.5em}

\begin{samepage}
\setlength{\intextsep}{0pt}
\begin{wraptable}{r}{0.5\linewidth}
\centering\footnotesize
\setlength{\abovecaptionskip}{0pt}
\caption{\textbf{Zero-shot cross-modal retrieval.}
R@1 with 100 candidates. Bold: best.}
\label{tab:retrieval}
\scriptsize
\setlength{\tabcolsep}{1.5pt}
\begin{tabular*}{\linewidth}{@{\extracolsep{\fill}}lcccccc@{}}
\toprule
\multicolumn{1}{c}{\multirow{2}{*}[-0.5ex]{\textbf{Model}}} & \multicolumn{2}{>{\columncolor[HTML]{EAF2E6}}c}{\textbf{MC-MED}} & \multicolumn{2}{>{\columncolor[HTML]{FDF1E6}}c}{\textbf{MIMIC-III}} & \multicolumn{2}{>{\columncolor[HTML]{F2ECFC}}c}{\textbf{VitalDB500}}\\
\cmidrule(lr){2-3}\cmidrule(lr){4-5}\cmidrule(lr){6-7}
 & S2T$\uparrow$ & T2S$\uparrow$ & S2T$\uparrow$ & T2S$\uparrow$ & S2T$\uparrow$ & T2S$\uparrow$\\
\midrule
GPT-5.6-luna & 0.105 & --- & 0.090 & --- & 0.075 & ---\\
GPT-6-luna & 0.110 & --- & 0.085 & --- & 0.085 & ---\\
GPT-6-sol & 0.240 & --- & 0.260 & --- & 0.200 & ---\\
Opus~5.5 & 0.190 & --- & 0.275 & --- & 0.230 & ---\\
\midrule
Qwen3-4B (SFT) & 0.385 & 0.600 & 0.160 & 0.185 & 0.125 & 0.180\\
\midrule
\noalign{\begingroup\color{ppgrowblue}\hrule height\dimexpr\ht\strutbox+\dp\strutbox\relax\endgroup\vskip-\dimexpr\ht\strutbox+\dp\strutbox\relax}
\textbf{PPG-LM} & \textbf{0.975} & \textbf{0.965} & \textbf{0.805} & \textbf{0.755} & \textbf{0.555} & \textbf{0.550}\\
\bottomrule
\end{tabular*}
\vspace{0.2em}
\end{wraptable}

\textbf{Zero-shot cross-modal retrieval.}
We test whether a PPG signal retrieves its paired description
(signal-to-text, S2T) and vice versa (text-to-signal, T2S).
Table~\ref{tab:retrieval} reports R@1, the fraction of queries whose paired
item ranks first, averaged over two 100-candidate pools.
Following SensorLM and SleepLM \citep{sensorlm,sleeplm}, we evaluate
general-purpose LLMs only in S2T to avoid the long numerical contexts
required to include all 100 candidate waveforms in a T2S prompt.
\ppglm{} leads both directions across all three datasets, with S2T R@1
of 0.975, 0.805, and 0.555 on MC-MED, MIMIC-III, and VitalDB, respectively;
the corresponding T2S scores are 0.965, 0.755, and 0.550.
These results support bidirectional signal--text alignment that transfers
to external datasets, with lower retrieval accuracy outside MC-MED.
\par
\end{samepage}

\subsection{Zero-Shot Event- and Visit-Level Physiological Recognition}
\label{sec:exp-recognition}

\begin{table}[!htbp]
\centering
\footnotesize
\vspace{-0.9em}
\caption{\textbf{Zero-shot event-level physiological recognition.} Class-versus-control AUROC ($\uparrow$)
for the five highest-scoring MC-MED classes for \ppglm{} and all available
VitalDB classes. Bold marks the best score in each column.}
\label{tab:medication-top-classes}
\scriptsize
\setlength{\tabcolsep}{2pt}
\begin{tabular*}{\linewidth}{@{\extracolsep{\fill}}l*{9}{c}@{}}
\toprule
\multicolumn{1}{c}{\multirow{2}{*}[-0.5ex]{\textbf{Model}}} & \multicolumn{5}{>{\columncolor[HTML]{EAF2E6}}c}{\textbf{MC-MED}}
 & \multicolumn{4}{>{\columncolor[HTML]{F2ECFC}}c}{\textbf{VitalDB}}\\
\cmidrule(lr){2-6}\cmidrule(lr){7-10}
 & Rate ctrl.$\uparrow$ & Vasopress.$\uparrow$ & Vasodil.$\uparrow$ & IV fluid$\uparrow$ & Diuretic$\uparrow$
      & Opioid$\uparrow$ & Sedative$\uparrow$ & Vasopress.$\uparrow$ & Vasodil.$\uparrow$\\
\midrule
GPT-5.6-luna & 0.674 & 0.707 & 0.553 & 0.587 & 0.499
             & 0.567 & 0.641 & 0.578 & 0.614\\
GPT-6-luna & 0.626 & 0.715 & 0.648 & 0.530 & 0.478
           & 0.600 & 0.594 & 0.480 & 0.693\\
GPT-6-sol & 0.651 & 0.699 & 0.648 & 0.536 & 0.467
          & 0.663 & 0.679 & 0.480 & 0.585\\
Opus~5.5 & 0.595 & 0.707 & 0.636 & 0.633 & 0.460
         & 0.722 & 0.766 & 0.571 & 0.634\\
\midrule
PulseLM & 0.591 & 0.589 & 0.458 & 0.493 & 0.553
               & 0.591 & 0.613 & 0.602 & 0.480\\
Qwen3-4B (SFT) & 0.324 & 0.464 & 0.507 & 0.531 & 0.453
               & 0.532 & 0.386 & 0.503 & 0.481\\
\midrule
\noalign{\begingroup\color{ppgrowblue}\hrule height\dimexpr\ht\strutbox+\dp\strutbox\relax\endgroup\vskip-\dimexpr\ht\strutbox+\dp\strutbox\relax}
\textbf{PPG-LM} & \textbf{0.860} & \textbf{0.830} & \textbf{0.772}
               & \textbf{0.759} & \textbf{0.754}
               & \textbf{0.731} & \textbf{0.807} & \textbf{0.612} & \textbf{0.741}\\
\bottomrule
\end{tabular*}
\end{table}
\vspace{-0.5em}

\textbf{Zero-shot event-level physiological recognition.}
We distinguish observations around medication administrations from
no-administration controls (Table~\ref{tab:medication-top-classes}). Each column
reports class-versus-control AUROC for a medication class: rate-control agents
(Rate ctrl.), vasopressors (Vasopress.), vasodilators (Vasodil.), intravenous
fluids (IV fluid), diuretics, opioids, or sedatives.
\ppglm{} outperforms the reported baselines on all displayed classes, with
AUROC of 0.754--0.860 for the five selected MC-MED classes and 0.612--0.807
for the four VitalDB classes. These results support recognition of documented
treatment context across emergency and surgical settings.

\begin{table}[!htbp]
\centering\small
\vspace{-0.5em}
\caption{\textbf{Zero-shot visit-level recognition.}
Balanced accuracy ($\uparrow$) for home-medication use (MC-MED) and clinical
status (VitalDB). \ppglm{} uses eight-window fact embeddings with held-out
threshold calibration. Bold marks the best score in each column.}
\label{tab:recognition-facts}
\setlength{\tabcolsep}{2pt}
\begin{tabular*}{\linewidth}{@{\extracolsep{\fill}}l*{8}{c}@{}}
\toprule
\multicolumn{1}{c}{\multirow{2}{*}[-0.5ex]{\textbf{Model}}} & \multicolumn{4}{>{\columncolor[HTML]{EAF2E6}}c}{\textbf{MC-MED}} & \multicolumn{4}{>{\columncolor[HTML]{F2ECFC}}c}{\textbf{VitalDB}}\\
\cmidrule(lr){2-5}\cmidrule(lr){6-9}
 & CCB$\uparrow$ & AntiHTN$\uparrow$ & AntiPL$\uparrow$ & AntiDM$\uparrow$ & HTN$\uparrow$ & DM$\uparrow$ & Anemia$\uparrow$ & eGFR$\uparrow$\\
\midrule
GPT-5.6-luna & 0.468 & 0.509 & 0.477 & 0.520 & 0.500 & 0.465 & 0.487 & 0.500\\
GPT-6-luna & 0.497 & 0.484 & 0.421 & 0.464 & 0.475 & 0.507 & 0.495 & 0.498\\
GPT-6-sol & 0.500 & 0.500 & 0.505 & 0.507 & \textbf{0.523} & 0.502 & 0.523 & 0.500\\
Opus~5.5 & 0.500 & 0.443 & 0.468 & 0.507 & 0.519 & 0.512 & 0.484 & 0.500\\
\midrule
PulseLM & 0.500 & 0.500 & 0.500 & 0.500 & 0.500 & 0.500 & 0.500 & 0.500\\
Qwen3-4B (SFT) & 0.535 & 0.540 & 0.438 & 0.444 & 0.500 & 0.500 & 0.500 & 0.500\\
\midrule
\noalign{\begingroup\color{ppgrowblue}\hrule height\dimexpr\ht\strutbox+\dp\strutbox\relax\endgroup\vskip-\dimexpr\ht\strutbox+\dp\strutbox\relax}
\textbf{PPG-LM} & \textbf{0.558} & \textbf{0.570} & \textbf{0.544} & \textbf{0.532} & 0.520 & \textbf{0.562} & \textbf{0.531} & \textbf{0.559}\\
\bottomrule
\end{tabular*}
\end{table}
\vspace{-0.5em}

\textbf{Zero-shot visit-level recognition.}
We assess whether eight PPG windows identify recorded home-medication use and
clinical status, reporting balanced accuracy in Table~\ref{tab:recognition-facts}.
The MC-MED columns denote calcium-channel blockers (CCB), antihypertensives
(AntiHTN), antiplatelets (AntiPL), and antidiabetic drugs (AntiDM).
VitalDB columns cover hypertension (HTN), diabetes (DM), anemia, and reduced
kidney function (eGFR below 60\,mL/min/1.73\,m$^2$).
With held-out threshold calibration, \ppglm{} leads on seven of eight tasks;
the exception is HTN (0.520 vs.\ 0.523 for GPT-6-sol). Balanced accuracy of
0.520--0.570 indicates modest discrimination of these clinical facts.
Additional MC-MED evaluations cover eight triage attributes and eleven
medical-history conditions in Appendix~\ref{app:experiment-facts},
Table~\ref{tab:recognition-context-extra}(a) and (b), respectively.

\subsection{Clinical Prediction with Frozen Representations}
\label{sec:exp-repr}

\begin{table}[!htbp]
\centering\small
\caption{\textbf{Clinical prediction with linear probes.}
Three classification and three regression tasks per dataset (mean $\pm$ SD).
Bold: best mean; $\dagger$: window-level target.}
\label{tab:repr}
\setlength{\tabcolsep}{1.4pt}
\renewcommand{\arraystretch}{0.96}
\newcommand{\clinicalgroup}[2]{%
  \multicolumn{9}{@{}l@{}}{%
    \begingroup\setlength{\fboxsep}{1.4pt}%
    \colorbox[HTML]{#1}{%
      \makebox[\dimexpr\linewidth*10/9-2\fboxsep\relax][l]{%
        \strut\textbf{\textit{#2}}}}%
    \endgroup}\\}
\scalebox{0.88}{%
\begin{tabular}{lcccccccc}
\toprule
\multicolumn{1}{c}{\multirow{2}{*}[-0.5ex]{\textbf{Task}}} & \multicolumn{6}{c}{Foundation models} & \multicolumn{2}{c}{\ppglm{}}\\
\cmidrule(lr){2-7}\cmidrule(lr){8-9}
 & \shortstack{PaPaGei-S} & \shortstack{PulsePPG} & AnyPPG & \shortstack{SIGMA-PPG} & \shortstack{MOMENT-L} & \shortstack{Chronos-2} & Stage 1 & Stage 2\\
\midrule
\clinicalgroup{EAF2E6}{MC-MED (ED) --- AUROC $\uparrow$}
Sex & 0.692{\fontsize{5.59}{5.59}\selectfont$\pm$0.006} & 0.719{\fontsize{5.59}{5.59}\selectfont$\pm$0.007} & 0.783{\fontsize{5.59}{5.59}\selectfont$\pm$0.008} & 0.712{\fontsize{5.59}{5.59}\selectfont$\pm$0.008} & 0.724{\fontsize{5.59}{5.59}\selectfont$\pm$0.005} & 0.739{\fontsize{5.59}{5.59}\selectfont$\pm$0.008} & 0.813{\fontsize{5.59}{5.59}\selectfont$\pm$0.006} & \textbf{0.856}{\fontsize{5.59}{5.59}\selectfont$\pm$0.005}\\
ED admission & 0.673{\fontsize{5.59}{5.59}\selectfont$\pm$0.011} & 0.681{\fontsize{5.59}{5.59}\selectfont$\pm$0.009} & 0.750{\fontsize{5.59}{5.59}\selectfont$\pm$0.013} & 0.679{\fontsize{5.59}{5.59}\selectfont$\pm$0.011} & 0.700{\fontsize{5.59}{5.59}\selectfont$\pm$0.011} & 0.728{\fontsize{5.59}{5.59}\selectfont$\pm$0.013} & 0.765{\fontsize{5.59}{5.59}\selectfont$\pm$0.012} & \textbf{0.787}{\fontsize{5.59}{5.59}\selectfont$\pm$0.010}\\
Atrial fibrillation & 0.742{\fontsize{5.59}{5.59}\selectfont$\pm$0.012} & 0.752{\fontsize{5.59}{5.59}\selectfont$\pm$0.013} & 0.845{\fontsize{5.59}{5.59}\selectfont$\pm$0.010} & 0.755{\fontsize{5.59}{5.59}\selectfont$\pm$0.013} & 0.768{\fontsize{5.59}{5.59}\selectfont$\pm$0.014} & 0.815{\fontsize{5.59}{5.59}\selectfont$\pm$0.012} & 0.857{\fontsize{5.59}{5.59}\selectfont$\pm$0.010} & \textbf{0.870}{\fontsize{5.59}{5.59}\selectfont$\pm$0.011}\\
\midrule
\clinicalgroup{EAF2E6}{MC-MED (ED) --- $R^2\uparrow$}
Age & 0.395{\fontsize{5.59}{5.59}\selectfont$\pm$0.016} & 0.428{\fontsize{5.59}{5.59}\selectfont$\pm$0.018} & 0.626{\fontsize{5.59}{5.59}\selectfont$\pm$0.020} & 0.389{\fontsize{5.59}{5.59}\selectfont$\pm$0.017} & 0.475{\fontsize{5.59}{5.59}\selectfont$\pm$0.017} & 0.521{\fontsize{5.59}{5.59}\selectfont$\pm$0.016} & 0.691{\fontsize{5.59}{5.59}\selectfont$\pm$0.016} & \textbf{0.782}{\fontsize{5.59}{5.59}\selectfont$\pm$0.014}\\
Hemoglobin & 0.083{\fontsize{5.59}{5.59}\selectfont$\pm$0.011} & 0.098{\fontsize{5.59}{5.59}\selectfont$\pm$0.013} & 0.196{\fontsize{5.59}{5.59}\selectfont$\pm$0.013} & 0.102{\fontsize{5.59}{5.59}\selectfont$\pm$0.012} & 0.123{\fontsize{5.59}{5.59}\selectfont$\pm$0.014} & 0.164{\fontsize{5.59}{5.59}\selectfont$\pm$0.014} & 0.238{\fontsize{5.59}{5.59}\selectfont$\pm$0.014} & \textbf{0.274}{\fontsize{5.59}{5.59}\selectfont$\pm$0.012}\\
Glucose & 0.032{\fontsize{5.59}{5.59}\selectfont$\pm$0.007} & 0.039{\fontsize{5.59}{5.59}\selectfont$\pm$0.007} & 0.065{\fontsize{5.59}{5.59}\selectfont$\pm$0.008} & 0.041{\fontsize{5.59}{5.59}\selectfont$\pm$0.009} & 0.028{\fontsize{5.59}{5.59}\selectfont$\pm$0.007} & 0.049{\fontsize{5.59}{5.59}\selectfont$\pm$0.011} & 0.071{\fontsize{5.59}{5.59}\selectfont$\pm$0.007} & \textbf{0.083}{\fontsize{5.59}{5.59}\selectfont$\pm$0.011}\\
\midrule
\clinicalgroup{FDF1E6}{MIMIC-III (ICU) --- AUROC $\uparrow$}
Sex & 0.651{\fontsize{5.59}{5.59}\selectfont$\pm$0.024} & 0.686{\fontsize{5.59}{5.59}\selectfont$\pm$0.014} & 0.713{\fontsize{5.59}{5.59}\selectfont$\pm$0.018} & 0.672{\fontsize{5.59}{5.59}\selectfont$\pm$0.016} & 0.682{\fontsize{5.59}{5.59}\selectfont$\pm$0.019} & 0.664{\fontsize{5.59}{5.59}\selectfont$\pm$0.020} & 0.730{\fontsize{5.59}{5.59}\selectfont$\pm$0.019} & \textbf{0.733}{\fontsize{5.59}{5.59}\selectfont$\pm$0.018}\\
In-hospital mortality & 0.697{\fontsize{5.59}{5.59}\selectfont$\pm$0.029} & 0.741{\fontsize{5.59}{5.59}\selectfont$\pm$0.034} & 0.756{\fontsize{5.59}{5.59}\selectfont$\pm$0.031} & 0.732{\fontsize{5.59}{5.59}\selectfont$\pm$0.025} & 0.751{\fontsize{5.59}{5.59}\selectfont$\pm$0.025} & 0.772{\fontsize{5.59}{5.59}\selectfont$\pm$0.020} & 0.786{\fontsize{5.59}{5.59}\selectfont$\pm$0.022} & \textbf{0.793}{\fontsize{5.59}{5.59}\selectfont$\pm$0.021}\\
Circulatory diagnosis & 0.666{\fontsize{5.59}{5.59}\selectfont$\pm$0.015} & 0.690{\fontsize{5.59}{5.59}\selectfont$\pm$0.022} & 0.725{\fontsize{5.59}{5.59}\selectfont$\pm$0.017} & 0.658{\fontsize{5.59}{5.59}\selectfont$\pm$0.018} & 0.672{\fontsize{5.59}{5.59}\selectfont$\pm$0.012} & 0.677{\fontsize{5.59}{5.59}\selectfont$\pm$0.017} & 0.720{\fontsize{5.59}{5.59}\selectfont$\pm$0.022} & \textbf{0.726}{\fontsize{5.59}{5.59}\selectfont$\pm$0.023}\\
\midrule
\clinicalgroup{FDF1E6}{MIMIC-III (ICU) --- $R^2\uparrow$}
Age & 0.295{\fontsize{5.59}{5.59}\selectfont$\pm$0.039} & 0.308{\fontsize{5.59}{5.59}\selectfont$\pm$0.031} & 0.468{\fontsize{5.59}{5.59}\selectfont$\pm$0.029} & 0.288{\fontsize{5.59}{5.59}\selectfont$\pm$0.036} & 0.347{\fontsize{5.59}{5.59}\selectfont$\pm$0.036} & 0.367{\fontsize{5.59}{5.59}\selectfont$\pm$0.028} & 0.502{\fontsize{5.59}{5.59}\selectfont$\pm$0.030} & \textbf{0.507}{\fontsize{5.59}{5.59}\selectfont$\pm$0.028}\\
BUN & 0.029{\fontsize{5.59}{5.59}\selectfont$\pm$0.011} & 0.029{\fontsize{5.59}{5.59}\selectfont$\pm$0.009} & 0.065{\fontsize{5.59}{5.59}\selectfont$\pm$0.019} & 0.034{\fontsize{5.59}{5.59}\selectfont$\pm$0.015} & 0.055{\fontsize{5.59}{5.59}\selectfont$\pm$0.015} & 0.074{\fontsize{5.59}{5.59}\selectfont$\pm$0.012} & 0.081{\fontsize{5.59}{5.59}\selectfont$\pm$0.012} & \textbf{0.086}{\fontsize{5.59}{5.59}\selectfont$\pm$0.013}\\
Hemoglobin & 0.035{\fontsize{5.59}{5.59}\selectfont$\pm$0.010} & 0.048{\fontsize{5.59}{5.59}\selectfont$\pm$0.010} & 0.079{\fontsize{5.59}{5.59}\selectfont$\pm$0.023} & 0.034{\fontsize{5.59}{5.59}\selectfont$\pm$0.009} & 0.046{\fontsize{5.59}{5.59}\selectfont$\pm$0.012} & 0.076{\fontsize{5.59}{5.59}\selectfont$\pm$0.018} & 0.092{\fontsize{5.59}{5.59}\selectfont$\pm$0.024} & \textbf{0.095}{\fontsize{5.59}{5.59}\selectfont$\pm$0.020}\\
\midrule
\clinicalgroup{F2ECFC}{VitalDB (OR) --- AUROC $\uparrow$}
ASA $\geq 3$ & 0.606{\fontsize{5.59}{5.59}\selectfont$\pm$0.021} & 0.642{\fontsize{5.59}{5.59}\selectfont$\pm$0.020} & 0.690{\fontsize{5.59}{5.59}\selectfont$\pm$0.026} & 0.605{\fontsize{5.59}{5.59}\selectfont$\pm$0.035} & 0.625{\fontsize{5.59}{5.59}\selectfont$\pm$0.035} & 0.655{\fontsize{5.59}{5.59}\selectfont$\pm$0.040} & 0.699{\fontsize{5.59}{5.59}\selectfont$\pm$0.034} & \textbf{0.704}{\fontsize{5.59}{5.59}\selectfont$\pm$0.025}\\
Hypertension & 0.598{\fontsize{5.59}{5.59}\selectfont$\pm$0.016} & 0.607{\fontsize{5.59}{5.59}\selectfont$\pm$0.020} & 0.635{\fontsize{5.59}{5.59}\selectfont$\pm$0.019} & 0.595{\fontsize{5.59}{5.59}\selectfont$\pm$0.025} & 0.610{\fontsize{5.59}{5.59}\selectfont$\pm$0.016} & 0.617{\fontsize{5.59}{5.59}\selectfont$\pm$0.023} & \textbf{0.656}{\fontsize{5.59}{5.59}\selectfont$\pm$0.016} & 0.655{\fontsize{5.59}{5.59}\selectfont$\pm$0.017}\\
ICU admission & 0.690{\fontsize{5.59}{5.59}\selectfont$\pm$0.009} & 0.700{\fontsize{5.59}{5.59}\selectfont$\pm$0.004} & 0.747{\fontsize{5.59}{5.59}\selectfont$\pm$0.002} & 0.683{\fontsize{5.59}{5.59}\selectfont$\pm$0.009} & 0.662{\fontsize{5.59}{5.59}\selectfont$\pm$0.011} & 0.714{\fontsize{5.59}{5.59}\selectfont$\pm$0.009} & 0.733{\fontsize{5.59}{5.59}\selectfont$\pm$0.004} & \textbf{0.751}{\fontsize{5.59}{5.59}\selectfont$\pm$0.006}\\
\midrule
\clinicalgroup{F2ECFC}{VitalDB (OR) --- $R^2\uparrow$}
Age & 0.240{\fontsize{5.59}{5.59}\selectfont$\pm$0.026} & 0.263{\fontsize{5.59}{5.59}\selectfont$\pm$0.036} & 0.403{\fontsize{5.59}{5.59}\selectfont$\pm$0.024} & 0.259{\fontsize{5.59}{5.59}\selectfont$\pm$0.026} & 0.302{\fontsize{5.59}{5.59}\selectfont$\pm$0.025} & 0.337{\fontsize{5.59}{5.59}\selectfont$\pm$0.032} & 0.418{\fontsize{5.59}{5.59}\selectfont$\pm$0.030} & \textbf{0.435}{\fontsize{5.59}{5.59}\selectfont$\pm$0.029}\\
EtCO$_2$$^\dagger$ & 0.098{\fontsize{5.59}{5.59}\selectfont$\pm$0.023} & 0.121{\fontsize{5.59}{5.59}\selectfont$\pm$0.034} & 0.105{\fontsize{5.59}{5.59}\selectfont$\pm$0.028} & 0.113{\fontsize{5.59}{5.59}\selectfont$\pm$0.024} & 0.133{\fontsize{5.59}{5.59}\selectfont$\pm$0.029} & 0.150{\fontsize{5.59}{5.59}\selectfont$\pm$0.031} & \textbf{0.153}{\fontsize{5.59}{5.59}\selectfont$\pm$0.038} & 0.152{\fontsize{5.59}{5.59}\selectfont$\pm$0.037}\\
BIS$^\dagger$ & 0.380{\fontsize{5.59}{5.59}\selectfont$\pm$0.019} & 0.398{\fontsize{5.59}{5.59}\selectfont$\pm$0.022} & 0.393{\fontsize{5.59}{5.59}\selectfont$\pm$0.021} & 0.387{\fontsize{5.59}{5.59}\selectfont$\pm$0.020} & 0.402{\fontsize{5.59}{5.59}\selectfont$\pm$0.021} & 0.404{\fontsize{5.59}{5.59}\selectfont$\pm$0.024} & 0.407{\fontsize{5.59}{5.59}\selectfont$\pm$0.022} & \textbf{0.408}{\fontsize{5.59}{5.59}\selectfont$\pm$0.022}\\
\bottomrule
\end{tabular}}
\end{table}

We evaluate frozen PPG representations on 32 dataset--task pairs across MC-MED, MIMIC-III, and VitalDB using logistic and ridge regression with patient-disjoint splits. All models use the same four windows per visit: baselines and Stage~1 mean-pool features, while Stage~2 uses its learned aggregator. Stage~2 \ppglm{} exceeds all six foundation-model baselines in mean performance on all 32 tasks and Stage~1 on 30, except VitalDB hypertension and EtCO$_2$, although glucose prediction remains weak ($R^2=0.083$). Table~\ref{tab:repr} presents selected results; Appendix~\ref{app:experiment-clinical-results}, Table~\ref{tab:repr-full} provides complete results and protocols.

\subsection{Model Analysis}
\label{sec:exp-analysis}

\begin{wraptable}[9]{r}{0.6\linewidth}
\vspace{-\intextsep}
\centering
\setlength{\abovecaptionskip}{0pt}
\caption{\textbf{Ablation studies.}}
\label{tab:ablation}
\footnotesize
\setlength{\tabcolsep}{2pt}
\renewcommand{\arraystretch}{0.94}
\begin{tabular*}{\linewidth}{@{\extracolsep{\fill}}lccc@{}}
\toprule
\textbf{Variant}
& \multicolumn{1}{>{\columncolor[HTML]{EAF2E6}}c}{\shortstack{Temporal\\acc.$\uparrow$}}
& \multicolumn{1}{>{\columncolor[HTML]{FDF1E6}}c}{\shortstack{Event\\AUROC$\uparrow$}}
& \multicolumn{1}{>{\columncolor[HTML]{F2ECFC}}c}{\shortstack{Segment\\R@1$\uparrow$}}\\
\midrule
w/o temporal loss & 0.511 & 0.698 & 0.446\\
w/o signed offsets & 0.923 & 0.617 & 0.445\\
w/o segment loss & 0.922 & 0.697 & 0.202\\
w/o distillation & 0.920 & 0.700 & 0.445\\
w/o seg. loss + distill. & 0.919 & 0.699 & 0.015\\
\midrule
\noalign{\begingroup\color{ppgrowblue}\hrule height\dimexpr\ht\strutbox+\dp\strutbox\relax\endgroup\vskip-\dimexpr\ht\strutbox+\dp\strutbox\relax}
\textbf{PPG-LM} & \textbf{0.925} & \textbf{0.707} & \textbf{0.447}\\
\bottomrule
\end{tabular*}
\vspace{\intextsep}
\end{wraptable}

\textbf{Ablation studies.}
Table~\ref{tab:ablation} shows that temporal matching and event
offsets support temporal and event discrimination, respectively.
Continued segment supervision preserves retrieval, while distillation helps
mainly when it is removed.
Comparing CLIP and SigLIP, CoCa gives the best Stage-2 rhythm, 
quality recognition and retrieval
(Appendix~\ref{app:objective-ablations}).

\begin{figure}[!htb]
\centering
\includegraphics[width=\linewidth]{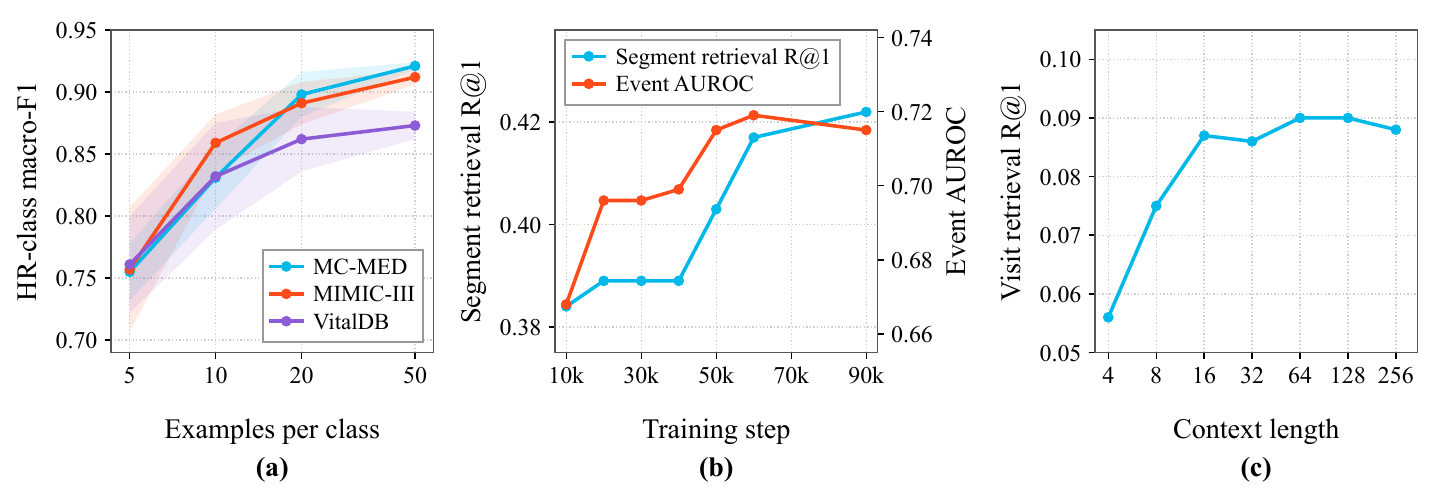}
\vspace{-2em}
\caption{\textbf{Model analysis.}
(a) HR-class macro-F1 with labeled examples per class; shading denotes
$\pm1$ SD over five support draws. VitalDB uses the 40\,Hz waveform view.
(b) Segment retrieval and medication-event AUROC during training.
(c) Visit retrieval versus context length in windows.}
\vspace{-1em}
\label{fig:adaptation-analysis}
\end{figure}

\textbf{Few-shot adaptation.}
Figure~\ref{fig:adaptation-analysis}(a) shows HR-class prediction with linear probes on
frozen \ppglm{} representations. Mean macro-F1 increases consistently as
labeled examples per class grow from 5 to 50 on MC-MED, MIMIC-III, and VitalDB.
Appendix~\ref{app:experiment-fewshot} reports complete results across tasks
and label budgets, along with the separate LLM in-context protocol, whose
label budget and adaptation procedure differ from those of the probes.

\textbf{Training dynamics.}
Figure~\ref{fig:adaptation-analysis}(b) tracks an extended Stage-2 CoCa run.
Segment retrieval improves with training, while medication-event AUROC
largely plateaus by 50--60k steps, indicating different benefits from
additional optimization across the two tasks.

\textbf{Visit context.}
With the Stage-2 CoCa model fixed, increasing context from four to sixteen
windows improves visit retrieval, with little further gain from longer
contexts (Figure~\ref{fig:adaptation-analysis}(c)). This suggests diminishing returns
from additional observations in this retrieval setting.

%% file: Sections/5_conclusion.tex
\section{Conclusion}
\label{sec:conclusion}

We presented \ppglm{}, a PPG--language model family that connects waveforms with clinical context.
Our framework combines signal- and EHR-derived captions with two-stage alignment across segment, event, and visit levels.
Experiments on MC-MED, MIMIC-III, and VitalDB support its effectiveness in language-based recognition, retrieval, captioning, and clinical prediction.

%% file: Sections/6_appendix.tex
\clearpage
\begingroup
\raggedbottom

\section{Limitations}
\label{app:limitations}

We pretrain \ppglm{} on single-center adult emergency-department data.
Although MIMIC-III and VitalDB provide external validation, broader
multi-center data would strengthen generalization across clinical settings.
Clinical supervision depends on available EHR information, whose
completeness and temporal accuracy affect caption quality and may limit
model understanding and generalization. Richer longitudinal records and
clinician-verified descriptions could improve this supervision.
Medication-event recognition captures associations rather than causal
drug effects, since treatment indication and concurrent interventions may
confound observed signals. The current model is intended for research
and is not ready for direct clinical decision-making; prospective
validation, clinician assessment, subgroup analyses, and uncertainty
calibration are needed before clinical deployment.

\section{Caption Construction Details}
\label{app:captions}

We summarize how waveform measurements and clinical records produce segment,
visit, event, and temporal captions. The tables describe their content;
Appendix~\ref{app:prompts} provides language templates and rendered examples.

\makeatletter
\newenvironment{appcaptiontable}{%
  \par\addvspace{4pt}\noindent\begin{minipage}{\linewidth}%
  \def\@captype{table}%
  \centering\small
  \setlength{\tabcolsep}{5pt}%
  \renewcommand{\arraystretch}{1.08}%
}{\end{minipage}\par\addvspace{4pt}}
\makeatother

\subsection{Segment captions and signal quality}
\label{app:caption-signal}

\textbf{Measurements.}
Each segment contains 30 seconds of PPG sampled at 125\,Hz. NeuroKit2 provides
pulse timing, amplitude, and quality measurements; pyPPG provides augmentation
and signal-quality indices \citep{neurokit2,pyppg}.
Synchronized ECG supplies rhythm annotations and supports quality checks;
the model input remains PPG.

\begin{appcaptiontable}
\caption{\textbf{Segment-caption content.} Perfusion and morphology are
waveform surrogates rather than independent clinical diagnoses.}
\label{tab:caption-signal-rules}
\begin{tabular*}{\linewidth}{>{\raggedright\arraybackslash}p{\dimexpr0.25\linewidth-10pt\relax}>{\raggedright\arraybackslash}p{\dimexpr0.75\linewidth-10pt\relax}}
\toprule
\rowcolor[HTML]{ECF4FF}
\textbf{Attribute} & \textbf{Evidence and description}\\
\midrule
Pulse rate & Median pulse-interval rate, expressed as a value and a category.\\[2pt]
Rate trend & Instantaneous rate slope: falling, stable, or rising.\\[2pt]
Rhythm & ECG R--R interval variability, expressed as a regularity category.\\[2pt]
Perfusion & Pulsatile amplitude relative to baseline: weak, moderate, or strong.\\[2pt]
Signal quality & NeuroKit2 quality score: clean, fair, or noisy but usable.\\[2pt]
Augmentation & The pyPPG augmentation index: low, moderate, or high.\\[2pt]
SVRI & Post-peak/pre-peak mean-amplitude ratio in normalized beats: low, mid-range, or high.\\[2pt]
Amplitude variability & Beat-amplitude variation: steady to highly variable.\\
\bottomrule
\end{tabular*}
\end{appcaptiontable}

\textbf{Quality filtering.}
Usability combines PPG quality with available signal-quality, flatline, and
clipping checks; missing optional checks do not reject a window. Good windows
meet stricter criteria, including PPG--ECG rate agreement when reliable ECG
exists; other usable windows are fair. Selection tiers differ from caption
quality labels. Unavailable or unreliable ECG rhythm annotations are omitted.

\textbf{Rendering.}
Missing features are omitted. Poor windows receive quality-only descriptions
and are excluded from training but retained for validation and testing.
Each feature has ten templates; template selection and sentence shuffling
use reproducible example-specific seeds.

\textbf{Signal-plus-vitals view.}
For the signal-plus-vitals configuration reported in
Section~\ref{sec:experiments}, available charted HR, SpO$_2$, RR, and BP
measurements augment the eight signal attributes described in
Section~\ref{sec:captions}. Each value is matched to the
nearest charted observation within 300 seconds of the window timestamp;
unavailable values are omitted. These measurements enter the text target,
while the waveform input remains PPG.

\subsection{Visit captions and clinical context}
\label{app:caption-clinical}

\textbf{Clinical records.}
Visit captions summarize encounter context
(Table~\ref{tab:caption-visit-content}). History entries are deduplicated and
ranked by clinical relevance, then recency.

\begin{appcaptiontable}
\caption{\textbf{Visit-caption content.}}
\label{tab:caption-visit-content}
\begin{tabular*}{\linewidth}{>{\raggedright\arraybackslash}p{\dimexpr0.25\linewidth-10pt\relax}>{\raggedright\arraybackslash}p{\dimexpr0.75\linewidth-10pt\relax}}
\toprule
\rowcolor[HTML]{F2ECFC}
\textbf{Component} & \textbf{Evidence and description}\\
\midrule
Demographics & Recorded age and sex.\\[2pt]
Presentation & Arrival mode, Emergency Severity Index, and normalized chief complaints.\\[2pt]
Triage measurements & Available heart rate, respiratory rate, oxygen saturation, blood pressure, and temperature.\\[2pt]
Clinical history & Selected prior ICD-10 diagnoses, expressed as clinical descriptions.\\[2pt]
Home medications & Recorded active home-medication classes.\\
\bottomrule
\end{tabular*}
\end{appcaptiontable}

\textbf{Field selection.}
Known post-arrival diagnoses are excluded; history with missing diagnosis or
arrival dates is retained. Home medications use patient-level active records
without encounter-specific temporal verification. These fields therefore have
incomplete evidence of availability time. Disposition, discharge diagnoses,
and length of stay are excluded.

\subsection{Event captions and within-visit controls}
\label{app:caption-events}

\textbf{Administration records.}
Eligible records require a parseable first-administration timestamp and a
normalized drug name. Route and class filters exclude topical, ophthalmic,
and other ineligible administrations. Name rules and class-keyword fallback
map the remaining records to opioids, intravenous fluids, rate-control agents,
vasodilators, bronchodilators, sedatives, non-opioid analgesics/antipyretics,
vasopressors, diuretics, or antiemetics. Unmapped records do not become targets;
near-duplicate administrations are removed within each visit and class.

\begin{appcaptiontable}
\caption{\textbf{Event-caption construction.}}
\label{tab:caption-event-content}
\begin{tabular*}{\linewidth}{>{\raggedright\arraybackslash}p{\dimexpr0.25\linewidth-10pt\relax}>{\raggedright\arraybackslash}p{\dimexpr0.75\linewidth-10pt\relax}}
\toprule
\rowcolor[HTML]{FDF1E6}
\textbf{Component} & \textbf{Evidence and description}\\
\midrule
Event anchor & A recorded administration timestamp for an eligible medication class.\\[2pt]
Pre-event context & The latest available PPG windows preceding the administration.\\[2pt]
Post-event context & PPG windows sampled evenly across the period following administration.\\[2pt]
Caption target & The medication class, with individual drug names, doses, and exact latencies omitted.\\[2pt]
Control view & The same sampling rule around a within-visit anchor free of nearby eligible administrations.\\
\bottomrule
\end{tabular*}
\end{appcaptiontable}

\textbf{Sampling and controls.}
Event views use up to three windows in the preceding 30 minutes and up to five in
the following 60 minutes, with at least one pre-administration and two
post-administration windows. Control anchors are sampled uniformly over the
same visit's monitored span; their exclusion interval also covers eligible
but unmapped drugs. Controls are capped at the retained event count, subject
to availability. This construction avoids a fixed window-start cue but does
not establish causal effects or remove all temporal confounding. Antiemetics
serve as an additional comparison class.

\subsection{Temporal descriptions and corrupted statements}
\label{app:caption-temporal}

\textbf{Trajectory evidence.}
Ordered segment-caption rates and rhythm labels produce the temporal
descriptions in Table~\ref{tab:caption-temporal-content}. They summarize
observed tachycardia, irregularity, and changes in pulse rate across a visit.

\begin{appcaptiontable}
\caption{\textbf{Temporal-caption content.}}
\label{tab:caption-temporal-content}
\begin{tabular*}{\linewidth}{>{\raggedright\arraybackslash}p{\dimexpr0.25\linewidth-10pt\relax}>{\raggedright\arraybackslash}p{\dimexpr0.75\linewidth-10pt\relax}}
\toprule
\rowcolor[HTML]{EAF5EF}
\textbf{Operator} & \textbf{Evidence and description}\\
\midrule
Occurrence & Whether tachycardia or irregularity appears among known observations.\\[2pt]
Proportion & The affected share of known windows, described as brief, part, or most.\\[2pt]
Contiguous duration & Consecutive affected windows, grouped into short, intermediate, or long runs.\\[2pt]
Rate ordering & Consecutive groups of observed rates, describing rising, falling, stable, rise-then-fall, or fall-then-rise patterns.\\
\bottomrule
\end{tabular*}
\end{appcaptiontable}

\textbf{Missing observations.}
Proportion and duration descriptions require the state to occur. Unknown
states and monitoring gaps interrupt runs; unsupported operators and ambiguous
rate profiles are omitted.

\textbf{Corrupted descriptions.}
Each supported operator value has three templates. A negative substitutes an
incompatible value in one sentence and re-renders that sentence, retaining
the remaining facts. Padded sentences and unavailable negatives are masked.

\par
\clearpage
\endgroup

\section{Method and Training Details}
\label{app:training-details}

This appendix expands the architecture and alignment objectives in
Section~\ref{sec:method}, using the same notation throughout.
The waveform feature is $g_i$, the level-conditioned text embedding is
$u^\ell(c)$, and $\operatorname{norm}$ denotes $\ell_2$ normalization.
We first specify the two encoders and their CoCa training interface, then
describe event and visit aggregation, preservation of segment alignment,
and the two-stage optimization schedule.

\subsection{Waveform and text encoders}
\label{app:training-architecture}

\textbf{Waveform encoder.}
The encoder $f_\theta$ operates on a single-channel, 30-second PPG window
$x_i\in\mathbb R^{3750}$ sampled at 125\,Hz. Each window is standardized
using its own mean and standard deviation, with $10^{-6}$ added to the
denominator. A one-dimensional convolution with kernel and stride 25
produces 150 non-overlapping patches, each spanning 0.2 seconds. Learned
absolute positional embeddings are added to the width-384 patch embeddings
before six Transformer blocks \citep{transformer}, each with six attention heads and an MLP
hidden width of $4\times384$. Denoting the final normalized patch states by
$H_i\in\mathbb R^{150\times384}$, the two waveform representations are
\begin{equation}
g_i=\operatorname{Pool}_{1}(H_i)\in\mathbb R^{384},
\qquad
z_i^{\mathrm{seg}}=\operatorname{norm}(W_{\mathrm{seg}}g_i)
\in\mathbb R^{512}.
\label{eq:app-waveform-readout}
\end{equation}
Here $\operatorname{Pool}_{1}$ is a single learned-query attention pooler
with a residual MLP, and $W_{\mathrm{seg}}\in\mathbb R^{512\times384}$
is bias-free. The waveform encoder, including this projection, has
approximately 12.7M parameters. As in the clinical prediction experiments,
the 768-dimensional window probe feature concatenates $g_i$ with the mean
of the 150 patch states, before the alignment projection.

\textbf{Text encoder.}
The shared text tower $\psi_\phi$ is an eight-layer causal Transformer
with width 512, eight attention heads, and an MLP hidden width of
$4\times512$. It uses CLIP's byte-pair encoding vocabulary of 49,408 tokens
\citep{clip}, learned token and absolute positional embeddings, and a
learned prefix $e_\ell$ for
$\ell\in\{\mathrm{seg},\mathrm{evt},\mathrm{vis}\}$.
The prefix is prepended to the text sequence and participates in causal
self-attention. The level-conditioned read-out in Section~\ref{sec:method} is
\begin{equation}
u^\ell(c)=\operatorname{norm}\!\left(P_\ell
\psi_\phi([e_\ell;c])_{\mathrm{EOS}}\right),
\qquad P_\ell\in\mathbb R^{512\times512}.
\label{eq:app-text-readout}
\end{equation}
The end-of-sequence state summarizes the available text, and the
level-specific projection selects its alignment space. The same tower
encodes full captions and individual temporal statements; the latter use
the visit prefix and projection. Both waveform and text towers are trained
from random initialization. CLIP supplies the tokenizer and contrastive
formulation, rather than pretrained encoder weights.

\subsection{CoCa alignment and captioning objectives}
\label{app:training-objectives}

\textbf{Contrastive branch.}
CoCa combines paired representation learning with conditional caption
generation \citep{coca}. In our waveform adaptation, the contrastive branch
connects the single-query waveform read-out in
Equation~\ref{eq:app-waveform-readout} to the text-only end-of-sequence
read-out in Equation~\ref{eq:app-text-readout}. For a batch of $B$ matched
window--caption pairs, the normalized embeddings give
\begin{equation}
\begin{aligned}
Q_{ij}&=s_{\mathrm{seg}}(z_i^{\mathrm{seg}})^\top u^{\mathrm{seg}}(c_j),
\qquad
\mathcal C_i^{\mathrm{seg}}=\{i\}\cup\{j:\kappa_j\ne\kappa_i\},\\
p^{\mathrm{seg}\rightarrow\mathrm{text}}_{ij}
&=\frac{\exp Q_{ij}}{\sum_{k\in\mathcal C_i^{\mathrm{seg}}}\exp Q_{ik}},
\qquad
p^{\mathrm{text}\rightarrow\mathrm{seg}}_{ij}
=\frac{\exp Q_{ji}}{\sum_{k\in\mathcal C_i^{\mathrm{seg}}}\exp Q_{ki}},
\quad j\in\mathcal C_i^{\mathrm{seg}}.
\end{aligned}
\label{eq:app-pair-probabilities}
\end{equation}
The scale $s_{\mathrm{seg}}>0$ is learned. The content key $\kappa_i$
represents the binned measurements underlying a caption, independently of
template wording and sentence order. Off-diagonal pairs with equal keys
are masked in both directions, while each matched pair is retained.
Symmetric InfoNCE
\citep{clip} minimizes the negative log-probability of the diagonal match
in both directions:
\begin{equation}
\mathcal L_{\mathrm{NCE}}(Q;\mathcal C^{\mathrm{seg}})
=-\frac{1}{2B}\sum_{i=1}^{B}
\left(\log p^{\mathrm{seg}\rightarrow\mathrm{text}}_{ii}
+\log p^{\mathrm{text}\rightarrow\mathrm{seg}}_{ii}\right).
\label{eq:app-segment-nce}
\end{equation}
This is Equation~\ref{eq:contrastive} with the segment candidate set from
Section~\ref{sec:method-stage1}. Excluded pairs do not contribute to either
denominator; they are not treated as additional positives.

\textbf{Captioning branch.}
A separate attention pooler with 32 learned queries reads the same patch
states $H_i$ and produces waveform memory
$M_i=\operatorname{Pool}_{32}(H_i)\in\mathbb R^{32\times384}$.
A learned projection maps the memory to width 512. The causal token states
from the eight-layer text tower, with the level-prefix position removed,
then pass through four additional multimodal decoder layers of width 512,
eight heads, and MLP ratio four. Each layer combines causal text
self-attention with cross-attention to the projected waveform memory.
This implements CoCa's separation between text-only contrastive features
and waveform-conditioned caption features \citep{coca}.

Let $d_{i,r}\in\mathbb R^{512}$ be the final layer-normalized decoder state
that predicts token $c_{i,r}$ from the preceding caption tokens, and let
$E_{\mathrm{tok}}\in\mathbb R^{49408\times512}$ be the shared token
embedding matrix. The tied output head and training objective are
\begin{equation}
\begin{aligned}
p_{\phi,\eta}(c_{i,r}\mid c_{i,<r},M_i)
&=\left[\operatorname{softmax}(E_{\mathrm{tok}}d_{i,r})\right]_{c_{i,r}},\\
\mathcal L_{\mathrm{cap}}
&=-\frac{1}{N_{\mathrm{tok}}}\sum_i\sum_{r\in\mathcal T_i}
\log p_{\phi,\eta}(c_{i,r}\mid c_{i,<r},M_i),\\
\mathcal L_{\mathrm{seg}}
&=\mathcal L_{\mathrm{NCE}}(Q;\mathcal C^{\mathrm{seg}})
+\mathcal L_{\mathrm{cap}}.
\end{aligned}
\label{eq:app-coca-objective}
\end{equation}
Here $\mathcal T_i$ contains non-padding target positions, including EOS,
$N_{\mathrm{tok}}=\sum_i|\mathcal T_i|$, and $\eta$ denotes the additional
parameters of the waveform-memory and multimodal decoding branch, as in
Section~\ref{sec:method-stage1}.
Training uses teacher forcing and unit weights for both losses. The two
branches share the waveform patch encoder and causal text tower, so
caption supervision also updates the representations used for alignment.
Caption loss applies to segment descriptions in every batch type; event
and visit descriptions are alignment targets. The CLIP objective compared
in the experiments retains the contrastive branch without caption loss.

\subsection{Event and visit aggregation}
\label{app:training-aggregator}

\textbf{Event read-out.}
For a medication or control anchor $a_j$, $E_j$ contains up to three latest
windows in $[a_j-30,a_j)$ and up to five evenly spaced windows in
$[a_j,a_j+60]$, following Appendix~\ref{app:caption-events}.
Both $t_i$ and $a_j$ use the visit clock, so $t_i-a_j$ is a signed offset
in minutes. Each view has at least one pre-anchor and two post-anchor windows.
The time embedding $\varphi$ maps 64 sinusoidal features, with periods
from two minutes to two weeks, through an MLP to width 384. The attention
operator in Equation~\ref{eq:event} includes layer normalization of the
waveform content and is implemented by
\begin{equation}
\begin{aligned}
U_j^{(0)}&=[q_{\mathrm{evt}};
(\operatorname{LN}(g_i)+\varphi(t_i-a_j))_{i\in E_j}],\\
U_j^{(b)}&=\mathcal B_b^{\mathrm{evt}}(U_j^{(b-1)}),\quad b=1,2,\\
z_j^{\mathrm{evt}}&=\operatorname{norm}\!\left(
W_{\mathrm{evt}}\operatorname{LN}([U_j^{(2)}]_0)\right).
\end{aligned}
\label{eq:app-event-readout}
\end{equation}
Each $\mathcal B_b^{\mathrm{evt}}$ is a self-attention block of width 384,
six heads, and MLP ratio four; key-padding masks exclude missing windows.
Reading the learned-query position yields one 512-dimensional event
embedding. Signed offsets preserve the distinction between observations
before and after the anchor, including for control views.

\textbf{Visit read-out and reference times.}
For a visit $V$, we use the input construction in
Section~\ref{sec:method-stage2}:
\begin{equation}
b_i=g_i+\gamma_{\mathrm{mon}}(t_i)+\gamma_{\mathrm{arr}}(\delta_i),
\qquad
B_V=[(b_i)_{i\in\mathcal I_V};G_V;e_V].
\label{eq:app-visit-inputs}
\end{equation}
Here $\mathcal I_V$ indexes at most 256 selected windows, $G_V$ contains
gap tokens, and $e_V$ is the evidence token. We prepend
eight learned latent queries and update the complete sequence through two
self-attention blocks of width 384, six heads, and MLP ratio four:
\begin{equation}
\begin{aligned}
U_V^{(0)}&=[q_0;\ldots;q_7;B_V],\\
U_V^{(b)}&=\mathcal B_b^{\mathrm{vis}}(U_V^{(b-1)};\tau_V),\quad b=1,2,\\
\ell_{Vk}&=\operatorname{LN}([U_V^{(2)}]_k),\qquad
h_{Vk}=\operatorname{norm}(W_{\mathrm{lat}}\ell_{Vk}),\quad k=0,\ldots,7.
\end{aligned}
\label{eq:latent-updates}
\end{equation}
The blocks use the token times $\tau_V$ and a validity mask; both the
latent and observation positions are updated. Inspired by Perceiver's
learned latent representations \citep{perceiver}, only the eight latent
positions are read out. Attention operates on the full concatenated
sequence, with at most $8+256+16+1=281$ tokens.
The global latent has reference time
$r_0=|\mathcal I_V|^{-1}\sum_{i\in\mathcal I_V}t_i$ and supplies
$z_V^{\mathrm{vis}}=h_{V0}$ for clinical-caption alignment.
The remaining reference times are
$(r_1,\ldots,r_7)=(0,15,30,60,120,240,480)$ minutes after the first window.
Following the reference-time idea in mTAN \citep{mtan}, these anchors give
the queries temporal identities while allowing attention to all valid
observations.

\textbf{Continuous-time rotary encoding.}
We adapt rotary position embeddings from RoFormer \citep{roformer} to
elapsed minutes. In each 64-dimensional attention head, 16 channel pairs
are rotated and the other 32 channels are unchanged. With
$T_p=2(20160/2)^{(p-1)/15}$ minutes and $\theta_p=2\pi/T_p$, define
\begin{equation}
R_\Theta(t)=\operatorname{diag}_{p=1}^{16}
\begin{pmatrix}
\cos(\theta_pt)&-\sin(\theta_pt)\\
\sin(\theta_pt)&\cos(\theta_pt)
\end{pmatrix}.
\label{eq:rotary-blocks}
\end{equation}
For query and key times $\tau$ and $\tau'$, respectively,
\begin{equation}
\begin{aligned}
\widetilde q(\tau)&=[R_\Theta(\tau)q^{\mathrm{rot}};q^{\mathrm{free}}],
&\widetilde k(\tau')&=[R_\Theta(\tau')k^{\mathrm{rot}};k^{\mathrm{free}}],\\
\widetilde q(\tau)^\top\widetilde k(\tau')
&=(q^{\mathrm{rot}})^\top R_\Theta(\tau'-\tau)k^{\mathrm{rot}}
+(q^{\mathrm{free}})^\top k^{\mathrm{free}}.
\end{aligned}
\label{eq:app-rope-relative}
\end{equation}
Setting $\tau=r_k$ and $\tau'=t_i$ recovers
Equation~\ref{eq:time-rope}. Each head divides these scores by
$\sqrt{64}$, applies the validity mask, and normalizes with softmax.
Using actual minutes distinguishes irregular intervals that share the
same window-index separation. The relative-rotation identity holds for
fixed query/key content; additive clock features also affect that content.

\textbf{Time content and monitoring coverage.}
As in Equation~\ref{eq:app-visit-inputs}, learned MLPs map 64-dimensional
sinusoidal features of monitoring time $t_i$ and time since ED arrival
$\delta_i$ into additive width-384 features. A learned vector represents
unavailable arrival time. These features expose time through the values
as well as the attention scores. Gaps longer than ten minutes are detected
over the full visit before context subsampling; the longest 16 are retained.
Each gap token uses its midpoint as its time and encodes the logarithmically
transformed duration, $\log(1+\mathrm{duration})$.
The evidence token maps six coverage descriptors through a
$6\rightarrow384\rightarrow384$ MLP: window count, coverage fraction,
monitoring span, fraction of good-quality windows, arrival-to-monitoring
delay, and an arrival-time availability indicator. Window count,
monitoring span, and available arrival delay use $\log(1+\cdot)$
transforms, with times measured in minutes. Coverage is
$\min(1,0.5n/\max(\mathrm{span},0.5))$ for $n$ observed 30-second windows,
with a half-minute lower bound on the denominator for a single-window visit.
These inputs describe the amount and timing of observed evidence without
imputing physiology within monitoring gaps.

\textbf{Alignment across scopes.}
Event alignment uses Equation~\ref{eq:contrastive} with
$Q_{ij}=s_{\mathrm{evt}}(z_i^{\mathrm{evt}})^\top u^{\mathrm{evt}}(c_j)$
and $\mathcal C_i^{\mathrm{evt}}=\{i\}\cup
\{j:\operatorname{class}(j)\ne\operatorname{class}(i)\}$;
all control views share one category. Global visit alignment uses
$Q_{ij}=s_{\mathrm{vis}}(z_i^{\mathrm{vis}})^\top u^{\mathrm{vis}}(c_j)$
with the full in-batch candidate set.
Temporal matching uses all eight $h_{Vk}$ and the mean-over-statements,
max-over-latents score $R_{ij}$ in Equation~\ref{eq:temporal-score},
adapting ColBERT's late interaction \citep{colbert} to whole statements.
Each visit supplies at most eight statements, with no fixed
statement-to-latent assignment. Equation~\ref{eq:temporal-objective}
combines in-batch alignment of $s_{\mathrm{vis}}R$ with a corruption
ranking term of weight 0.5. For the temporal contrastive term, we retain
the matched sub-batch of visits with $m_V>0$ and use all candidates within
that sub-batch; the term is zero if no visit has available statements.
Padded statements do not enter the mean defining $R_{ij}$.
The ranking term averages only over visits with an available corruption;
if none are available, this term is zero. A visit without temporal
statements still participates in clinical-caption alignment.
Appendix~\ref{app:caption-temporal} specifies the corruptions.
No cross-batch negative queue is used at any scope.

\subsection{Preserving segment alignment}
\label{app:training-distillation}

Stage~2 introduces event and visit targets whose temporal scope differs
from that of a segment caption. Updating the shared encoder solely for
these broader targets can change the local waveform--language geometry
learned in Stage~1. We therefore retain the segment objective on every
batch and use a frozen Stage-1 waveform encoder as a reference, alongside
the multilevel training strategy motivated by HierVL \citep{hiervl}.
The ongoing segment objective continues to learn from paired captions;
the reference penalty limits changes to the normalized waveform embeddings.

Let $\mathcal Q$ be the windows re-encoded in the current batch. A frozen
copy of the Stage-1 waveform encoder and segment projection supplies
$z_i^{(1)}=\operatorname{norm}(W_{\mathrm{seg}}^{(1)}g_i^{(1)})$.
With no gradients through this reference branch, we use
\begin{equation}
\begin{aligned}
\mathcal L_{\mathrm{dst}}(\mathcal Q)
&=\frac{1}{|\mathcal Q|}\sum_{i\in\mathcal Q}
\left(1-(z_i^{\mathrm{seg}})^\top z_i^{(1)}\right)\\
&=\frac{1}{2|\mathcal Q|}\sum_{i\in\mathcal Q}
\left\|z_i^{\mathrm{seg}}-z_i^{(1)}\right\|_2^2.
\end{aligned}
\label{eq:anchor}
\end{equation}
The equality follows from unit normalization. The penalty constrains
embedding directions while allowing the pre-projection waveform features
to adapt. The teacher remains in evaluation mode with dropout disabled;
the current waveform encoder and projection receive gradients after the
initial freezing period in Appendix~\ref{app:training-schedule}.

The alignment terms in Equation~\ref{eq:joint} can then be written as
\begin{equation}
\begin{aligned}
\mathcal L^{(\mathrm{seg})}
&=\mathcal L_{\mathrm{seg}}(\mathcal Q)
+\lambda_{\mathrm{dst}}\mathcal L_{\mathrm{dst}}(\mathcal Q),\\
\mathcal L^{(\mathrm{evt})}
&=\mathcal L_{\mathrm{seg}}(\mathcal Q)
+\lambda_{\mathrm{dst}}\mathcal L_{\mathrm{dst}}(\mathcal Q)
+\mathcal L_{\mathrm{evt}},\\
\mathcal L^{(\mathrm{vis})}
&=\mathcal L_{\mathrm{seg}}(\mathcal Q)
+\lambda_{\mathrm{dst}}\mathcal L_{\mathrm{dst}}(\mathcal Q)
+\mathcal L_{\mathrm{vis}}+\mathcal L_{\mathrm{tmp}},
\end{aligned}
\label{eq:app-batch-objectives}
\end{equation}
where $\lambda_{\mathrm{dst}}=0.1$.
For segment and event batches, $\mathcal Q$ contains their valid input
windows, each paired with its own segment caption. For visit batches, it
contains up to eight live windows per visit; detached context features
participate in aggregation but incur no segment or distillation loss.
The same content-key mask is applied whenever the segment objective is
evaluated, including on live windows from event and visit batches.

\subsection{Optimization, feature caching, and inference}
\label{app:training-schedule}

\textbf{Two-stage schedule.}
Stage~1 uses the segment objective in Section~\ref{sec:method-stage1}.
Both towers are trained from scratch for six epochs over
8,842,271 training windows, using batches
of 256 and dropping the final incomplete batch. This gives 207,240 updates.
Stage~2 initializes from Stage~1 and trains the shared encoders and
scope-specific read-outs under the freezing schedule below for 30,000
global updates, cycling through segment, visit, and event
batches in a $4{:}2{:}1$ ratio. Batch sizes are 256 windows, 64 visits,
and 64 event views, respectively. The extended Stage-2 run used for
the training-dynamics analysis in Figure~\ref{fig:adaptation-analysis}(b)
is separate from this standard schedule.

\textbf{Optimization.}
Both stages use AdamW with $\beta=(0.9,0.98)$, $\epsilon=10^{-6}$,
peak learning rate $5\times10^{-4}$, and weight decay 0.2.
The learning rate follows a 2,000-update linear warmup and cosine decay
to zero over each stage's own schedule. Weight decay excludes biases,
normalization parameters, token and positional embeddings, learned
queries and prefixes, missing-time embeddings, and logit scales.
During the first 2,000 Stage-2 updates, the waveform and text towers and
the segment logit scale have zero learning rate; subsequently they use
one tenth of the scheduled rate. Aggregators and other read-out parameters
use the full scheduled rate from the start. Global gradient norms are
clipped at 1.0. Dropout is 0.1 in waveform Transformer blocks and the
event and visit aggregators, and zero in the text tower, caption decoder,
and attention poolers. Each level has a learned scale
$s_\ell=\min(\exp(\alpha_\ell),100)$; newly introduced scales are
initialized to $1/0.07$, while Stage~2 retains the learned segment scale.
Token budgets are
256 for segment and visit captions, 64 for event captions, and 48 for
temporal statements.

\textbf{Loss weights.}
All contrastive alignment terms and the segment captioning loss use unit
weights. Stage~2 uses a distillation weight of $0.1$ and a temporal
corruption-ranking weight of $0.5$.

\textbf{Live windows and cached context.}
The Stage-1 waveform encoder initializes an fp16 cache of the
384-dimensional window features $g_i$. Each visit batch selects a context
of at most 256 windows per visit. Up to eight distinct windows per visit
are sampled for live re-encoding, using all windows when fewer than eight
are available; their features replace the corresponding cached entries
within this context, so the live windows count toward the context limit.
For the selected set $\mathcal I_V$ and live subset
$\mathcal Q_V=\mathcal Q\cap\mathcal I_V$, aggregation uses
\begin{equation}
\widetilde g_i=
\begin{cases}
f_\theta(x_i),&i\in\mathcal Q_V,\\
\operatorname{stopgrad}(g_i^{\mathrm{cache}}),
&i\in\mathcal I_V\setminus\mathcal Q_V.
\end{cases}
\label{eq:app-cached-features}
\end{equation}
This lets visit-level losses update the aggregators over a longer context
while limiting waveform-encoder activations to the live subset.
The context cap at global Stage-2 update $s$ is
\begin{equation}
C_{\max}(s)=\left\lfloor 8+248\min\!\left(1,\frac{s}{10{,}000}\right)\right\rfloor.
\label{eq:app-context-curriculum}
\end{equation}
Half of the visit batches use this cap; the other half draw a context budget
log-uniformly between 8 and the current cap, subject to the available
windows. Longer visits are subsampled uniformly over time while retaining
the live windows. Detached live features are written back to the cache
after visit batches, and segment batches also refresh their encoded
windows. Event views are encoded directly from raw windows and do not
update the cache. Each training run uses its own cache.

\textbf{Inference.}
All model parameters are frozen. We encode the windows selected by each
benchmark using the evaluation checkpoint; the training feature cache is
not required. Segment and event tasks use $z^{\mathrm{seg}}$ and
$z^{\mathrm{evt}}$, respectively, while clinical-caption visit retrieval
uses the global latent $z_V^{\mathrm{vis}}=h_{V0}$. Retrieval and class-prompt
scoring compare normalized waveform and corresponding level-conditioned
text embeddings by cosine similarity; task-specific prompt sets and
read-outs are specified in Appendix~\ref{app:experiment-details}.
Temporal descriptions use all eight latents and the score in
Equation~\ref{eq:temporal-score}. Segment captioning starts with a
beginning-of-sequence token and uses greedy autoregressive decoding
conditioned on $M_i$, stopping at EOS or the segment token limit.

\textbf{Compute.}
The main CoCa model is trained with seed 0 and bfloat16 autocast on one
NVIDIA H200 GPU. Recorded training times are approximately 26.3 hours for
Stage~1 and 6.0 hours for Stage~2, excluding cache construction and
evaluation. Caption cross-entropy is computed in chunks of 8,192 target
tokens with gradient checkpointing to reduce memory use, without changing
the token-averaged objective in Equation~\ref{eq:app-coca-objective}.

\section{Additional Evaluation Details}
\label{app:experiment-details}

\subsection{Experimental setup}
\label{app:experiment-setup}

\textbf{Datasets and patient splits.}
The three datasets cover emergency, intensive-care, and intraoperative monitoring.
Only MC-MED is used to pretrain \ppglm{}; MIMIC-III and VitalDB evaluate transfer
to external clinical settings.
\begin{itemize}
\renewcommand{\labelitemi}{$\bullet$}
\setlength{\itemsep}{3pt}
\setlength{\parskip}{0pt}
\item \textbf{MC-MED} \citep{mcmed} pairs PPG from Stanford's adult emergency
department (ED) with structured electronic health records. The processed cohort
contains 36,430 patients and 53,473 visits. The patient-random 80/10/10 split yields
the retained counts in Table~\ref{tab:experiment-mcmed-splits}. PPG is divided
into non-overlapping 30-second windows at 125\,Hz; the training set provides
approximately 73,700 hours. Poor-quality windows are excluded from pretraining,
while the validation and test corpus splits retain them.
\item \textbf{MIMIC-III} \citep{mimic3} links intensive-care bedside waveforms
from Beth Israel Deaconess Medical Center to clinical records. The external
benchmark uses PPG paired with synchronized ECG where available, charted vital
signs, demographics, laboratory measurements, and hospital outcomes. Language
evaluation uses fixed patient-level samples. Clinical probes are fitted on the
target dataset's training patients and evaluated on patient-disjoint test patients.
\item \textbf{VitalDB} \citep{vitaldb} pairs intraoperative biosignals from Seoul
National University Hospital with perioperative clinical information. It provides
preoperative conditions and laboratory measurements, intraoperative monitoring,
medication records, and postoperative ICU outcomes. The language benchmark has
official 40\,Hz and original 500\,Hz waveform views of the same evaluation items.
The main \ppglm{} language results, retrieval analysis, and prompt-sensitivity
analysis use the 500\,Hz source; supplementary caption/read-out and few-shot
analyses use the 40\,Hz view. Baselines retain their recorded input views.
Clinical probes use target-dataset training patients and patient-disjoint testing.
\end{itemize}

\begin{table}[!htbp]
\centering\small
\caption{\textbf{Processed MC-MED splits.} Patients are disjoint across splits;
window counts refer to the retained corpus.}
\label{tab:experiment-mcmed-splits}
\setlength{\tabcolsep}{6pt}
\renewcommand{\arraystretch}{1.08}
\begin{tabular*}{\linewidth}{@{\extracolsep{\fill}}lrrr@{}}
\toprule
Split & Patients & Visits & 30-second windows\\
\midrule
Training & 29,177 & 42,856 & 8,842,271\\
Validation & 3,641 & 5,421 & 1,183,278\\
Test & 3,612 & 5,196 & 1,099,395\\
\midrule
Total & 36,430 & 53,473 & 11,124,944\\
\bottomrule
\end{tabular*}
\end{table}

\textbf{Waveform preparation and references.}
\ppglm{} inputs are resampled to 125\,Hz and standardized within each window
as $(x-\mu)/(\sigma+10^{-6})$. The captioning pipeline supplies reference
descriptions on all three datasets; Appendix~\ref{app:captions} describes its
signal measurements, quality gates, and clinical-record matching. ECG supports
rhythm annotation and the independent HR reference, but is not a model input.

\textbf{Baseline configurations.}
For language tasks, GPT-5.6-luna, GPT-6-luna, GPT-6-sol, and Claude Opus~5.5 receive
each 30-second PPG window mean-pooled to 25\,Hz and standardized, giving 750
numerical values, and return JSON answers. Prompts include the annotation class
definitions without reference measurements. Signal-to-text retrieval shuffles
candidate-caption order independently for each query and requests five preferred
indices. PulseLM uses its Qwen2.5-7B-Instruct variant. Our Qwen3-4B (SFT) baseline
uses a frozen PaPaGei-S encoder and full language-model fine-tuning on 100,000
window and 40,000 visit examples from our captioning corpus; evaluation uses
prompt scoring or generated answers as appropriate to the task.
Clinical prediction compares frozen PaPaGei-S, Pulse-PPG, AnyPPG, SIGMA-PPG,
MOMENT-L, and Chronos-2 features with the same logistic- or ridge-regression
probe protocol (Appendix~\ref{app:experiment-probes}). 
For visit-level clinical
targets, all baselines and both \ppglm{} stages receive the same four windows;
baselines and Stage~1 use mean pooling, while Stage~2 uses learned visit aggregation.
MOMENT-L is the largest
evaluated MOMENT variant; PaPaGei-S is the morphology-aware variant.
Unless otherwise stated, \ppglm{} uses the caption-style template prompt bank.

\textbf{Benchmark sampling and scoring.}
Language evaluations use fixed items and cached model responses. Window
recognition and HR evaluation use 300 windows per dataset, with one window per
patient, stratified by rhythm and HR class and excluding the poor-quality tier.
Captioning uses the first 200 of these windows. Clinical-fact recognition in
Table~\ref{tab:recognition-facts} uses 400 MC-MED visits and 300 VitalDB cases,
with eight windows per case and enriched sampling of rare labels. Its thresholds
are calibrated on held-out MC-MED validation visits or separate VitalDB patients,
without using test labels. The broader visit-inference benchmark instead samples
200 encounters per dataset, requires at least eight windows, and stratifies by
sex, age group, and outcome. Visit benchmarks retain one encounter per patient.
The medication-class results in Table~\ref{tab:medication-top-classes} use 393
administration views and 393 controls on MC-MED and 139 of each on VitalDB;
Appendix~\ref{app:experiment-events} additionally reports the smaller event cohort.
Off-label or missing classification answers count as errors. Numerical MAE uses
items with both a numerical answer and an available reference; the HR comparisons
have 300 valid ECG-referenced items per system and dataset. Dashes denote
unavailable or unreported evaluations.

\textbf{Evaluation protocol.} 
Unless otherwise stated, \ppglm{} denotes the Stage-2 model. All model backbones remain frozen during evaluation. Within each benchmark, comparisons use PPG inputs at the same temporal level and with the same number of windows across models. Language benchmarks use the same evaluation examples and class definitions across models; \ppglm{} performs prompt-based recognition, cross-modal retrieval using cosine similarity, and caption generation with greedy decoding. External zero-shot segment recognition, retrieval, and captioning use no target-dataset fitting. Clinical-fact recognition uses read-out thresholds calibrated on held-out MC-MED visits or separate VitalDB patients. For supervised clinical prediction, we fit logistic regression probes for classification and ridge regression probes for continuous targets on frozen features from training patients, and evaluate on patient-disjoint test sets. 

\makeatletter
\newenvironment{appdefinitionpanel}{%
  \par\addvspace{8pt}\noindent\begin{minipage}{\linewidth}%
  \def\@captype{table}%
  \centering\footnotesize
  \setlength{\tabcolsep}{5pt}%
  \renewcommand{\arraystretch}{1.06}%
}{\end{minipage}\par\addvspace{8pt}}
\makeatother

\raggedbottom
\renewcommand{\floatpagefraction}{0.8}
\renewcommand{\topfraction}{0.95}
\setcounter{topnumber}{3}
\subsection{Clinical task definitions}
\label{app:experiment-clinical-tasks}

We define the physiological attributes, medication and visit facts, and clinical
prediction targets used in the main experiments. Tables~\ref{tab:task-physiology}
and~\ref{tab:task-medication-facts} explain the language-based tasks;
Tables~\ref{tab:task-classification} and~\ref{tab:task-regression} cover all 32
clinical prediction targets (17 MC-MED, six MIMIC-III, and nine VitalDB).
In these tables, M denotes MC-MED, I denotes MIMIC-III, and V denotes VitalDB.
Definitions distinguish the recorded target from its clinical interpretation.

\textbf{Physiological understanding.}
Window-level tasks describe rate, regularity, signal properties, and charted
physiology (Table~\ref{tab:task-physiology}). Their annotation rules are given
in Appendix~\ref{app:caption-signal}. Rhythm regularity is a window attribute,
whereas atrial fibrillation in the clinical probe is a recorded prior diagnosis.
Cross-modal retrieval tests whether a signal retrieves its paired description
and vice versa; R@$k$ measures whether that pair appears in the top $k$ candidates
(Appendix~\ref{app:experiment-retrieval}). It is a correspondence task rather
than a separate disease label.

\begin{appdefinitionpanel}
\caption{\textbf{Physiological task definitions.} Tasks in
Tables~\ref{tab:recognition} and~\ref{tab:caption}, including the HR-class target
used in few-shot evaluation. CV denotes coefficient of variation.}
\label{tab:task-physiology}
\begin{tabular}{@{}>{\raggedright\arraybackslash}p{0.19\linewidth}
>{\raggedright\arraybackslash}p{0.75\linewidth}@{}}
\toprule
\rowcolor{ppgrowblue}
Task & Reference and clinical interpretation\\
\midrule
Rhythm & Regular, slightly irregular, or markedly irregular according to
synchronized ECG R--R interval CV ($<0.05$, $[0.05,0.12)$, or $\geq0.12$).
Measures beat-timing regularity, without assigning a specific arrhythmia.\\[1.5pt]
Quality (Qual.) & Clean, fair, or noisy according to the annotation quality
score and usability gate. Describes waveform reliability and artifact burden.\\[1.5pt]
Perfusion (Perf.) & Weak, moderate, or strong pulsatile amplitude relative to
the signal baseline, using the waveform-derived AC/DC index. A peripheral
perfusion surrogate, not a direct blood-flow measurement.\\[1.5pt]
Morphology (Morph.) & Low, moderate, or high augmentation from the pyPPG
augmentation index. Describes pulse-wave shape; it does not establish a
vascular disease diagnosis.\\[1.5pt]
HR class & Five pulse-rate bins: $<50$, 50--59, 60--100, 101--120, and
$>120$\,bpm, spanning markedly slow to fast rates. These are benchmark
annotation categories.\\[1.5pt]
HR estimation & Heart rate in beats/min, evaluated against synchronized ECG.
This independent reference differs from the PPG-derived pulse rate used
to construct training captions.\\[1.5pt]
RR class / estimation & Low, normal, or elevated respiratory rate, or its
numerical value in breaths/min. Both use charted respiratory rate as the
reference and describe breathing frequency.\\[1.5pt]
SBP / MAP & Systolic blood pressure (peak arterial pressure during a beat)
and mean arterial pressure (pressure averaged over the cardiac cycle),
in mmHg, evaluated against the corresponding charted measurements.\\[1.5pt]
HR trend & Falling, stable, or rising rate within a window, from the slope
of instantaneous pulse rate. Included in caption factuality.\\[1.5pt]
Amplitude variability & Steady, moderately variable, or highly variable
beat amplitudes, from beat-amplitude CV. Included in caption factuality.\\[1.5pt]
Caption factuality (Fact) & Pooled accuracy of stated facts across six shared
fields: HR, HR class, rhythm, HR trend, amplitude variability, and quality.
Numerical HR is correct within $\pm5$\,bpm of the caption reference;
categorical fields require agreement.\\
\bottomrule
\end{tabular}
\end{appdefinitionpanel}

\textbf{Medication events and visit facts.}
Table~\ref{tab:task-medication-facts} defines every medication class and binary
visit fact displayed in the main results. Event AUROC compares each medication
class with no-administration controls. The class describes the recorded
administration, without asserting a causal physiological response. Home-medication
facts instead use patient-level active medication records, whose timing is not
verified for each encounter (Appendix~\ref{app:caption-clinical}); their classes
may overlap. VitalDB facts describe preoperative status. In particular, the
binary eGFR target identifies reduced estimated filtration, without establishing
the duration required for a chronic kidney disease diagnosis.

\begin{appdefinitionpanel}
\caption{\textbf{Medication and visit-fact definitions.} Dataset abbreviations
refer to the tasks displayed in Tables~\ref{tab:medication-top-classes}
and~\ref{tab:recognition-facts}.}
\label{tab:task-medication-facts}
\begin{tabular}{@{}>{\raggedright\arraybackslash}p{0.19\linewidth}
>{\raggedright\arraybackslash}p{0.09\linewidth}
>{\raggedright\arraybackslash}p{0.65\linewidth}@{}}
\toprule
Task & Data & Recorded target and clinical meaning\\
\midrule
\rowcolor{ppgrowblue}
\multicolumn{3}{@{}l}{\textbf{Administration events: medication class versus control}}\\[2pt]
Rate control & M & Administration of a rate-controlling medication, used to
slow heart rate.\\[1.5pt]
Vasopressor & M, V & Administration of an agent used to increase arterial
blood pressure.\\[1.5pt]
Vasodilator & M, V & Administration of a medication that dilates blood vessels.\\[1.5pt]
IV fluid & M & Intravenous fluid administration, representing fluid delivery
to the circulation.\\[1.5pt]
Diuretic & M & Administration of medication that promotes renal salt and
water excretion.\\[1.5pt]
Opioid & V & Administration of an opioid analgesic, used for pain control.\\[1.5pt]
Sedative & V & Administration of medication used to induce sedation or
reduce arousal.\\[1.5pt]
\bottomrule
\end{tabular}
\end{appdefinitionpanel}

\begin{appdefinitionpanel}
{\raggedright\textit{Table~\ref{tab:task-medication-facts} (continued)}\par}\smallskip
\begin{tabular}{@{}>{\raggedright\arraybackslash}p{0.19\linewidth}
>{\raggedright\arraybackslash}p{0.09\linewidth}
>{\raggedright\arraybackslash}p{0.65\linewidth}@{}}
\toprule
Task & Data & Recorded target and clinical meaning\\
\midrule
\rowcolor[HTML]{FDF1E6}
\multicolumn{3}{@{}l}{\textbf{Recorded home-medication use: binary visit facts}}\\[2pt]
CCB & M & Recorded use of calcium-channel blockers, which act on vascular
and/or cardiac muscle.\\[1.5pt]
AntiHTN & M & Recorded use of antihypertensive medication, used to lower
blood pressure.\\[1.5pt]
AntiPL & M & Recorded use of antiplatelet medication, which inhibits
platelet activation or aggregation.\\[1.5pt]
AntiDM & M & Recorded use of glucose-lowering medication for diabetes
management.\\[1.5pt]
\midrule
\rowcolor[HTML]{F2ECFC}
\multicolumn{3}{@{}l}{\textbf{Preoperative status: binary visit facts}}\\[2pt]
HTN & V & Preoperative hypertension, a recorded condition of persistently
elevated blood pressure.\\[1.5pt]
DM & V & Preoperative diabetes mellitus, a disorder of glucose regulation.\\[1.5pt]
Anemia & V & The benchmark's preoperative anemia label, indicating reduced
hemoglobin or red-cell mass.\\[1.5pt]
eGFR $<60$ & V & Estimated glomerular filtration rate below
60\,mL/min/1.73\,m$^2$, indicating reduced kidney filtration.\\
\bottomrule
\end{tabular}
\end{appdefinitionpanel}

\textbf{Clinical prediction targets.}
The 15 classification and 17 regression targets in
Table~\ref{tab:repr-full} use recorded categories and numerical measurements,
respectively (Tables~\ref{tab:task-classification}--\ref{tab:task-regression}).
All are visit-level except VitalDB EtCO$_2$ and BIS, which are window-level
targets from capnography and EEG monitoring. MC-MED disease and device labels
use prior ICD-10 records. Its laboratory targets use the first available
ED-visit value; eGFR is calculated with the race-free CKD-EPI 2021 equation
from creatinine, age, and sex. For MIMIC-III, laboratory matching takes the
earliest available value from the 24 hours preceding the recording and the
recording itself; a measurement during the recording can therefore be retained
when no prior value is available. VitalDB albumin is preoperative. These tasks
evaluate agreement with clinical records and measurements at their specified
times, rather than uniformly prospective outcome prediction.

\begin{appdefinitionpanel}
\caption{\textbf{Clinical classification targets.} These rows cover all 15
dataset--task pairs in Table~\ref{tab:repr-full}; sex occurs in all three datasets.
Positive disease labels refer to recorded conditions.}
\label{tab:task-classification}
\begin{tabular}{@{}>{\raggedright\arraybackslash}p{0.23\linewidth}
>{\raggedright\arraybackslash}p{0.09\linewidth}
>{\raggedright\arraybackslash}p{0.61\linewidth}@{}}
\toprule
Task & Data & Definition and clinical interpretation\\
\midrule
\rowcolor[HTML]{F1F3F5}
\multicolumn{3}{@{}l}{\textbf{Shared demographic target}}\\[2pt]
Sex & M, I, V & Recorded patient sex.\\[1.5pt]
\midrule
\rowcolor{ppgrowblue}
\multicolumn{3}{@{}l}{\textbf{MC-MED: ED disposition and prior clinical history}}\\[2pt]
ED admission & M & Hospital admission following the ED encounter,
versus no admission.\\[1.5pt]
Atrial fibrillation & M & Prior atrial fibrillation: disorganized atrial
electrical activity associated with an irregular rhythm. The label records
history, not necessarily AF in the observed window.\\[1.5pt]
Heart failure & M & Prior heart failure, a clinical syndrome involving
impaired cardiac pumping and/or filling.\\[1.5pt]
Cardiac device & M & Prior recorded cardiac-device status.\\[1.5pt]
Chronic kidney disease & M & Prior chronic kidney disease, indicating
persistent impairment of kidney structure or function.\\[1.5pt]
Diabetes & M & Prior diabetes mellitus, a disorder of glucose regulation.\\[1.5pt]
Anemia & M & Prior anemia, characterized by reduced hemoglobin or red-cell
mass; the target is a diagnosis record, not a threshold applied to current Hb.\\[1.5pt]
\midrule
\rowcolor[HTML]{FDF1E6}
\multicolumn{3}{@{}l}{\textbf{MIMIC-III: hospital outcome and diagnostic category}}\\[2pt]
In-hospital mortality & I & Death during the indexed hospitalization.\\[1.5pt]
Circulatory diagnosis & I & The primary coded hospital diagnosis belongs
to the circulatory-system chapter of ICD-9.\\[1.5pt]
\midrule
\rowcolor[HTML]{F2ECFC}
\multicolumn{3}{@{}l}{\textbf{VitalDB: preoperative status and postoperative disposition}}\\[2pt]
ASA $\geq3$ & V & Preoperative American Society of Anesthesiologists physical
status III or higher; class III denotes severe systemic disease.\\[1.5pt]
Hypertension & V & Recorded preoperative hypertension, indicating
persistently elevated blood pressure.\\[1.5pt]
ICU admission & V & Subsequent postoperative admission to intensive care.\\
\bottomrule
\end{tabular}
\end{appdefinitionpanel}

\begin{appdefinitionpanel}
\caption{\textbf{Clinical regression targets.} These rows cover all 17
dataset--task pairs in Table~\ref{tab:repr-full}. Shared measurements are defined
once; sampling times and target levels are specified in the accompanying text.}
\label{tab:task-regression}
\begin{tabular}{@{}>{\raggedright\arraybackslash}p{0.23\linewidth}
>{\raggedright\arraybackslash}p{0.09\linewidth}
>{\raggedright\arraybackslash}p{0.61\linewidth}@{}}
\toprule
Task & Data & Measurement and clinical interpretation\\
\midrule
\rowcolor[HTML]{F1F3F5}
\multicolumn{3}{@{}l}{\textbf{Demographics and body size}}\\[2pt]
Age & M, I, V & Patient age in years. MIMIC-III values above 120 are treated
as missing because of de-identification shifts.\\[1.5pt]
BMI & V & Body mass index: weight divided by squared height (kg/m$^2$),
a measure of body size relative to height.\\[1.5pt]
\midrule
\rowcolor{ppgrowblue}
\multicolumn{3}{@{}l}{\textbf{Kidney function, hematology, and blood chemistry}}\\[2pt]
eGFR & M & Estimated glomerular filtration rate (mL/min/1.73\,m$^2$),
reflecting kidney filtration function. This target is continuous.\\[1.5pt]
Hemoglobin (Hb) & M, I & Blood concentration of the oxygen-carrying
protein in red blood cells.\\[1.5pt]
Albumin & M, V & Serum concentration of albumin, a major protein
synthesized by the liver.\\[1.5pt]
BUN & M, I & Blood urea nitrogen: the concentration of nitrogen in
circulating urea, related to nitrogen metabolism and renal handling.\\[1.5pt]
Potassium & M & Blood potassium concentration, relevant to electrical
activity of nerve and muscle cells, including the heart.\\[1.5pt]
Calcium & M & Blood calcium concentration, relevant to neuromuscular
function and multiple cellular processes.\\[1.5pt]
Sodium & M & Blood sodium concentration, relevant to water balance
and extracellular fluid regulation.\\[1.5pt]
Glucose & M & Blood glucose concentration, reflecting glycemic status
at the sampled time.\\[1.5pt]
\midrule
\rowcolor[HTML]{F2ECFC}
\multicolumn{3}{@{}l}{\textbf{Independent intraoperative monitoring channels}}\\[2pt]
EtCO$_2$ & V & End-tidal carbon dioxide partial pressure measured by
capnography (mmHg), describing expired CO$_2$ at the end of a breath.\\[1.5pt]
BIS & V & EEG-derived, unitless bispectral index, reflecting the
hypnotic effect of anesthesia.\\
\bottomrule
\end{tabular}
\end{appdefinitionpanel}

\subsection{Cross-modal retrieval}
\label{app:experiment-retrieval}

We extend the 100-candidate evaluation in Table~\ref{tab:retrieval} to pools
of 2,000 candidates, testing how well \ppglm{} identifies paired signals and
descriptions as the search space grows. Table~\ref{tab:retrieval-extra} reports
both signal-to-text (S2T) and text-to-signal (T2S) retrieval on MC-MED,
MIMIC-III, and VitalDB. Candidates are ranked by cosine similarity in the
aligned embedding space. We report R@1, R@5, and R@10: the fraction of queries
whose paired target appears among the top one, five, or ten candidates.
The 100-candidate results average two disjoint pools, each with one window
per patient and distinct reference captions.

\textbf{Effect of candidate-pool size.}
Expanding the pool from 100 to 2,000 candidates reduces recall in both
directions, with larger R@1 declines on the external datasets. On MC-MED,
S2T/T2S R@1 decreases from 0.975/0.965 to 0.780/0.744, while R@10 remains
0.998/0.995. At the same pool size, MIMIC-III reaches R@10 of 0.853/0.762
and VitalDB reaches 0.510/0.483. Thus, matching the exact pair becomes more
difficult as the candidate pool grows, particularly after transfer to the
external datasets, while MC-MED retains nearly all paired targets within
the top ten results.

\begin{center}
\begin{minipage}{0.82\linewidth}
\makeatletter\def\@captype{table}\makeatother
\centering\small
\caption{\textbf{Bidirectional retrieval with varying candidate pools.}
\ppglm{} recall ($\uparrow$) with $N=100$ or 2,000 candidates.
S2T: signal to text; T2S: text to signal.}
\label{tab:retrieval-extra}
\setlength{\tabcolsep}{6pt}
\renewcommand{\arraystretch}{1.12}
\begin{tabular*}{\linewidth}{@{\extracolsep{\fill}}lrccc@{}}
\toprule
Direction & Candidates ($N$) & R@1$\uparrow$ & R@5$\uparrow$ & R@10$\uparrow$\\
\midrule
\rowcolor[HTML]{ECF4FF}[0pt][0pt]
\multicolumn{5}{@{}l}{\strut\textbf{MC-MED (ED)}}\\
\multirow{2}{*}{S2T} & 100 & 0.975 & 1.000 & 1.000\\
 & 2,000 & 0.780 & 0.988 & 0.998\\
\cmidrule(l){2-5}
\multirow{2}{*}{T2S} & 100 & 0.965 & 1.000 & 1.000\\
 & 2,000 & 0.744 & 0.980 & 0.995\\
\midrule
\rowcolor[HTML]{FDF1E6}[0pt][0pt]
\multicolumn{5}{@{}l}{\strut\textbf{MIMIC-III (ICU)}}\\
\multirow{2}{*}{S2T} & 100 & 0.805 & 1.000 & 1.000\\
 & 2,000 & 0.258 & 0.671 & 0.853\\
\cmidrule(l){2-5}
\multirow{2}{*}{T2S} & 100 & 0.755 & 0.990 & 1.000\\
 & 2,000 & 0.207 & 0.560 & 0.762\\
\midrule
\rowcolor[HTML]{F2ECFC}[0pt][0pt]
\multicolumn{5}{@{}l}{\strut\textbf{VitalDB (OR)}}\\
\multirow{2}{*}{S2T} & 100 & 0.555 & 0.885 & 0.965\\
 & 2,000 & 0.108 & 0.344 & 0.510\\
\cmidrule(l){2-5}
\multirow{2}{*}{T2S} & 100 & 0.550 & 0.905 & 0.955\\
 & 2,000 & 0.097 & 0.324 & 0.483\\
\bottomrule
\end{tabular*}
\end{minipage}
\end{center}

\subsection{Caption evaluation and numerical read-out}
\label{app:experiment-caption}

We assess the factual accuracy of PPG descriptions and the accuracy of heart
rates read from those descriptions. We compare two \ppglm{} interfaces:
\textbf{direct caption generation}, which produces a new description from the
waveform using the caption decoder with greedy decoding; and
\textbf{nearest-caption retrieval}, which returns the highest-scoring existing
caption by embedding similarity. Retrieval uses a fixed bank of 20,000 MC-MED
captions from patients disjoint from the evaluation queries, including for
external evaluation on MIMIC-III and VitalDB.

The VitalDB analyses here use the official 40\,Hz waveform view of the same
evaluation items. The main \ppglm{} results in Table~\ref{tab:caption} use the
original 500\,Hz source, accounting for the different VitalDB values in
Tables~\ref{tab:caption-extra} and~\ref{tab:hr-extra}.

\textbf{Caption evaluation.}
The deterministic parser checks stated fields against reference measurements;
a numerical HR statement is correct within $\pm5$\,bpm. Fact accuracy is the
number of correct stated fields divided by all stated fields, and coverage is
the fraction of gradable fields that are stated. Table~\ref{tab:caption-extra}
uses the six common fields across datasets for fact accuracy. The original
MC-MED evaluation additionally grades perfusion and morphology, yielding
an eight-field accuracy of 0.932 for direct generation and 0.889 for
nearest-caption retrieval, versus 0.923 and 0.877 for the common six fields.

The LLM judge, GPT-5.6-luna, receives the reference fact sheet and candidate caption. It
classifies fields as correct, incorrect, or absent, counts unsupported claims,
and scores factuality and completeness from 1 to 5. Optional charted vitals
serve as supporting evidence but do not count toward completeness.
Pairwise evaluation uses both presentation orders (400 judgments for 200
windows); generated \ppglm{} captions are preferred to GPT-5.6-luna captions
at rates of 0.922, 0.855, and 0.777 on MC-MED, MIMIC-III, and VitalDB.
As a calibration check, reference captions receive factuality scores of
4.40, 4.80, and 4.85, with 0.86, 0.23, and 0.17 unsupported claims per caption,
respectively, indicating that the judge is an imperfect evaluator.
Direct generation has higher parser fact accuracy on all three datasets,
whereas nearest-caption retrieval receives higher judge factuality scores
and fewer unsupported claims (Table~\ref{tab:caption-extra}).

\begin{center}
\begin{minipage}{0.98\linewidth}
\makeatletter\def\@captype{table}\makeatother
\centering\footnotesize
\caption{\textbf{Caption factuality by output method.} Direct generation
produces a new caption; nearest-caption retrieval selects an existing caption
from the fixed bank. Fact accuracy covers six common fields; coverage includes
all gradable fields. Judge factuality is on a 1--5 scale; unsupported claims
are counted per caption. VitalDB uses the 40\,Hz view. Bold marks the best
value within each dataset.}
\label{tab:caption-extra}
\setlength{\tabcolsep}{4pt}
\renewcommand{\arraystretch}{1.10}
\begin{tabular*}{\linewidth}{@{\extracolsep{\fill}}lcccc@{}}
\toprule
 & \multicolumn{2}{c}{Rule-based evaluation} & \multicolumn{2}{c}{LLM-judge evaluation}\\
\cmidrule(lr){2-3}\cmidrule(l){4-5}
Model / caption source & Fact acc.$\uparrow$ & Coverage$\uparrow$ & Factuality$\uparrow$ & Unsupported$\downarrow$\\
\midrule
\rowcolor[HTML]{ECF4FF}[0pt][0pt]
\multicolumn{5}{@{}l}{\strut\textbf{MC-MED (ED)}}\\
GPT-5.6-luna & 0.499 & 0.99 & 2.63 & 6.54\\
\ppglm{}: direct generation & \textbf{0.923} & \textbf{1.00} & 3.64 & 2.94\\
\ppglm{}: nearest-caption retrieval & 0.877 & \textbf{1.00} & \textbf{3.83} & \textbf{1.51}\\
\midrule
\rowcolor[HTML]{FDF1E6}[0pt][0pt]
\multicolumn{5}{@{}l}{\strut\textbf{MIMIC-III (ICU)}}\\
GPT-5.6-luna & 0.456 & 0.95 & 2.83 & 3.31\\
\ppglm{}: direct generation & \textbf{0.920} & \textbf{1.00} & 3.59 & 2.88\\
\ppglm{}: nearest-caption retrieval & 0.859 & \textbf{1.00} & \textbf{3.67} & \textbf{1.47}\\
\midrule
\rowcolor[HTML]{F2ECFC}[0pt][0pt]
\multicolumn{5}{@{}l}{\strut\textbf{VitalDB (OR)}}\\
GPT-5.6-luna & 0.486 & 0.98 & 2.88 & 3.21\\
\ppglm{}: direct generation & \textbf{0.837} & 0.98 & 3.30 & 3.17\\
\ppglm{}: nearest-caption retrieval & 0.778 & \textbf{1.00} & \textbf{3.43} & \textbf{1.52}\\
\bottomrule
\end{tabular*}
\end{minipage}
\end{center}

\textbf{Heart-rate read-out.}
For each \ppglm{} interface, we extract the numerical HR stated in its output
caption and compare it with the synchronized ECG reference. Table~\ref{tab:hr-extra}
also includes two digital signal processing (DSP) baselines based on the dominant
frequency and detected pulse peaks. HR read from directly generated captions
has the lowest MAE and the highest fraction of estimates within 5\,bpm on all
three datasets. On MC-MED, its MAE is 0.27\,bpm against the PPG-derived annotation
rate but 3.90\,bpm against ECG; the independent ECG reference therefore assesses
agreement beyond the caption-construction target.

\begin{center}
\begin{minipage}{0.88\linewidth}
\makeatletter\def\@captype{table}\makeatother
\centering\small
\caption{\textbf{Heart-rate estimation from waveforms or captions.}
DSP estimates HR directly from the waveform; \ppglm{} reads HR from a generated
or retrieved caption. All entries use 300 valid ECG-referenced windows per
dataset. MAE is in bpm; ``Within 5 bpm'' is a fraction. VitalDB uses the 40\,Hz
view. Bold marks the best value within each dataset.}
\label{tab:hr-extra}
\setlength{\tabcolsep}{5pt}
\renewcommand{\arraystretch}{1.10}
\begin{tabular*}{\linewidth}{@{\extracolsep{\fill}}lcc@{}}
\toprule
Method / HR source & MAE$\downarrow$ & Within 5 bpm$\uparrow$\\
\midrule
\rowcolor[HTML]{ECF4FF}[0pt][0pt]
\multicolumn{3}{@{}l}{\strut\textbf{MC-MED (ED)}}\\
DSP: dominant frequency & 6.07 & 0.857\\
DSP: pulse peaks & 6.36 & 0.690\\
\ppglm{}: HR from retrieved caption & 4.70 & 0.847\\
\ppglm{}: HR from generated caption & \textbf{3.90} & \textbf{0.877}\\
\midrule
\rowcolor[HTML]{FDF1E6}[0pt][0pt]
\multicolumn{3}{@{}l}{\strut\textbf{MIMIC-III (ICU)}}\\
DSP: dominant frequency & 6.23 & 0.823\\
DSP: pulse peaks & 6.40 & 0.680\\
\ppglm{}: HR from retrieved caption & 5.36 & 0.833\\
\ppglm{}: HR from generated caption & \textbf{4.27} & \textbf{0.860}\\
\midrule
\rowcolor[HTML]{F2ECFC}[0pt][0pt]
\multicolumn{3}{@{}l}{\strut\textbf{VitalDB (OR)}}\\
DSP: dominant frequency & 5.53 & 0.870\\
DSP: pulse peaks & 7.93 & 0.723\\
\ppglm{}: HR from retrieved caption & 4.74 & 0.867\\
\ppglm{}: HR from generated caption & \textbf{4.08} & \textbf{0.900}\\
\bottomrule
\end{tabular*}
\end{minipage}
\end{center}

\subsection{Additional Visit-Level Clinical-Fact Recognition}
\label{app:experiment-facts}
\label{app:experiment-visit}

Complementing Section~\ref{sec:exp-recognition} and
Table~\ref{tab:recognition-facts}, Table~\ref{tab:recognition-context-extra}
reports additional clinical-fact recognition on 400 MC-MED visits with eight
PPG windows each. \ppglm{} averages frozen pretrained fact embeddings across
windows and scores them against textual fact descriptions. Read-out thresholds are calibrated
on held-out MC-MED validation visits, with no calibration on test visits.

Panel~(a) covers eight triage-context tasks: high acuity (ESI 1--2), arrival by
ambulance versus self-arrival, three heart-rate (HR) classes, low oxygen saturation
(SpO$_2<95\%$), high respiratory rate (RR $>20$ breaths/min), fever
(Temp. $\geq38\,^{\circ}$C), four blood-pressure (BP)
classes, and ten chief-complaint (CC) groups. Balanced accuracy accounts for class
imbalance in both binary and multiclass tasks.

Panel~(b) covers eleven binary recorded conditions: hypertension (HTN), atrial
fibrillation (AF), heart failure (HF), coronary artery disease (CAD), type~2 diabetes
mellitus (DM), chronic kidney disease (CKD), anemia, obstructive sleep apnea (OSA),
asthma, obesity, and stroke. \ppglm{} achieves the highest reported balanced
accuracy on every task in both panels, reaching 0.785 for HR and 0.729 for fever,
with medical-history scores ranging from 0.580 to 0.687. These results measure
associations with recorded visit attributes rather than independent clinical
diagnoses.

\begin{table}[!htbp]
\centering\small
\caption{\textbf{Additional visit-level clinical-fact recognition.}
Balanced accuracy ($\uparrow$) for (a) triage context and (b) medical history on
400 MC-MED visits using eight windows each. \ppglm{} read-outs use thresholds
calibrated on held-out MC-MED validation visits. Bold marks column maxima.}
\label{tab:recognition-context-extra}
\footnotesize
\setlength{\tabcolsep}{2pt}
\renewcommand{\arraystretch}{1.08}
\begin{tabular*}{\linewidth}{@{\extracolsep{\fill}}l*{8}{c}@{}}
\toprule
\multicolumn{9}{@{}l}{\textbf{(a) Triage-context recognition}}\\
\midrule
Model & ESI$\uparrow$ & Arrival$\uparrow$ & HR$\uparrow$ & SpO$_2$$\uparrow$ & RR$\uparrow$ & Temp.$\uparrow$ & BP$\uparrow$ & CC$\uparrow$\\
\midrule
GPT-5.6-luna & 0.501 & 0.445 & 0.463 & 0.497 & 0.506 & 0.500 & 0.269 & 0.134\\
GPT-6-luna & 0.500 & 0.498 & 0.486 & 0.500 & 0.500 & 0.500 & 0.238 & 0.128\\
GPT-6-sol & 0.536 & 0.509 & 0.736 & 0.499 & 0.506 & 0.512 & 0.265 & 0.126\\
Opus~5.5 & 0.501 & 0.504 & 0.528 & 0.500 & 0.500 & 0.500 & 0.259 & 0.100\\
PulseLM & 0.483 & 0.500 & 0.656 & 0.499 & 0.500 & 0.498 & 0.250 & 0.113\\
Qwen3-4B (SFT) & 0.547 & 0.500 & 0.625 & 0.500 & 0.500 & 0.500 & 0.250 & 0.100\\
\noalign{\begingroup\color{ppgrowblue}\hrule height\dimexpr\ht\strutbox+\dp\strutbox\relax\endgroup\vskip-\dimexpr\ht\strutbox+\dp\strutbox\relax}
\textbf{PPG-LM} & \textbf{0.618} & \textbf{0.607} & \textbf{0.785} & \textbf{0.650} & \textbf{0.692} & \textbf{0.729} & \textbf{0.442} & \textbf{0.202}\\
\bottomrule
\end{tabular*}
\par\vspace{0.75em}
\begin{tabular*}{\linewidth}{@{\extracolsep{\fill}}l*{11}{c}@{}}
\toprule
\multicolumn{12}{@{}l}{\textbf{(b) Medical-history recognition}}\\
\midrule
Model & HTN$\uparrow$ & AF$\uparrow$ & HF$\uparrow$ & CAD$\uparrow$ & DM$\uparrow$ & CKD$\uparrow$ & Anemia$\uparrow$ & OSA$\uparrow$ & Asthma$\uparrow$ & Obesity$\uparrow$ & Stroke$\uparrow$\\
\midrule
GPT-5.6-luna & 0.500 & 0.502 & 0.495 & 0.460 & 0.483 & 0.481 & 0.496 & 0.531 & 0.500 & 0.478 & 0.512\\
GPT-6-luna & 0.506 & 0.499 & 0.494 & 0.464 & 0.525 & 0.510 & 0.498 & 0.513 & 0.499 & 0.526 & 0.500\\
GPT-6-sol & 0.498 & 0.598 & 0.524 & 0.495 & 0.490 & 0.500 & 0.500 & 0.500 & 0.500 & 0.522 & 0.500\\
Opus~5.5 & 0.547 & 0.659 & 0.546 & 0.530 & 0.487 & 0.494 & 0.490 & 0.500 & 0.500 & 0.500 & 0.500\\
PulseLM & 0.504 & 0.534 & 0.500 & 0.500 & 0.500 & 0.500 & 0.500 & 0.500 & 0.500 & 0.500 & 0.500\\
Qwen3-4B (SFT) & 0.500 & 0.500 & 0.500 & 0.500 & 0.500 & 0.500 & 0.500 & 0.500 & 0.500 & 0.500 & 0.500\\
\noalign{\begingroup\color{ppgrowblue}\hrule height\dimexpr\ht\strutbox+\dp\strutbox\relax\endgroup\vskip-\dimexpr\ht\strutbox+\dp\strutbox\relax}
\textbf{PPG-LM} & \textbf{0.583} & \textbf{0.676} & \textbf{0.687} & \textbf{0.656} & \textbf{0.586} & \textbf{0.682} & \textbf{0.638} & \textbf{0.599} & \textbf{0.609} & \textbf{0.580} & \textbf{0.593}\\
\bottomrule
\end{tabular*}
\end{table}

\subsection{Medication-event evaluation}
\label{app:experiment-events}

\textbf{Main medication-class benchmark.}
Table~\ref{tab:medication-top-classes} uses 393 administrations and 393
controls on MC-MED and 139 administrations and 139 controls on VitalDB.
Medication names are mapped to the classes used during training. Each view
selects up to three nearest windows within 30 minutes before the anchor and
five uniformly spaced windows within 60 minutes after it, retaining signed
time offsets. Each reported AUROC compares one medication class with
no-administration controls; other medication classes are excluded from that
comparison's negative group. These are class-specific ranking tasks, rather
than a single measure of whether any medication was administered.

\textbf{Supplementary smaller-cohort evaluation.}
The earlier language event benchmark contains 150 administrations and 150 controls on
MC-MED, with up to 15 administrations per medication class. Its selection windows
are 30 minutes before and 60 minutes after the anchor, with up to three pre-event
and five post-event observations. VitalDB contains 60 events and 60 controls within
$\pm2$ minutes; MIMIC-III has no event source in this benchmark. Same-visit paired
accuracy uses the 56 MC-MED and 19 VitalDB pairs with both views available, assigning
half credit to ties. It is not an accuracy over all event/control items. GPT-5.6-luna
has tie fractions 0.70 and 0.53, respectively. Its reported AUROC comes from a Boolean
answer and equals balanced accuracy, whereas \ppglm{} uses a continuous difference
between the highest medication-class score and the no-medication score. On MC-MED,
\ppglm{} balanced accuracy is 0.647, compared with 0.550 for the LLM; on VitalDB the
values are 0.533 and 0.633. The separate validation diagnostic in the training report
uses a different item set (AUROC 0.711, paired accuracy 0.687); those values are not
directly comparable to the class-versus-control AUROCs in
Table~\ref{tab:medication-top-classes}.

\subsection{Full Clinical Prediction Results}
\label{app:experiment-clinical-results}
\label{app:experiment-probes}

This subsection provides all 32 dataset--task pairs evaluated in
Section~\ref{sec:exp-repr}: 17 on MC-MED, six on MIMIC-III, and nine on VitalDB,
comprising 15 classification and 17 regression tasks.
Table~\ref{tab:repr-full} retains the complete results;
Table~\ref{tab:repr} presents six tasks per dataset.
Task definitions are given in Appendix~\ref{app:experiment-clinical-tasks}.

\textbf{Evaluation protocol.}
Logistic-regression classifiers and ridge-regression models are fitted on frozen
representations using each dataset's training patients and evaluated on its
test patients. Classification uses AUROC and regression uses $R^2$; results
are reported as mean $\pm$ standard deviation. External-dataset probes therefore
assess supervised transfer of the pretrained representations.
For visit-level tasks on MC-MED, MIMIC-III, and VitalDB, all foundation-model
baselines (including AnyPPG) and both \ppglm{} stages use the same four PPG
windows per visit, with evenly spaced windows on MC-MED.
Baselines and Stage~1 mean-pool the four window
features, while Stage~2 applies its learned visit aggregator to those same four
windows. The observation budget is therefore matched across models and stages,
while the visit read-out differs. VitalDB EtCO$_2$ and BIS are evaluated on the
same individual windows across models.

MC-MED diagnosis targets denote recorded prior diagnoses. Its eGFR target is derived
from creatinine, age, and sex and is not independent of the demographic targets.
Laboratory timing differs across datasets: MC-MED uses the first ED-visit value,
MIMIC-III uses the available admission/recording-associated value, and VitalDB uses
preoperative measurements. MIMIC-III's matching can retain a value measured during
the recording when no earlier value is present. These probes therefore assess
association with recorded clinical labels rather than a uniform prospective endpoint.

\textbf{Full clinical prediction results.}
Stage~2 exceeds all six foundation-model baselines in mean performance on all
32 tasks and improves over Stage~1 on 30. Stage~1 remains slightly stronger on
VitalDB hypertension (0.656 vs.\ 0.655) and EtCO$_2$ (0.153 vs.\ 0.152).
Despite these relative gains, low $R^2$ for electrolytes, glucose, and BMI
indicates limited predictive performance for these targets.

\begin{table}[!htbp]
\centering\small
\caption{\textbf{Full clinical prediction results.}
Test AUROC and $R^2$ (mean $\pm$ SD; $\uparrow$) for all 32 tasks.
All models use the same four windows for visit-level targets.
Bold: best mean; $\dagger$: window-level target.}
\label{tab:repr-full}
\setlength{\tabcolsep}{1.4pt}
\renewcommand{\arraystretch}{0.96}
\newcommand{\clinicalgroup}[2]{%
  \multicolumn{9}{@{}l@{}}{%
    \begingroup\setlength{\fboxsep}{1.4pt}%
    \colorbox[HTML]{#1}{%
      \makebox[\dimexpr\linewidth*10/9-2\fboxsep\relax][l]{%
        \strut\textbf{\textit{#2}}}}%
    \endgroup}\\}
\scalebox{0.88}{%
\begin{tabular}{lcccccccc}
\toprule
& \multicolumn{6}{c}{Foundation models} & \multicolumn{2}{c}{\ppglm{}}\\
\cmidrule(lr){2-7}\cmidrule(lr){8-9}
Task & \shortstack{PaPaGei-S} & \shortstack{PulsePPG} & AnyPPG & \shortstack{SIGMA-PPG} & \shortstack{MOMENT-L} & \shortstack{Chronos-2} & Stage 1 & Stage 2\\
\midrule
\clinicalgroup{ECF4FF}{MC-MED (ED) --- AUROC $\uparrow$}
Sex & 0.692{\fontsize{5.59}{5.59}\selectfont$\pm$0.006} & 0.719{\fontsize{5.59}{5.59}\selectfont$\pm$0.007} & 0.783{\fontsize{5.59}{5.59}\selectfont$\pm$0.008} & 0.712{\fontsize{5.59}{5.59}\selectfont$\pm$0.008} & 0.724{\fontsize{5.59}{5.59}\selectfont$\pm$0.005} & 0.739{\fontsize{5.59}{5.59}\selectfont$\pm$0.008} & 0.813{\fontsize{5.59}{5.59}\selectfont$\pm$0.006} & \textbf{0.856}{\fontsize{5.59}{5.59}\selectfont$\pm$0.005}\\
ED admission & 0.673{\fontsize{5.59}{5.59}\selectfont$\pm$0.011} & 0.681{\fontsize{5.59}{5.59}\selectfont$\pm$0.009} & 0.750{\fontsize{5.59}{5.59}\selectfont$\pm$0.013} & 0.679{\fontsize{5.59}{5.59}\selectfont$\pm$0.011} & 0.700{\fontsize{5.59}{5.59}\selectfont$\pm$0.011} & 0.728{\fontsize{5.59}{5.59}\selectfont$\pm$0.013} & 0.765{\fontsize{5.59}{5.59}\selectfont$\pm$0.012} & \textbf{0.787}{\fontsize{5.59}{5.59}\selectfont$\pm$0.010}\\
Atrial fibrillation & 0.742{\fontsize{5.59}{5.59}\selectfont$\pm$0.012} & 0.752{\fontsize{5.59}{5.59}\selectfont$\pm$0.013} & 0.845{\fontsize{5.59}{5.59}\selectfont$\pm$0.010} & 0.755{\fontsize{5.59}{5.59}\selectfont$\pm$0.013} & 0.768{\fontsize{5.59}{5.59}\selectfont$\pm$0.014} & 0.815{\fontsize{5.59}{5.59}\selectfont$\pm$0.012} & 0.857{\fontsize{5.59}{5.59}\selectfont$\pm$0.010} & \textbf{0.870}{\fontsize{5.59}{5.59}\selectfont$\pm$0.011}\\
Heart failure & 0.708{\fontsize{5.59}{5.59}\selectfont$\pm$0.018} & 0.723{\fontsize{5.59}{5.59}\selectfont$\pm$0.015} & 0.825{\fontsize{5.59}{5.59}\selectfont$\pm$0.013} & 0.711{\fontsize{5.59}{5.59}\selectfont$\pm$0.017} & 0.731{\fontsize{5.59}{5.59}\selectfont$\pm$0.015} & 0.781{\fontsize{5.59}{5.59}\selectfont$\pm$0.012} & 0.836{\fontsize{5.59}{5.59}\selectfont$\pm$0.010} & \textbf{0.860}{\fontsize{5.59}{5.59}\selectfont$\pm$0.009}\\
Cardiac device & 0.713{\fontsize{5.59}{5.59}\selectfont$\pm$0.023} & 0.729{\fontsize{5.59}{5.59}\selectfont$\pm$0.018} & 0.835{\fontsize{5.59}{5.59}\selectfont$\pm$0.022} & 0.733{\fontsize{5.59}{5.59}\selectfont$\pm$0.022} & 0.719{\fontsize{5.59}{5.59}\selectfont$\pm$0.017} & 0.784{\fontsize{5.59}{5.59}\selectfont$\pm$0.023} & 0.836{\fontsize{5.59}{5.59}\selectfont$\pm$0.022} & \textbf{0.857}{\fontsize{5.59}{5.59}\selectfont$\pm$0.025}\\
Chronic kidney disease & 0.647{\fontsize{5.59}{5.59}\selectfont$\pm$0.017} & 0.666{\fontsize{5.59}{5.59}\selectfont$\pm$0.016} & 0.761{\fontsize{5.59}{5.59}\selectfont$\pm$0.016} & 0.666{\fontsize{5.59}{5.59}\selectfont$\pm$0.014} & 0.673{\fontsize{5.59}{5.59}\selectfont$\pm$0.013} & 0.735{\fontsize{5.59}{5.59}\selectfont$\pm$0.013} & 0.774{\fontsize{5.59}{5.59}\selectfont$\pm$0.013} & \textbf{0.792}{\fontsize{5.59}{5.59}\selectfont$\pm$0.014}\\
Diabetes & 0.644{\fontsize{5.59}{5.59}\selectfont$\pm$0.012} & 0.649{\fontsize{5.59}{5.59}\selectfont$\pm$0.014} & 0.717{\fontsize{5.59}{5.59}\selectfont$\pm$0.015} & 0.660{\fontsize{5.59}{5.59}\selectfont$\pm$0.020} & 0.661{\fontsize{5.59}{5.59}\selectfont$\pm$0.016} & 0.698{\fontsize{5.59}{5.59}\selectfont$\pm$0.016} & 0.728{\fontsize{5.59}{5.59}\selectfont$\pm$0.016} & \textbf{0.740}{\fontsize{5.59}{5.59}\selectfont$\pm$0.016}\\
Anemia & 0.608{\fontsize{5.59}{5.59}\selectfont$\pm$0.008} & 0.604{\fontsize{5.59}{5.59}\selectfont$\pm$0.011} & 0.668{\fontsize{5.59}{5.59}\selectfont$\pm$0.014} & 0.620{\fontsize{5.59}{5.59}\selectfont$\pm$0.010} & 0.621{\fontsize{5.59}{5.59}\selectfont$\pm$0.008} & 0.654{\fontsize{5.59}{5.59}\selectfont$\pm$0.010} & 0.686{\fontsize{5.59}{5.59}\selectfont$\pm$0.015} & \textbf{0.709}{\fontsize{5.59}{5.59}\selectfont$\pm$0.012}\\
\midrule
\clinicalgroup{ECF4FF}{MC-MED (ED) --- $R^2\uparrow$}
Age & 0.395{\fontsize{5.59}{5.59}\selectfont$\pm$0.016} & 0.428{\fontsize{5.59}{5.59}\selectfont$\pm$0.018} & 0.626{\fontsize{5.59}{5.59}\selectfont$\pm$0.020} & 0.389{\fontsize{5.59}{5.59}\selectfont$\pm$0.017} & 0.475{\fontsize{5.59}{5.59}\selectfont$\pm$0.017} & 0.521{\fontsize{5.59}{5.59}\selectfont$\pm$0.016} & 0.691{\fontsize{5.59}{5.59}\selectfont$\pm$0.016} & \textbf{0.782}{\fontsize{5.59}{5.59}\selectfont$\pm$0.014}\\
eGFR & 0.155{\fontsize{5.59}{5.59}\selectfont$\pm$0.015} & 0.173{\fontsize{5.59}{5.59}\selectfont$\pm$0.016} & 0.320{\fontsize{5.59}{5.59}\selectfont$\pm$0.020} & 0.169{\fontsize{5.59}{5.59}\selectfont$\pm$0.012} & 0.202{\fontsize{5.59}{5.59}\selectfont$\pm$0.017} & 0.262{\fontsize{5.59}{5.59}\selectfont$\pm$0.017} & 0.356{\fontsize{5.59}{5.59}\selectfont$\pm$0.021} & \textbf{0.404}{\fontsize{5.59}{5.59}\selectfont$\pm$0.024}\\
Hemoglobin & 0.083{\fontsize{5.59}{5.59}\selectfont$\pm$0.011} & 0.098{\fontsize{5.59}{5.59}\selectfont$\pm$0.013} & 0.196{\fontsize{5.59}{5.59}\selectfont$\pm$0.013} & 0.102{\fontsize{5.59}{5.59}\selectfont$\pm$0.012} & 0.123{\fontsize{5.59}{5.59}\selectfont$\pm$0.014} & 0.164{\fontsize{5.59}{5.59}\selectfont$\pm$0.014} & 0.238{\fontsize{5.59}{5.59}\selectfont$\pm$0.014} & \textbf{0.274}{\fontsize{5.59}{5.59}\selectfont$\pm$0.012}\\
Albumin & 0.093{\fontsize{5.59}{5.59}\selectfont$\pm$0.015} & 0.094{\fontsize{5.59}{5.59}\selectfont$\pm$0.011} & 0.207{\fontsize{5.59}{5.59}\selectfont$\pm$0.015} & 0.097{\fontsize{5.59}{5.59}\selectfont$\pm$0.012} & 0.122{\fontsize{5.59}{5.59}\selectfont$\pm$0.016} & 0.185{\fontsize{5.59}{5.59}\selectfont$\pm$0.017} & 0.239{\fontsize{5.59}{5.59}\selectfont$\pm$0.018} & \textbf{0.286}{\fontsize{5.59}{5.59}\selectfont$\pm$0.013}\\
BUN & 0.053{\fontsize{5.59}{5.59}\selectfont$\pm$0.009} & 0.062{\fontsize{5.59}{5.59}\selectfont$\pm$0.015} & 0.152{\fontsize{5.59}{5.59}\selectfont$\pm$0.015} & 0.063{\fontsize{5.59}{5.59}\selectfont$\pm$0.009} & 0.070{\fontsize{5.59}{5.59}\selectfont$\pm$0.014} & 0.114{\fontsize{5.59}{5.59}\selectfont$\pm$0.010} & 0.178{\fontsize{5.59}{5.59}\selectfont$\pm$0.019} & \textbf{0.213}{\fontsize{5.59}{5.59}\selectfont$\pm$0.019}\\
Potassium & 0.025{\fontsize{5.59}{5.59}\selectfont$\pm$0.006} & 0.023{\fontsize{5.59}{5.59}\selectfont$\pm$0.009} & 0.057{\fontsize{5.59}{5.59}\selectfont$\pm$0.008} & 0.029{\fontsize{5.59}{5.59}\selectfont$\pm$0.006} & 0.025{\fontsize{5.59}{5.59}\selectfont$\pm$0.008} & 0.041{\fontsize{5.59}{5.59}\selectfont$\pm$0.008} & 0.067{\fontsize{5.59}{5.59}\selectfont$\pm$0.007} & \textbf{0.086}{\fontsize{5.59}{5.59}\selectfont$\pm$0.010}\\
Calcium & 0.002{\fontsize{5.59}{5.59}\selectfont$\pm$0.006} & 0.011{\fontsize{5.59}{5.59}\selectfont$\pm$0.010} & 0.041{\fontsize{5.59}{5.59}\selectfont$\pm$0.008} & 0.010{\fontsize{5.59}{5.59}\selectfont$\pm$0.005} & 0.001{\fontsize{5.59}{5.59}\selectfont$\pm$0.008} & 0.024{\fontsize{5.59}{5.59}\selectfont$\pm$0.009} & 0.051{\fontsize{5.59}{5.59}\selectfont$\pm$0.009} & \textbf{0.076}{\fontsize{5.59}{5.59}\selectfont$\pm$0.012}\\
Sodium & 0.017{\fontsize{5.59}{5.59}\selectfont$\pm$0.010} & 0.023{\fontsize{5.59}{5.59}\selectfont$\pm$0.008} & 0.056{\fontsize{5.59}{5.59}\selectfont$\pm$0.010} & 0.026{\fontsize{5.59}{5.59}\selectfont$\pm$0.008} & 0.017{\fontsize{5.59}{5.59}\selectfont$\pm$0.007} & 0.045{\fontsize{5.59}{5.59}\selectfont$\pm$0.010} & 0.068{\fontsize{5.59}{5.59}\selectfont$\pm$0.012} & \textbf{0.091}{\fontsize{5.59}{5.59}\selectfont$\pm$0.015}\\
Glucose & 0.032{\fontsize{5.59}{5.59}\selectfont$\pm$0.007} & 0.039{\fontsize{5.59}{5.59}\selectfont$\pm$0.007} & 0.065{\fontsize{5.59}{5.59}\selectfont$\pm$0.008} & 0.041{\fontsize{5.59}{5.59}\selectfont$\pm$0.009} & 0.028{\fontsize{5.59}{5.59}\selectfont$\pm$0.007} & 0.049{\fontsize{5.59}{5.59}\selectfont$\pm$0.011} & 0.071{\fontsize{5.59}{5.59}\selectfont$\pm$0.007} & \textbf{0.083}{\fontsize{5.59}{5.59}\selectfont$\pm$0.011}\\
\midrule
\clinicalgroup{FDF1E6}{MIMIC-III (ICU) --- AUROC $\uparrow$}
Sex & 0.651{\fontsize{5.59}{5.59}\selectfont$\pm$0.024} & 0.686{\fontsize{5.59}{5.59}\selectfont$\pm$0.014} & 0.713{\fontsize{5.59}{5.59}\selectfont$\pm$0.018} & 0.672{\fontsize{5.59}{5.59}\selectfont$\pm$0.016} & 0.682{\fontsize{5.59}{5.59}\selectfont$\pm$0.019} & 0.664{\fontsize{5.59}{5.59}\selectfont$\pm$0.020} & 0.730{\fontsize{5.59}{5.59}\selectfont$\pm$0.019} & \textbf{0.733}{\fontsize{5.59}{5.59}\selectfont$\pm$0.018}\\
In-hospital mortality & 0.697{\fontsize{5.59}{5.59}\selectfont$\pm$0.029} & 0.741{\fontsize{5.59}{5.59}\selectfont$\pm$0.034} & 0.756{\fontsize{5.59}{5.59}\selectfont$\pm$0.031} & 0.732{\fontsize{5.59}{5.59}\selectfont$\pm$0.025} & 0.751{\fontsize{5.59}{5.59}\selectfont$\pm$0.025} & 0.772{\fontsize{5.59}{5.59}\selectfont$\pm$0.020} & 0.786{\fontsize{5.59}{5.59}\selectfont$\pm$0.022} & \textbf{0.793}{\fontsize{5.59}{5.59}\selectfont$\pm$0.021}\\
Circulatory diagnosis & 0.666{\fontsize{5.59}{5.59}\selectfont$\pm$0.015} & 0.690{\fontsize{5.59}{5.59}\selectfont$\pm$0.022} & 0.725{\fontsize{5.59}{5.59}\selectfont$\pm$0.017} & 0.658{\fontsize{5.59}{5.59}\selectfont$\pm$0.018} & 0.672{\fontsize{5.59}{5.59}\selectfont$\pm$0.012} & 0.677{\fontsize{5.59}{5.59}\selectfont$\pm$0.017} & 0.720{\fontsize{5.59}{5.59}\selectfont$\pm$0.022} & \textbf{0.726}{\fontsize{5.59}{5.59}\selectfont$\pm$0.023}\\
\midrule
\clinicalgroup{FDF1E6}{MIMIC-III (ICU) --- $R^2\uparrow$}
Age & 0.295{\fontsize{5.59}{5.59}\selectfont$\pm$0.039} & 0.308{\fontsize{5.59}{5.59}\selectfont$\pm$0.031} & 0.468{\fontsize{5.59}{5.59}\selectfont$\pm$0.029} & 0.288{\fontsize{5.59}{5.59}\selectfont$\pm$0.036} & 0.347{\fontsize{5.59}{5.59}\selectfont$\pm$0.036} & 0.367{\fontsize{5.59}{5.59}\selectfont$\pm$0.028} & 0.502{\fontsize{5.59}{5.59}\selectfont$\pm$0.030} & \textbf{0.507}{\fontsize{5.59}{5.59}\selectfont$\pm$0.028}\\
BUN & 0.029{\fontsize{5.59}{5.59}\selectfont$\pm$0.011} & 0.029{\fontsize{5.59}{5.59}\selectfont$\pm$0.009} & 0.065{\fontsize{5.59}{5.59}\selectfont$\pm$0.019} & 0.034{\fontsize{5.59}{5.59}\selectfont$\pm$0.015} & 0.055{\fontsize{5.59}{5.59}\selectfont$\pm$0.015} & 0.074{\fontsize{5.59}{5.59}\selectfont$\pm$0.012} & 0.081{\fontsize{5.59}{5.59}\selectfont$\pm$0.012} & \textbf{0.086}{\fontsize{5.59}{5.59}\selectfont$\pm$0.013}\\
Hemoglobin & 0.035{\fontsize{5.59}{5.59}\selectfont$\pm$0.010} & 0.048{\fontsize{5.59}{5.59}\selectfont$\pm$0.010} & 0.079{\fontsize{5.59}{5.59}\selectfont$\pm$0.023} & 0.034{\fontsize{5.59}{5.59}\selectfont$\pm$0.009} & 0.046{\fontsize{5.59}{5.59}\selectfont$\pm$0.012} & 0.076{\fontsize{5.59}{5.59}\selectfont$\pm$0.018} & 0.092{\fontsize{5.59}{5.59}\selectfont$\pm$0.024} & \textbf{0.095}{\fontsize{5.59}{5.59}\selectfont$\pm$0.020}\\
\midrule
\clinicalgroup{F2ECFC}{VitalDB (OR) --- AUROC $\uparrow$}
Sex & 0.586{\fontsize{5.59}{5.59}\selectfont$\pm$0.023} & 0.626{\fontsize{5.59}{5.59}\selectfont$\pm$0.022} & 0.658{\fontsize{5.59}{5.59}\selectfont$\pm$0.016} & 0.616{\fontsize{5.59}{5.59}\selectfont$\pm$0.016} & 0.620{\fontsize{5.59}{5.59}\selectfont$\pm$0.017} & 0.608{\fontsize{5.59}{5.59}\selectfont$\pm$0.016} & 0.684{\fontsize{5.59}{5.59}\selectfont$\pm$0.012} & \textbf{0.691}{\fontsize{5.59}{5.59}\selectfont$\pm$0.013}\\
ASA $\geq 3$ & 0.606{\fontsize{5.59}{5.59}\selectfont$\pm$0.021} & 0.642{\fontsize{5.59}{5.59}\selectfont$\pm$0.020} & 0.690{\fontsize{5.59}{5.59}\selectfont$\pm$0.026} & 0.605{\fontsize{5.59}{5.59}\selectfont$\pm$0.035} & 0.625{\fontsize{5.59}{5.59}\selectfont$\pm$0.035} & 0.655{\fontsize{5.59}{5.59}\selectfont$\pm$0.040} & 0.699{\fontsize{5.59}{5.59}\selectfont$\pm$0.034} & \textbf{0.704}{\fontsize{5.59}{5.59}\selectfont$\pm$0.025}\\
Hypertension & 0.598{\fontsize{5.59}{5.59}\selectfont$\pm$0.016} & 0.607{\fontsize{5.59}{5.59}\selectfont$\pm$0.020} & 0.635{\fontsize{5.59}{5.59}\selectfont$\pm$0.019} & 0.595{\fontsize{5.59}{5.59}\selectfont$\pm$0.025} & 0.610{\fontsize{5.59}{5.59}\selectfont$\pm$0.016} & 0.617{\fontsize{5.59}{5.59}\selectfont$\pm$0.023} & \textbf{0.656}{\fontsize{5.59}{5.59}\selectfont$\pm$0.016} & 0.655{\fontsize{5.59}{5.59}\selectfont$\pm$0.017}\\
ICU admission & 0.690{\fontsize{5.59}{5.59}\selectfont$\pm$0.009} & 0.700{\fontsize{5.59}{5.59}\selectfont$\pm$0.004} & 0.747{\fontsize{5.59}{5.59}\selectfont$\pm$0.002} & 0.683{\fontsize{5.59}{5.59}\selectfont$\pm$0.009} & 0.662{\fontsize{5.59}{5.59}\selectfont$\pm$0.011} & 0.714{\fontsize{5.59}{5.59}\selectfont$\pm$0.009} & 0.733{\fontsize{5.59}{5.59}\selectfont$\pm$0.004} & \textbf{0.751}{\fontsize{5.59}{5.59}\selectfont$\pm$0.006}\\
\midrule
\clinicalgroup{F2ECFC}{VitalDB (OR) --- $R^2\uparrow$}
Age & 0.240{\fontsize{5.59}{5.59}\selectfont$\pm$0.026} & 0.263{\fontsize{5.59}{5.59}\selectfont$\pm$0.036} & 0.403{\fontsize{5.59}{5.59}\selectfont$\pm$0.024} & 0.259{\fontsize{5.59}{5.59}\selectfont$\pm$0.026} & 0.302{\fontsize{5.59}{5.59}\selectfont$\pm$0.025} & 0.337{\fontsize{5.59}{5.59}\selectfont$\pm$0.032} & 0.418{\fontsize{5.59}{5.59}\selectfont$\pm$0.030} & \textbf{0.435}{\fontsize{5.59}{5.59}\selectfont$\pm$0.029}\\
BMI & 0.015{\fontsize{5.59}{5.59}\selectfont$\pm$0.005} & 0.022{\fontsize{5.59}{5.59}\selectfont$\pm$0.003} & 0.036{\fontsize{5.59}{5.59}\selectfont$\pm$0.008} & 0.026{\fontsize{5.59}{5.59}\selectfont$\pm$0.005} & 0.035{\fontsize{5.59}{5.59}\selectfont$\pm$0.003} & 0.013{\fontsize{5.59}{5.59}\selectfont$\pm$0.010} & 0.041{\fontsize{5.59}{5.59}\selectfont$\pm$0.007} & \textbf{0.043}{\fontsize{5.59}{5.59}\selectfont$\pm$0.007}\\
Albumin & 0.075{\fontsize{5.59}{5.59}\selectfont$\pm$0.026} & 0.085{\fontsize{5.59}{5.59}\selectfont$\pm$0.034} & 0.136{\fontsize{5.59}{5.59}\selectfont$\pm$0.046} & 0.085{\fontsize{5.59}{5.59}\selectfont$\pm$0.037} & 0.089{\fontsize{5.59}{5.59}\selectfont$\pm$0.038} & 0.101{\fontsize{5.59}{5.59}\selectfont$\pm$0.040} & 0.134{\fontsize{5.59}{5.59}\selectfont$\pm$0.054} & \textbf{0.138}{\fontsize{5.59}{5.59}\selectfont$\pm$0.048}\\
EtCO$_2$$^\dagger$ & 0.098{\fontsize{5.59}{5.59}\selectfont$\pm$0.023} & 0.121{\fontsize{5.59}{5.59}\selectfont$\pm$0.034} & 0.105{\fontsize{5.59}{5.59}\selectfont$\pm$0.028} & 0.113{\fontsize{5.59}{5.59}\selectfont$\pm$0.024} & 0.133{\fontsize{5.59}{5.59}\selectfont$\pm$0.029} & 0.150{\fontsize{5.59}{5.59}\selectfont$\pm$0.031} & \textbf{0.153}{\fontsize{5.59}{5.59}\selectfont$\pm$0.038} & 0.152{\fontsize{5.59}{5.59}\selectfont$\pm$0.037}\\
BIS$^\dagger$ & 0.380{\fontsize{5.59}{5.59}\selectfont$\pm$0.019} & 0.398{\fontsize{5.59}{5.59}\selectfont$\pm$0.022} & 0.393{\fontsize{5.59}{5.59}\selectfont$\pm$0.021} & 0.387{\fontsize{5.59}{5.59}\selectfont$\pm$0.020} & 0.402{\fontsize{5.59}{5.59}\selectfont$\pm$0.021} & 0.404{\fontsize{5.59}{5.59}\selectfont$\pm$0.024} & 0.407{\fontsize{5.59}{5.59}\selectfont$\pm$0.022} & \textbf{0.408}{\fontsize{5.59}{5.59}\selectfont$\pm$0.022}\\
\bottomrule
\end{tabular}}
\end{table}

\subsection{Few-Shot Adaptation}
\label{app:experiment-fewshot}

We assess how labeled support affects adaptation across MC-MED, MIMIC-III,
and VitalDB, complementing the clinical-transfer evaluation in
Section~\ref{sec:exp-repr}.

\textbf{Few-shot adaptation of \ppglm{}.}
We fit linear classifiers on frozen representations using
$k\in\{5,10,20,50\}$ labeled examples per class from a support pool that is
patient-disjoint from the query set. Table~\ref{tab:fewshot-extra} reports
macro-F1 as mean $\pm$ standard deviation over five support draws; columns
correspond to the four per-class label budgets.
Tasks cover rhythm regularity, heart-rate (HR) class, sex, and dataset-specific
outcomes: ED admission on MC-MED, in-hospital mortality on MIMIC-III, and
ICU admission on VitalDB.
HR-class performance increases consistently with more support, reaching
0.921, 0.912, and 0.873 at $k=50$ on the three datasets, respectively.
Eleven of twelve tasks attain their highest mean at $k=50$; the exception is
VitalDB ICU admission, which peaks at $k=10$ (0.591) and falls to 0.557 at
$k=50$. Thus, additional labels generally improve adaptation, but gains
remain task-dependent.

For comparison, the separate LLM in-context protocol uses ten labeled windows
in total, rather than $k$ examples per class. On MC-MED, PulseLM's linear probes
at $k=10/50$ reach rhythm macro-F1 of 0.395/0.403 and HR-class macro-F1 of
0.750/0.811. These evaluations use distinct adaptation procedures and should
be interpreted under their respective support budgets.

\begin{table}[!htbp]
\centering\small
\caption{\textbf{Few-shot adaptation of \ppglm{}.}
Macro-F1 (mean $\pm$ SD; $\uparrow$) over five support draws.
VitalDB uses the 40\,Hz waveform view.
$k$: examples per class. Bold: best mean per row.}
\label{tab:fewshot-extra}
\footnotesize
\setlength{\tabcolsep}{3pt}
\renewcommand{\arraystretch}{0.96}
\newcommand{\fewshotgroup}[2]{%
  \rowcolor[HTML]{#1}[0pt][0pt]%
  \multicolumn{5}{@{}l@{}}{\strut\textbf{\textit{#2}}}\\}
\begin{tabular*}{\linewidth}{@{\extracolsep{\fill}}lcccc@{}}
\toprule
Task & $k=5$ & $k=10$ & $k=20$ & $k=50$\\
\midrule
\fewshotgroup{ECF4FF}{MC-MED (ED) --- Macro-F1 $\uparrow$}
Rhythm & 0.488{\fontsize{5.59}{5.59}\selectfont$\pm$0.051} & 0.537{\fontsize{5.59}{5.59}\selectfont$\pm$0.019} & 0.567{\fontsize{5.59}{5.59}\selectfont$\pm$0.037} & \textbf{0.597}{\fontsize{5.59}{5.59}\selectfont$\pm$0.022}\\
HR class & 0.755{\fontsize{5.59}{5.59}\selectfont$\pm$0.023} & 0.831{\fontsize{5.59}{5.59}\selectfont$\pm$0.025} & 0.898{\fontsize{5.59}{5.59}\selectfont$\pm$0.018} & \textbf{0.921}{\fontsize{5.59}{5.59}\selectfont$\pm$0.003}\\
Sex & 0.568{\fontsize{5.59}{5.59}\selectfont$\pm$0.042} & 0.645{\fontsize{5.59}{5.59}\selectfont$\pm$0.054} & 0.696{\fontsize{5.59}{5.59}\selectfont$\pm$0.041} & \textbf{0.710}{\fontsize{5.59}{5.59}\selectfont$\pm$0.026}\\
ED admission & 0.563{\fontsize{5.59}{5.59}\selectfont$\pm$0.014} & 0.558{\fontsize{5.59}{5.59}\selectfont$\pm$0.026} & 0.583{\fontsize{5.59}{5.59}\selectfont$\pm$0.018} & \textbf{0.617}{\fontsize{5.59}{5.59}\selectfont$\pm$0.008}\\
\midrule
\fewshotgroup{FDF1E6}{MIMIC-III (ICU) --- Macro-F1 $\uparrow$}
Rhythm & 0.506{\fontsize{5.59}{5.59}\selectfont$\pm$0.029} & 0.585{\fontsize{5.59}{5.59}\selectfont$\pm$0.037} & 0.572{\fontsize{5.59}{5.59}\selectfont$\pm$0.025} & \textbf{0.591}{\fontsize{5.59}{5.59}\selectfont$\pm$0.021}\\
HR class & 0.757{\fontsize{5.59}{5.59}\selectfont$\pm$0.050} & 0.859{\fontsize{5.59}{5.59}\selectfont$\pm$0.023} & 0.891{\fontsize{5.59}{5.59}\selectfont$\pm$0.017} & \textbf{0.912}{\fontsize{5.59}{5.59}\selectfont$\pm$0.006}\\
Sex & 0.505{\fontsize{5.59}{5.59}\selectfont$\pm$0.058} & 0.538{\fontsize{5.59}{5.59}\selectfont$\pm$0.039} & 0.562{\fontsize{5.59}{5.59}\selectfont$\pm$0.040} & \textbf{0.580}{\fontsize{5.59}{5.59}\selectfont$\pm$0.033}\\
In-hospital mortality & 0.548{\fontsize{5.59}{5.59}\selectfont$\pm$0.053} & 0.576{\fontsize{5.59}{5.59}\selectfont$\pm$0.042} & 0.597{\fontsize{5.59}{5.59}\selectfont$\pm$0.042} & \textbf{0.636}{\fontsize{5.59}{5.59}\selectfont$\pm$0.015}\\
\midrule
\fewshotgroup{F2ECFC}{VitalDB (OR) --- Macro-F1 $\uparrow$}
Rhythm & 0.375{\fontsize{5.59}{5.59}\selectfont$\pm$0.023} & 0.365{\fontsize{5.59}{5.59}\selectfont$\pm$0.013} & 0.360{\fontsize{5.59}{5.59}\selectfont$\pm$0.020} & \textbf{0.399}{\fontsize{5.59}{5.59}\selectfont$\pm$0.006}\\
HR class & 0.761{\fontsize{5.59}{5.59}\selectfont$\pm$0.039} & 0.832{\fontsize{5.59}{5.59}\selectfont$\pm$0.043} & 0.862{\fontsize{5.59}{5.59}\selectfont$\pm$0.026} & \textbf{0.873}{\fontsize{5.59}{5.59}\selectfont$\pm$0.011}\\
Sex & 0.498{\fontsize{5.59}{5.59}\selectfont$\pm$0.018} & 0.501{\fontsize{5.59}{5.59}\selectfont$\pm$0.039} & 0.496{\fontsize{5.59}{5.59}\selectfont$\pm$0.035} & \textbf{0.558}{\fontsize{5.59}{5.59}\selectfont$\pm$0.013}\\
ICU admission & 0.543{\fontsize{5.59}{5.59}\selectfont$\pm$0.033} & \textbf{0.591}{\fontsize{5.59}{5.59}\selectfont$\pm$0.017} & 0.552{\fontsize{5.59}{5.59}\selectfont$\pm$0.045} & 0.557{\fontsize{5.59}{5.59}\selectfont$\pm$0.041}\\
\bottomrule
\end{tabular*}
\end{table}

\clearpage
\subsection{Objective ablations}
\label{app:objective-ablations}

\begin{table}[!htbp]
\centering\small
\caption{\textbf{Objective ablations on MC-MED.}
Rhythm/quality: macro-F1; retrieval: signal-to-text R@1 with 2,000 candidates.
Stage-2 CLIP/CoCa report per-metric bests across configurations.}
\label{tab:language-ablations}
\setlength{\tabcolsep}{3pt}
\begin{tabular*}{0.72\linewidth}{@{\extracolsep{\fill}}lccc@{}}
\toprule
Variant & Rhythm~$\uparrow$ & Quality~$\uparrow$ & R@1~$\uparrow$\\
\midrule
\rowcolor[HTML]{FCEBDD}[0pt][0pt]
\multicolumn{4}{@{}l@{}}{\hspace{3pt}\textbf{Stage 1}}\\
\midrule
CLIP & 0.650 & 0.505 & 0.790\\
SigLIP & 0.620 & 0.400 & 0.815\\
CoCa & 0.640 & 0.614 & 0.783\\
\midrule
\rowcolor[HTML]{E0F0EB}[0pt][0pt]
\multicolumn{4}{@{}l@{}}{\hspace{3pt}\textbf{Stage 2}}\\
\midrule
CLIP & 0.534 & 0.580 & 0.797\\
SigLIP & 0.600 & 0.470 & \textbf{0.820}\\
CoCa & \textbf{0.651} & \textbf{0.673} & \textbf{0.820}\\
\bottomrule
\end{tabular*}
\end{table}

Table~\ref{tab:language-ablations} compares three segment objectives across
training stages. CLIP~\citep{clip} uses symmetric InfoNCE,
$\mathcal{L}_{\mathrm{CLIP}}=\mathcal{L}_{\mathrm{NCE}}(Q;\mathcal{C})$,
while SigLIP~\citep{siglip} uses the pairwise sigmoid loss
$\ell_{ij}=-\log\sigma[y_{ij}(Q_{ij}+b)]$, where $y_{ij}\in\{+1,-1\}$
labels matched/mismatched pairs and $b$ is a learned bias.
CoCa additionally optimizes caption likelihood.
At Stage~1, CLIP leads rhythm recognition, SigLIP leads retrieval,
and CoCa leads quality assessment. At Stage~2, CoCa improves all three
metrics, achieving the highest rhythm and quality macro-F1
and matching SigLIP's R@1. This combination supports CoCa for
attribute recognition and retrieval while retaining caption generation.
The Stage-2 CoCa R@1 of 0.820 uses signal-only captions; the main model's
signal-plus-vitals configuration reaches 0.780 at the same candidate-pool
size (Table~\ref{tab:retrieval-extra}).

\textbf{Component-ablation setting.}
Table~\ref{tab:ablation} uses the CLIP configuration on MC-MED
validation, with shared Stage-1 initialization and matched Stage-2 training
budgets. Temporal accuracy distinguishes true from corrupted statements;
event AUROC compares any medication event against no-administration controls;
segment R@1 measures segment--text retrieval.

\subsection{Prompt sensitivity}
\label{app:experiment-prompts}

We examine whether expanding the class descriptions improves zero-shot
recognition while keeping the model and evaluation items fixed.
The \emph{Template} bank contains the caption-style descriptions used in the
main experiments; the \emph{Combined} bank augments them with reworded variants.
For each class, we average its text embeddings and score the resulting
representation against the waveform embedding by cosine similarity.
Table~\ref{tab:prompt-sensitivity} reports macro-F1 for three-class rhythm,
five-class heart rate (HR), and three-class signal quality on MC-MED,
MIMIC-III, and VitalDB.

Template prompts achieve higher rhythm and quality scores on all three
datasets. Combined prompts improve HR-class macro-F1 from 0.355 to 0.393 on
MC-MED and from 0.337 to 0.350 on MIMIC-III, but reduce it from 0.327 to
0.282 on VitalDB. The template bank thus performs better on seven of the
nine dataset--task pairs. Expanding the prompt bank does not consistently
improve recognition, indicating that the choice of class descriptions
remains consequential even when the model is unchanged.

\begin{table}[!htbp]
\centering\small
\caption{\textbf{Prompt sensitivity of \ppglm{}.}
Template uses caption-style class descriptions; Combined adds reworded variants.
Macro-F1 ($\uparrow$); bold marks the best score per column.}
\label{tab:prompt-sensitivity}
\setlength{\tabcolsep}{3pt}
\begin{tabular*}{\linewidth}{@{\extracolsep{\fill}}lccccccccc@{}}
\toprule
 & \multicolumn{3}{c}{\cellcolor[HTML]{ECF4FF}\textbf{MC-MED}}
 & \multicolumn{3}{c}{\cellcolor[HTML]{FDF1E6}\textbf{MIMIC-III}}
 & \multicolumn{3}{c}{\cellcolor[HTML]{F2ECFC}\textbf{VitalDB}}\\
\cmidrule(lr){2-4}\cmidrule(lr){5-7}\cmidrule(lr){8-10}
Prompt bank & Rhythm & HR class & Quality & Rhythm & HR class & Quality & Rhythm & HR class & Quality\\
\midrule
Template & \textbf{0.651} & 0.355 & \textbf{0.673} & \textbf{0.670} & 0.337 & \textbf{0.786} & \textbf{0.389} & \textbf{0.327} & \textbf{0.583}\\
Combined & 0.629 & \textbf{0.393} & 0.592 & 0.642 & \textbf{0.350} & 0.677 & 0.384 & 0.282 & 0.515\\
\bottomrule
\end{tabular*}
\end{table}

\clearpage
\section{Language Templates and Examples}
\label{app:prompts}

This appendix presents the language used for training and evaluation. Training captions are rendered from structured measurements and clinical records using fixed template banks; the example identifier and random seed determine template selection and sentence order. No external LLM is used. We distinguish these training targets from zero-shot class descriptions and the model's learned text-prefix embeddings.

\subsection{Representative Caption Templates}
\label{app:caption-templates}

\textbf{Caption content.}
Table~\ref{tab:caption-template-examples} reproduces representative template strings, with braces marking substituted fields. Segment captions describe available rate, rhythm, and quality evidence; missing features are omitted, and poor-quality segments receive a quality-only description. Event captions identify the medication class associated with the sampled windows. Visit captions combine available demographics, triage context, and prior history.

\begin{table}[!htbp]
\centering
\small
\setlength{\tabcolsep}{6pt}
\renewcommand{\arraystretch}{1.12}
\caption{\textbf{Representative training templates.} Strings are reproduced verbatim; braced fields are filled from measurements or records.}
\label{tab:caption-template-examples}
\begin{tabular*}{\linewidth}{>{\raggedright\arraybackslash}p{\dimexpr0.26\linewidth-12pt\relax}>{\raggedright\arraybackslash}p{\dimexpr0.74\linewidth-12pt\relax}}
\toprule
\rowcolor[gray]{0.94}
\textbf{Target} & \textbf{Template} \\
\midrule
Segment rate & The pulse rate is about \{val\} beats per minute, \{level\}. \\
Segment rhythm & The cardiac rhythm is \{level\}. \\
Segment quality & The PPG trace is \{level\}. \\
Poor-quality segment & This segment's pulse waveform is too noisy to characterize reliably. \\
\addlinespace[3pt]
Medication event & During this period the patient was given \{desc\}. \\
Control interval & This period of monitoring contains no medication administration. \\
\addlinespace[3pt]
Chief complaint & The chief complaint at triage was \{cc\}. \\
\bottomrule
\end{tabular*}
\end{table}

\textbf{Temporal captions and negatives.}
Temporal templates express occurrence, prevalence, contiguous duration, or temporal order. A negative caption replaces one eligible sentence and retains the others. For example, the implemented positive sentence ``Over the stay the pulse rate trended upward.'' can be replaced by ``Over the stay the pulse rate trended downward.'' This pair illustrates the corruption rule, rather than an observed trajectory.

\subsection{Zero-Shot Evaluation Prompts}
\label{app:zero-shot-prompts}

\textbf{Segment scoring.}
Table~\ref{tab:evaluation-prompt-examples} reproduces representative prompts from the language benchmark in Section~\ref{sec:experiments}. The main segment-recognition results use two caption-style descriptions per class, whose embeddings are averaged for cosine scoring. Appendix~\ref{app:experiment-prompts} compares this template bank with a combined bank that adds reworded class descriptions. Quality recognition uses a separate clean/fair/noisy prompt bank.

\begin{table}[!htbp]
\centering
\small
\setlength{\tabcolsep}{6pt}
\renewcommand{\arraystretch}{1.10}
\caption{\textbf{Representative zero-shot prompts.} Verbatim examples from the segment template bank and the visit and event class descriptions, grouped by task.}
\label{tab:evaluation-prompt-examples}
\begin{tabular*}{\linewidth}{>{\raggedright\arraybackslash}p{\dimexpr0.26\linewidth-12pt\relax}>{\raggedright\arraybackslash}p{\dimexpr0.74\linewidth-12pt\relax}}
\toprule
\rowcolor[gray]{0.94}
\textbf{Concept} & \textbf{Prompt}\\
\midrule
\multicolumn{2}{@{}l}{\textit{Pulse rate}}\\
Marked bradycardia & The measured pulse rate is markedly bradycardic.\\
Bradycardia & The measured pulse rate is bradycardic.\\
Normal rate & The measured pulse rate is within the normal range.\\
Mild tachycardia & The measured pulse rate is mildly tachycardic.\\
Tachycardia & The measured pulse rate is tachycardic.\\
\addlinespace[3pt]
\multicolumn{2}{@{}l}{\textit{Rhythm}}\\
Regular rhythm & The pulse rhythm appears regular.\\
Irregular rhythm & The pulse rhythm appears slightly irregular.\\
Marked irregularity & The pulse rhythm appears markedly irregular.\\
\addlinespace[3pt]
\multicolumn{2}{@{}l}{\textit{Signal quality}}\\
Clean signal & Pulse-wave quality here is clean and reliable.\\
Fair signal & Pulse-wave quality here is of fair quality.\\
Noisy signal & The signal is somewhat noisy but usable for analysis.\\
\addlinespace[3pt]
\multicolumn{2}{@{}l}{\textit{Visit attributes}}\\
Visit acuity & The visit is an emergent, high-acuity presentation.\\
Visit demographics & The recording is from a male patient.\\
Visit complaint & The visit is for a patient presenting with chest pain.\\
\addlinespace[3pt]
\multicolumn{2}{@{}l}{\textit{Medication events}}\\
Medication class & The patient received a bronchodilator during this period.\\
Control & No medication was given during this period.\\
\bottomrule
\end{tabular*}
\end{table}

\textbf{Visit and event scoring.}
Visit prompts describe individual attributes and are scored against the selected visit read-out. The MC-MED event bank contains ten medication classes and a no-medication description. For the supplementary any-administration diagnostic in Appendix~\ref{app:experiment-events}, the score is the highest medication-class score minus the no-medication score. The main medication table instead reports separate class-versus-control AUROCs.

\newpage
\subsection{Synthetic Caption Examples and Decoder Context}
\label{app:caption-examples}

These captions were rendered from invented feature values and clinical fields using the captioning pipeline. They illustrate training targets; they are neither patient records nor model-generated outputs. Missing fields contribute no sentences.

\begin{samepage}
\textbf{Segment caption.}
\begin{quote}
Overall the PPG signal is clean and reliable. Over the window the pulse rate stays stable. The heartbeat is regular across the window. Heart rate estimated from the pulse waveform is around 82 bpm, within the normal range.
\end{quote}
\end{samepage}

\begin{samepage}
\textbf{Event caption.}
\begin{quote}
This segment covers the time around a dose of an inhaled bronchodilator.
\end{quote}
\end{samepage}

\begin{samepage}
\textbf{Visit caption with history.}
\begin{quote}
The individual is a 67-year-old male patient (older adult). Arrival vital signs: heart rate 108 bpm (mildly tachycardic), SpO2 96\% (normal). The patient got to the ED by ambulance (EMS). Emergency severity was emergent, ESI level 2. Chief complaint: shortness of breath. Past medical history includes essential hypertension. The home medication list includes statins.
\end{quote}
\end{samepage}

\textbf{CoCa decoder context.}
CoCa tokenizes each segment caption into a beginning-of-sequence token, caption subwords, and an end-of-sequence token. The shared text encoder prepends the learned segment prefix $e_{\mathrm{seg}}$ (denoted \texttt{[WIN]} in the implementation), which is a vector rather than a literal word. After removing the prefix position, the causal text states enter the separate multimodal decoder, which cross-attends to waveform memory $M_i$ and predicts the next caption token; padding targets are ignored. Training uses no natural-language generation prompt. The language benchmark additionally evaluates greedy waveform-conditioned decoding alongside retrieval and class-prompt scoring.

%% file: reference.bib
@article{ppgpt,
  title   = {{PPGPT}: Transferring Next-Token Modeling from Language to {PPG} Signals},
  author  = {Zhang, Zexing and Lu, Huimin and Zhao, Qingxin},
  journal = {Proceedings of the AAAI Conference on Artificial Intelligence},
  volume  = {40},
  number  = {34},
  pages   = {28573--28581},
  year    = {2026},
  doi     = {10.1609/aaai.v40i34.40088}
}

@inproceedings{llmbp,
  title     = {Large Language Models for Cuffless Blood Pressure Measurement
               From Wearable Biosignals},
  author    = {Liu, Zengding and Chen, Chen and Cao, Jiannong and Pan, Minglei
               and Liu, Jikui and Li, Nan and Miao, Fen and Li, Ye},
  booktitle = {Proceedings of the 15th ACM International Conference on
               Bioinformatics, Computational Biology and Health Informatics},
  pages     = {1--11},
  year      = {2024},
  publisher = {Association for Computing Machinery},
  doi       = {10.1145/3698587.3701447}
}

@inproceedings{peftvitals,
  title     = {{PEFT QLORA}-Based Fine-Tuning of Foundation Models for Vitals
               Estimation Using {PPG} and {ECG}-Based Medical {IoT} Data:
               A Feasibility Study},
  author    = {Ali, Syed Anas and Nawaz, Muhammad Wasim and Rashid, Junaid
               and Mahmood, Ali Hamid and Kim, Jungeun
               and Rahman, Muhammad Mahboob Ur and Abbasi, Qammer H.},
  booktitle = {2025 IEEE International Conference on Data Mining Workshops (ICDMW)},
  pages     = {468--477},
  year      = {2025},
  publisher = {IEEE Computer Society},
  doi       = {10.1109/ICDMW69685.2025.00059}
}

@article{ppgreport,
  title   = {Multimodal {PPG}-Based Arrhythmia Detection Using a
             {CLIP}-Initialized Multi-Task {U-Net} and {LLM}-Assisted Reporting},
  author  = {Huh, Youngho and Noh, Minhwan and Ji, Dongwoo
             and Oh, Yuna and Sun, Sukkyu},
  journal = {Sensors},
  volume  = {26},
  number  = {8},
  pages   = {2316},
  year    = {2026},
  doi     = {10.3390/s26082316}
}

@inproceedings{physllm,
  title     = {{PhysLLM}: Harnessing Large Language Models for
               Cross-Modal Remote Physiological Sensing},
  author    = {Xie, Yiping and Zhao, Bo and Dai, Mingtong and Zhou, Jian-Ping
               and Sun, Yue and Tan, Tao and Xie, Weicheng
               and Shen, Linlin and Yu, Zitong},
  booktitle = {The Fourteenth International Conference on Learning Representations},
  year      = {2026},
  url       = {https://openreview.net/forum?id=aR43t8OEeW}
}

@inproceedings{biovilt,
  title     = {Learning to Exploit Temporal Structure for Biomedical
               Vision-Language Processing},
  author    = {Bannur, Shruthi and Hyland, Stephanie and Liu, Qianchu
               and P{\'e}rez-Garc{\'\i}a, Fernando and Ilse, Maximilian
               and Castro, Daniel C. and Boecking, Benedikt and Sharma, Harshita
               and Bouzid, Kenza and Thieme, Anja and Schwaighofer, Anton
               and Wetscherek, Maria and Lungren, Matthew P. and Nori, Aditya
               and Alvarez-Valle, Javier and Oktay, Ozan},
  booktitle = {Proceedings of the IEEE/CVF Conference on Computer Vision and Pattern Recognition},
  pages     = {15016--15027},
  year      = {2023}
}

@article{allen2007ppg,
  title   = {Photoplethysmography and its application in clinical physiological measurement},
  author  = {Allen, John},
  journal = {Physiological Measurement},
  volume  = {28},
  number  = {3},
  pages   = {R1--R39},
  year    = {2007},
  doi     = {10.1088/0967-3334/28/3/R01}
}

@article{charlton2022wearable,
  title   = {Wearable Photoplethysmography for Cardiovascular Monitoring},
  author  = {Charlton, Peter H. and Kyriacou, Panicos A. and Mant, Jonathan
             and Marozas, Vaidotas and Chowienczyk, Phil and Alastruey, Jordi},
  journal = {Proceedings of the IEEE},
  volume  = {110},
  number  = {3},
  pages   = {355--381},
  year    = {2022},
  doi     = {10.1109/JPROC.2022.3149785}
}

@inproceedings{papagei,
  title     = {{PaPaGei}: Open Foundation Models for Optical Physiological Signals},
  author    = {Pillai, Arvind and Spathis, Dimitris and Kawsar, Fahim
               and Malekzadeh, Mohammad},
  booktitle = {The Thirteenth International Conference on Learning Representations},
  year      = {2025},
  url       = {https://openreview.net/forum?id=kYwTmlq6Vn}
}

@article{pulseppg,
  title     = {{Pulse-PPG}: An Open-Source Field-Trained {PPG} Foundation Model
               for Wearable Applications across Lab and Field Settings},
  author    = {Saha, Mithun and Xu, Maxwell A. and Mao, Wanting
               and Neupane, Sameer and Rehg, James M. and Kumar, Santosh},
  journal   = {Proceedings of the ACM on Interactive, Mobile, Wearable
               and Ubiquitous Technologies},
  volume    = {9},
  number    = {3},
  pages     = {1--35},
  year      = {2025},
  publisher = {Association for Computing Machinery},
  doi       = {10.1145/3749494}
}

@inproceedings{anyppg,
  title={Anyppg: An ecg-guided ppg foundation model trained on over 100,000 hours of recordings for holistic health profiling},
  author={Nie, Guangkun and Fang, Xiaocheng and Tang, Gongzheng and Xiao, Yujie and Li, Jun and Liu, Bo and Li, Hongyan and Hong, Shenda},
  booktitle={Proceedings of the 32nd ACM SIGKDD Conference on Knowledge Discovery and Data Mining V. 2},
  pages={11717--11727},
  year={2026}
}

@article{sensorlm,
  title={SensorLM: Learning the language of wearable sensors},
  author={Zhang, Yuwei and Ayush, Kumar and Qiao, Siyuan and Heydari, A Ali and Narayanswamy, Girish and Xu, Max and Metwally, Ahmed and Xu, Jinhua and Garrison, Jake and Xu, Xuhai and others},
  journal={Advances in Neural Information Processing Systems},
  volume={38},
  pages={46813--46849},
  year={2026}
}

@article{chen2026learning,
  title   = {Learning Transferable Sensor Models via Language-Informed Pretraining},
  author  = {Chen, Yuliang and Pillai, Arvind and Wu, Yu Yvonne
             and Griffin, Tess Z. and Marsch, Lisa and Heinz, Michael V.
             and Jacobson, Nicholas C. and Campbell, Andrew},
  journal = {arXiv preprint arXiv:2603.11950},
  year    = {2026},
  doi     = {10.48550/arXiv.2603.11950},
  url     = {https://arxiv.org/abs/2603.11950}
}

@inproceedings{sleeplm,
  title     = {{SleepLM}: Natural-Language Intelligence for Human Sleep},
  author    = {Xu, Zongzhe and Shuai, Zitao and Mozaffari, Eideen
               and Aysola, Ravi S. and Kumar, Rajesh and Yang, Yuzhe},
  booktitle = {Proceedings of the 43rd International Conference on Machine Learning},
  year      = {2026},
  url       = {https://icml.cc/virtual/2026/poster/65807}
}

@inproceedings{opentslm,
  title     = {{OpenTSLM}: Time-Series Language Models for Reasoning over
               Multivariate Medical Text- and Time-Series Data},
  author    = {Langer, Patrick and Kaar, Thomas and Rosenblattl, Max
               and Xu, Maxwell A. and Chow, Winnie and Maritsch, Martin
               and Jakob, Robert and Wang, Ning and Liu, Juncheng
               and Verma, Aradhana and Han, Brian and Kim, Daniel
               and Chubb, Henry and Ceresnak, Scott and Zahedivash, Aydin
               and Sandhu, Alexander and Rodriguez, Fatima and McDuff, Daniel
               and Fleisch, Elgar and Aalami, Oliver and Barata, Filipe
               and Schmiedmayer, Paul},
  booktitle = {Proceedings of the 43rd International Conference on Machine Learning},
  year      = {2026},
  url       = {https://icml.cc/virtual/2026/poster/65261}
}

@inproceedings{wang2026position,
  title     = {Position: Beyond Prediction: Toward Verifiable Physiological
               Waveform Reasoning with Foundation Models and Agentic {LLMs}},
  author    = {Wang, Xiaoda and Chang, Ching and Cao, Defu and Han, Kaiqiao
               and Sun, Fang and Huang, Yue and Wang, Minxiao and Xu, Chang
               and Luo, Xiao and Yan, Runze and Zhang, Xiangliang and Hu, Xiao
               and Liu, Yan and Sun, Yizhou and Wang, Wei and Yang, Carl},
  booktitle = {Proceedings of the 43rd International Conference on Machine Learning},
  year      = {2026},
  url       = {https://icml.cc/virtual/2026/poster/67119}
}

@article{pulselm,
  title   = {{PulseLM}: A Foundation Dataset and Benchmark for {PPG}-Text Learning},
  author  = {Pham, Hung Manh and Wu, Jinyang and Ma, Xiao and Zhang, Yiming
             and Xu, Yixin and Saeed, Aaqib and Zhu, Bin and Pan, Zhou
             and Ma, Dong},
  journal = {arXiv preprint arXiv:2603.03331},
  year    = {2026},
  doi     = {10.48550/arXiv.2603.03331}
}

@inproceedings{cap,
  title     = {{CAP}: Towards {PPG} Universal Representation Learning
               with Patient-level Supervision},
  author    = {He, Chenyang and Shao, Xinyi and Huang, Shun
               and Huang, Bosong and Zhang, Daoqiang and Jin, Ming
               and Ding, Cheng},
  booktitle = {Proceedings of the 32nd ACM SIGKDD Conference on
               Knowledge Discovery and Data Mining V.2},
  pages     = {11050--11060},
  year      = {2026},
  publisher = {Association for Computing Machinery},
  doi       = {10.1145/3770855.3818881}
}

@article{siamquality,
  title   = {{SiamQuality}: A {ConvNet}-Based Foundation Model for
             Photoplethysmography Signals},
  author  = {Ding, Cheng and Guo, Zhicheng and Chen, Zhaoliang
             and Lee, Randall J. and Rudin, Cynthia and Hu, Xiao},
  journal = {Physiological Measurement},
  volume  = {45},
  number  = {8},
  pages   = {085004},
  year    = {2024},
  doi     = {10.1088/1361-6579/ad6747}
}

@article{gptppg,
  title   = {{GPT-PPG}: A {GPT}-Based Foundation Model for
             Photoplethysmography Signals},
  author  = {Chen, Zhaoliang and Ding, Cheng and Kataria, Saurabh
             and Yan, Runze and Wang, Minxiao and Lee, Randall J. and Hu, Xiao},
  journal = {Physiological Measurement},
  volume  = {46},
  number  = {5},
  pages   = {055004},
  year    = {2025},
  doi     = {10.1088/1361-6579/add988}
}

@inproceedings{moment,
  title     = {{MOMENT}: A Family of Open Time-Series Foundation Models},
  author    = {Goswami, Mononito and Szafer, Konrad and Choudhry, Arjun
               and Cai, Yifu and Li, Shuo and Dubrawski, Artur},
  booktitle = {Proceedings of the 41st International Conference on Machine Learning},
  series    = {Proceedings of Machine Learning Research},
  volume    = {235},
  pages     = {16115--16152},
  year      = {2024},
  publisher = {PMLR}
}

@article{chronos2,
  title   = {{Chronos-2}: From Univariate to Universal Forecasting},
  author  = {Ansari, Abdul Fatir and Shchur, Oleksandr and K{\"u}ken, Jaris
             and Auer, Andreas and Han, Boran and Mercado, Pedro
             and Rangapuram, Syama Sundar and Shen, Huibin and Stella, Lorenzo
             and Zhang, Xiyuan and Goswami, Mononito and Kapoor, Shubham
             and Maddix, Danielle C. and Guerron, Pablo and Hu, Tony
             and Yin, Junming and Erickson, Nick and Desai, Prateek Mutalik
             and Wang, Hao and Rangwala, Huzefa and Karypis, George
             and Wang, Yuyang and Bohlke-Schneider, Michael},
  journal = {arXiv preprint arXiv:2510.15821},
  year    = {2025},
  doi     = {10.48550/arXiv.2510.15821}
}

@inproceedings{transformer,
  title     = {Attention Is All You Need},
  author    = {Vaswani, Ashish and Shazeer, Noam and Parmar, Niki
               and Uszkoreit, Jakob and Jones, Llion and Gomez, Aidan N.
               and Kaiser, {\L}ukasz and Polosukhin, Illia},
  booktitle = {Advances in Neural Information Processing Systems},
  volume    = {30},
  year      = {2017},
  url       = {https://papers.nips.cc/paper_files/paper/2017/hash/3f5ee243547dee91fbd053c1c4a845aa-Abstract.html}
}

@article{cpc,
  title   = {Representation Learning with Contrastive Predictive Coding},
  author  = {van den Oord, Aaron and Li, Yazhe and Vinyals, Oriol},
  journal = {arXiv preprint arXiv:1807.03748},
  year    = {2018},
  url     = {https://arxiv.org/abs/1807.03748}
}

@inproceedings{clip,
  title     = {Learning Transferable Visual Models from Natural Language Supervision},
  author    = {Radford, Alec and Kim, Jong Wook and Hallacy, Chris
               and Ramesh, Aditya and Goh, Gabriel and Agarwal, Sandhini
               and Sastry, Girish and Askell, Amanda and Mishkin, Pamela
               and Clark, Jack and Krueger, Gretchen and Sutskever, Ilya},
  booktitle = {Proceedings of the 38th International Conference on Machine Learning},
  series    = {Proceedings of Machine Learning Research},
  volume    = {139},
  pages     = {8748--8763},
  year      = {2021},
  publisher = {PMLR}
}

@inproceedings{siglip,
  title     = {Sigmoid Loss for Language Image Pre-Training},
  author    = {Zhai, Xiaohua and Mustafa, Basil and Kolesnikov, Alexander
               and Beyer, Lucas},
  booktitle = {Proceedings of the IEEE/CVF International Conference on Computer Vision},
  pages     = {11975--11986},
  year      = {2023}
}

@article{coca,
  title   = {{CoCa}: Contrastive Captioners Are Image-Text Foundation Models},
  author  = {Yu, Jiahui and Wang, Zirui and Vasudevan, Vijay and Yeung, Legg
             and Seyedhosseini, Mojtaba and Wu, Yonghui},
  journal = {Transactions on Machine Learning Research},
  year    = {2022},
  url     = {https://openreview.net/forum?id=Ee277P3AYC}
}

@inproceedings{hiervl,
  title     = {{HierVL}: Learning Hierarchical Video-Language Embeddings},
  author    = {Ashutosh, Kumar and Girdhar, Rohit and Torresani, Lorenzo
               and Grauman, Kristen},
  booktitle = {Proceedings of the IEEE/CVF Conference on Computer Vision and Pattern Recognition},
  pages     = {23066--23078},
  year      = {2023}
}

@inproceedings{hecvl,
  title     = {{HecVL}: Hierarchical Video-Language Pretraining for Zero-Shot
               Surgical Phase Recognition},
  author    = {Yuan, Kun and Srivastav, Vinkle and Navab, Nassir and Padoy, Nicolas},
  booktitle = {Medical Image Computing and Computer Assisted Intervention -- MICCAI 2024},
  series    = {Lecture Notes in Computer Science},
  volume    = {15006},
  pages     = {306--316},
  year      = {2024},
  publisher = {Springer Nature Switzerland},
  doi       = {10.1007/978-3-031-72089-5_29}
}

@article{mcmed,
  title     = {{MC-MED}, Multimodal Clinical Monitoring in the Emergency Department},
  author    = {Kansal, Aman and Chen, Emma and Jin, Boyang Tom
               and Rajpurkar, Pranav and Kim, David A.},
  journal   = {Scientific Data},
  volume    = {12},
  number    = {1},
  pages     = {1094},
  year      = {2025},
  publisher = {Nature Publishing Group},
  doi       = {10.1038/s41597-025-05419-5}
}

@article{pyppg,
  title   = {{pyPPG}: A {Python} Toolbox for Comprehensive
             Photoplethysmography Signal Analysis},
  author  = {Goda, M{\'a}rton {\'A}. and Charlton, Peter H. and Behar, Joachim A.},
  journal = {Physiological Measurement},
  volume  = {45},
  number  = {4},
  pages   = {045001},
  year    = {2024},
  doi     = {10.1088/1361-6579/ad33a2}
}

@inproceedings{wearablefm,
  title     = {Large-Scale Training of Foundation Models for Wearable Biosignals},
  author    = {Abbaspourazad, Salar and Elachqar, Oussama and Miller, Andrew C.
               and Emrani, Saba and Nallasamy, Udhyakumar and Shapiro, Ian},
  booktitle = {The Twelfth International Conference on Learning Representations},
  year      = {2024},
  url       = {https://openreview.net/forum?id=pC3WJHf51j}
}

@inproceedings{imu2clip,
  title     = {{IMU}2{CLIP}: Language-Grounded Motion Sensor Translation
               with Multimodal Contrastive Learning},
  author    = {Moon, Seungwhan and Madotto, Andrea and Lin, Zhaojiang
               and Saraf, Aparajita and Bearman, Amy and Damavandi, Babak},
  booktitle = {Findings of the Association for Computational Linguistics: EMNLP 2023},
  pages     = {13246--13253},
  year      = {2023},
  publisher = {Association for Computational Linguistics},
  doi       = {10.18653/v1/2023.findings-emnlp.883}
}

@inproceedings{sensorllm,
  title     = {{SensorLLM}: Aligning Large Language Models with Motion Sensors
               for Human Activity Recognition},
  author    = {Li, Zechen and Deldari, Shohreh and Chen, Linyao
               and Xue, Hao and Salim, Flora D.},
  booktitle = {Proceedings of the 2025 Conference on Empirical Methods
               in Natural Language Processing},
  pages     = {354--379},
  year      = {2025},
  publisher = {Association for Computational Linguistics},
  doi       = {10.18653/v1/2025.emnlp-main.19}
}

@inproceedings{neurolm,
  title={NeuroLM: A universal multi-task foundation model for bridging the gap between language and EEG signals},
  author={Jiang, Wei-Bang and Wang, Yansen and Lu, Bao-Liang and Li, Dongsheng},
  booktitle={International Conference on Learning Representations},
  volume={2025},
  pages={55436--55457},
  year={2025}
}

@inproceedings{gloria,
  title     = {{GLoRIA}: A Multimodal Global-Local Representation Learning
               Framework for Label-Efficient Medical Image Recognition},
  author    = {Huang, Shih-Cheng and Shen, Liyue
               and Lungren, Matthew P. and Yeung, Serena},
  booktitle = {Proceedings of the IEEE/CVF International Conference on Computer Vision},
  pages     = {3942--3951},
  year      = {2021}
}

@inproceedings{mgca,
  title     = {Multi-Granularity Cross-Modal Alignment for Generalized
               Medical Visual Representation Learning},
  author    = {Wang, Fuying and Zhou, Yuyin and Wang, Shujun
               and Vardhanabhuti, Varut and Yu, Lequan},
  booktitle = {Advances in Neural Information Processing Systems},
  volume    = {35},
  pages     = {33536--33549},
  year      = {2022},
  publisher = {Curran Associates, Inc.},
  doi       = {10.52202/068431-2430}
}

@inproceedings{sigmappg,
  title     = {{SIGMA-PPG}: Statistical-Prior Informed Generative Masking
               Architecture for {PPG} Foundation Model},
  author    = {Guo, Zongheng and Chen, Tao and Jiao, Yang
               and Pan, Yi and Hu, Xiao and Ferrario, Manuela},
  booktitle = {Proceedings of the 43rd International Conference on Machine Learning},
  year      = {2026},
  url       = {https://icml.cc/virtual/2026/poster/64268}
}

@inproceedings{robustppg,
  title     = {A Robust {PPG} Foundation Model Using Multimodal
               Physiological Supervision},
  author    = {Geenjaar, Eloy and Calhoun, Vince D. and Daly, Scott
               and KV, Gouthaman and Lu, Lie and Mittal, Trisha and Darcy, Daniel P.},
  booktitle = {Proceedings of the 43rd International Conference on Machine Learning},
  year      = {2026},
  url       = {https://icml.cc/virtual/2026/poster/64846}
}

@inproceedings{eevr,
  title     = {{EEVR}: A Dataset of Paired Physiological Signals and Textual
               Descriptions for Joint Emotion Representation Learning},
  author    = {Singh, Pragya and Budhiraja, Ritvik and Gupta, Ankush
               and Goswami, Anshul and Kumar, Mohan and Singh, Pushpendra},
  booktitle = {Advances in Neural Information Processing Systems},
  volume    = {37},
  pages     = {15765--15778},
  year      = {2024},
  publisher = {Curran Associates, Inc.},
  doi       = {10.52202/079017-0503}
}

@article{neurokit2,
  title   = {{NeuroKit2}: A {Python} Toolbox for Neurophysiological Signal Processing},
  author  = {Makowski, Dominique and Pham, Tam and Lau, Zen J.
             and Brammer, Jan C. and Lespinasse, Fran{\c{c}}ois and Pham, Hung
             and Sch{\"o}lzel, Christopher and Chen, S. H. Annabel},
  journal = {Behavior Research Methods},
  volume  = {53},
  number  = {4},
  pages   = {1689--1696},
  year    = {2021},
  doi     = {10.3758/s13428-020-01516-y}
}

@article{roformer,
  title   = {{RoFormer}: Enhanced Transformer with Rotary Position Embedding},
  author  = {Su, Jianlin and Ahmed, Murtadha and Lu, Yu
             and Pan, Shengfeng and Bo, Wen and Liu, Yunfeng},
  journal = {Neurocomputing},
  volume  = {568},
  pages   = {127063},
  year    = {2024},
  doi     = {10.1016/j.neucom.2023.127063}
}

@inproceedings{colbert,
  title     = {{ColBERT}: Efficient and Effective Passage Search via
               Contextualized Late Interaction over {BERT}},
  author    = {Khattab, Omar and Zaharia, Matei},
  booktitle = {Proceedings of the 43rd International ACM SIGIR Conference
               on Research and Development in Information Retrieval},
  pages     = {39--48},
  year      = {2020},
  publisher = {Association for Computing Machinery},
  doi       = {10.1145/3397271.3401075}
}

@inproceedings{melp,
  title     = {From Token to Rhythm: A Multi-Scale Approach for
               {ECG}-Language Pretraining},
  author    = {Wang, Fuying and Xu, Jiacheng and Yu, Lequan},
  booktitle = {Proceedings of the 42nd International Conference on Machine Learning},
  series    = {Proceedings of Machine Learning Research},
  volume    = {267},
  pages     = {65059--65074},
  year      = {2025},
  publisher = {PMLR}
}

@article{vitaldb,
  title   = {{VitalDB}, a high-fidelity multi-parameter vital signs database in surgical patients},
  author  = {Lee, Hyung-Chul and Park, Yoonsang and Yoon, Soo Bin and Yang, Seong Mi and Park, Dongnyeok and Jung, Chul-Woo},
  journal = {Scientific Data},
  volume  = {9},
  number  = {1},
  pages   = {279},
  year    = {2022},
  doi     = {10.1038/s41597-022-01411-5}
}

@article{mimic3,
  title   = {{MIMIC-III}, a freely accessible critical care database},
  author  = {Johnson, Alistair E. W. and Pollard, Tom J. and Shen, Lu and Lehman, Li-wei H. and Feng, Mengling and Ghassemi, Mohammad and Moody, Benjamin and Szolovits, Peter and Celi, Leo Anthony and Mark, Roger G.},
  journal = {Scientific Data},
  volume  = {3},
  number  = {1},
  pages   = {160035},
  year    = {2016},
  doi     = {10.1038/sdata.2016.35}
}

@inproceedings{mtan,
  title     = {Multi-Time Attention Networks for Irregularly Sampled Time Series},
  author    = {Shukla, Satya Narayan and Marlin, Benjamin M.},
  booktitle = {International Conference on Learning Representations},
  year      = {2021}
}

@inproceedings{perceiver,
  title     = {Perceiver: General Perception with Iterative Attention},
  author    = {Jaegle, Andrew and Gimeno, Felix and Brock, Andrew
               and Vinyals, Oriol and Zisserman, Andrew and Carreira, Jo{\~a}o},
  booktitle = {Proceedings of the 38th International Conference on Machine Learning},
  series    = {Proceedings of Machine Learning Research},
  volume    = {139},
  pages     = {4651--4664},
  year      = {2021},
  publisher = {PMLR}
}

@article{normwear,
  title   = {Toward Foundation Model for Multivariate Wearable Sensing of Physiological Signals},
  author  = {Luo, Yunfei and Chen, Yuliang and Salekin, Asif and Rahman, Tauhidur},
  journal = {arXiv preprint arXiv:2412.09758},
  year    = {2024},
  doi     = {10.48550/arXiv.2412.09758},
  url     = {https://arxiv.org/abs/2412.09758}
}

@article{csfm,
  title   = {Cardiac health assessment across scenarios and devices using a multimodal foundation model pretrained on data from 1.7 million individuals},
  author  = {Gu, Xiao and Tang, Wei and Han, Jinpei and Sangha, Veer and
             Liu, Fenglin and Gowda, Shreyank N. and Ribeiro, Antonio H. and
             Schwab, Patrick and Branson, Kim and Clifton, Lei and
             Ribeiro, Antonio Luiz P. and Liu, Zhangdaihong and Clifton, David A.},
  journal = {Nature Machine Intelligence},
  volume  = {8},
  number  = {2},
  pages   = {220--233},
  year    = {2026},
  doi     = {10.1038/s42256-026-01180-5},
  url     = {https://www.nature.com/articles/s42256-026-01180-5}
}

@misc{gpt56,
  title        = {{GPT-5.6 Luna} Model},
  author       = {{OpenAI}},
  year         = {2026},
  howpublished = {\url{https://developers.openai.com/api/docs/models/gpt-5.6-luna}},
  note         = {OpenAI API documentation. Accessed September 25, 2026}
}

@misc{gpt6luna,
  title        = {{GPT-6 Luna} Model},
  author       = {{OpenAI}},
  year         = {2026},
  howpublished = {\url{https://developers.openai.com/api/docs/models/gpt-6-luna}},
  note         = {OpenAI API documentation. Accessed September 25, 2026}
}

@misc{gpt6sol,
  title        = {{GPT-6 Sol} Model},
  author       = {{OpenAI}},
  year         = {2026},
  howpublished = {\url{https://developers.openai.com/api/docs/models/gpt-6-sol}},
  note         = {OpenAI API documentation. Accessed September 25, 2026}
}

@misc{claudeopus55,
  title        = {Introducing {Claude Opus 5.5}},
  author       = {{Anthropic}},
  year         = {2026},
  month        = sep,
  howpublished = {\url{https://www.anthropic.com/claude-opus-5-5}},
  note         = {Released September 22, 2026. Accessed September 25, 2026}
}

@misc{qwen3,
  title         = {{Qwen3} Technical Report},
  author        = {{Qwen Team}},
  year          = {2025},
  eprint        = {2505.09388},
  archivePrefix = {arXiv},
  primaryClass  = {cs.CL},
  url           = {https://arxiv.org/abs/2505.09388}
}
